%% file: Main_ICML2020.tex
\documentclass{article}

\usepackage{microtype}
\usepackage{graphicx}
\usepackage{subfigure}
\usepackage{booktabs} 

\usepackage{hyperref}

\usepackage[accepted]{icml2020}

\usepackage{url}            
\usepackage{booktabs}       
\usepackage{amsfonts,amsmath}       
\usepackage{nicefrac}       
\usepackage{microtype}      
\usepackage{multirow,times}
\usepackage{graphicx}
\usepackage{makecell}
\usepackage[para,online,flushleft]{threeparttable}
\usepackage{verbatim}
\usepackage{wrapfig}
\usepackage{paralist}
\usepackage[normalem]{ulem}
\usepackage{enumitem}
\usepackage{algorithm}
\usepackage{algpseudocode}
 \usepackage[small,bf]{caption}
\usepackage{mathtools}
 \usepackage{lipsum}
\usepackage{color}
 \usepackage[textsize=scriptsize]{todonotes}
 \usepackage{amssymb,amsthm}
\usepackage{adjustbox}

\DeclarePairedDelimiterX{\inp}[2]{\langle}{\rangle}{#1, #2}
\definecolor{revision_color}{rgb}{0.858, 0.188, 0.478}

\newcommand{\argmax}{\operatornamewithlimits{arg\,max}}

\DeclareMathOperator{\bx}{\mathbf x}
\DeclareMathOperator{\by}{\mathbf y}
\DeclareMathOperator{\bv}{\mathbf v}

\DeclareMathOperator{\bxi}{\boldsymbol {\xi}}

\def\leref#1{Lemma~\ref{#1}}

\def\thref#1{Theorem~\ref{#1}}
\def\coref#1{Corollary~\ref{#1}}
\newtheorem{lemma}{Lemma}
\newtheorem{theorem}{Theorem}
\newtheorem{corollary}{Corollary}

\DeclareMathOperator*{\minimize}{\text{minimize}}
\DeclareMathOperator*{\maximize}{\text{maximize}}
\DeclareMathOperator*{\st}{\text{subject to}}
\DeclareMathAlphabet\mathbfcal{OMS}{cmsy}{b}{n}
\newcommand{\Def}[0]{\mathrel{\mathop:}=}
\icmltitlerunning{Min-Max Optimization without Gradients for ICML 2020}

\begin{document}

\twocolumn[
\icmltitle{Min-Max Optimization without Gradients: 
 Convergence    and Applications to  Black-Box   Evasion and Poisoning  Attacks}

\icmlsetsymbol{equal}{*}

\begin{icmlauthorlist}
\icmlauthor{Sijia Liu}{equal,mit-ibm}
\icmlauthor{Songtao Lu}{equal,tj}
\icmlauthor{Xiangyi Chen}{equal,umn}
\icmlauthor{Yao Feng}{equal,tsinghua}
\icmlauthor{Kaidi Xu}{equal,neu}
\icmlauthor{Abdullah Al-Dujaili}{equal,mit,mit-ibm}
\icmlauthor{Mingyi Hong}{umn}
\icmlauthor{Una-May O'Reilly}{mit,mit-ibm}
\end{icmlauthorlist}

\icmlaffiliation{mit-ibm}{MIT-IBM Watson AI Lab, IBM Research}
\icmlaffiliation{tj}{Thomas J. Watson Research Center, IBM Research}
\icmlaffiliation{umn}{ECE, University of Minnesota}
\icmlaffiliation{tsinghua}{Tsinghua University}
\icmlaffiliation{neu}{ECE, Northeastern University}
\icmlaffiliation{mit}{CSAIL, MIT}
\icmlcorrespondingauthor{Sijia Liu}{Sijia.Liu@ibm.com}

\icmlkeywords{Machine Learning, ICML}

\vskip 0.3in
]

\printAffiliationsAndNotice{\icmlEqualContribution} 

\begin{abstract}
In this paper, we study the problem of constrained  min-max optimization {in a black-box setting}, where  
the desired optimizer cannot access the gradients of the objective function but may query its values.  We present a principled optimization framework,  integrating a  
zeroth-order (ZO) gradient estimator with 
an alternating projected stochastic gradient descent-ascent method, where the former only requires a small number of function queries and the later   needs  just one-step descent/ascent update.  
We show that the proposed framework,  referred to as \textit{ZO-Min-Max}, has a sub-linear convergence rate under mild conditions and scales  gracefully with problem size.
We also
explore a promising connection between black-box min-max optimization and  black-box evasion and poisoning attacks in adversarial machine learning (ML). Our empirical evaluations on these use cases demonstrate the effectiveness of our approach and its scalability to dimensions that prohibit using 
recent  black-box solvers. 
\end{abstract}

\section{Introduction}
Min-max optimization problems 
have been studied for multiple decades~\citep{wald1945statistical}, and the majority of the proposed methods assume access to first-order (FO) information, i.e. gradients, to find or approximate robust solutions \citep{nesterov2007dual,gidel17a,yazdandoost2018iteration,qian2019robust, rafique2018non,sanjabi2018convergence, luts19, nouiehed2019solving,lu2019hybrid,jin2019minmax}.
Different from \emph{standard} optimization, 
min-max optimization  tackles a composition of an inner maximization problem and an outer minimization problem.
It can be used in
   many real-world applications, which are faced with  various forms of uncertainty or  adversary. For instance, when training a ML model on user-provided data, malicious users can carry out a {data poisoning attack}: providing false data with the aim of
corrupting the learned model \citep{steinhardt2017certified,tran2018spectral,jagielski2018manipulating}. At inference time, malicious users can evade   detection of multiple models  in the form of {adversarial example attacks} \citep{goodfellow2014explaining,liu2016delving, liu2018caad}.
{Our study is particularly motivated  by the design of   \emph{data poisoning} and \emph{evasion} adversarial attacks from \textit{black-box} machine learning (ML) or deep learning (DL) systems, whose    internal configuration  and operating mechanism are unknown to adversaries.}
We propose
\textit{zeroth-order (gradient-free)} min-max optimization methods, where gradients are neither symbolically nor numerically available, or they are tedious to compute.

Recently, \textit{zeroth-order (ZO) optimization} has attracted increasing  attention in solving ML/DL problems, where
FO  gradients (or stochastic gradients) are approximated   based  only on the function
values.
For example, ZO optimization serves as a powerful and practical tool for generation of      adversarial  example   to evaluate the adversarial robustness of black-box  ML/DL models \citep{chen2017zoo,ilyas2018black,tu2018autozoom,ilyas2018prior,chen2018frank,li2019nattack}. 
 ZO optimization can also help to solve automated ML problems, where the gradients with respect to ML pipeline configuration parameters are intractable \citep{aggarwal2019can,wang2019flo}.
Furthermore,
ZO optimization provides computationally-efficient alternatives of high-order optimization methods for solving complex ML/DL tasks, e.g., robust training by leveraging input gradient or curvature regularization \citep{finlay2019scaleable,moosavi2019robustness},  network control and management \citep{chen2017bandit,liu2017zeroth}, and  data processing in high dimension \citep{liu2017zeroth,golovin2019gradientless}. Other recent applications include generating model-agnostic contrastive explanations \citep{dhurandhar2019model} and escaping saddle points \citep{flokas2019efficiently}.

Current
studies \citep{ghadimi2013stochastic,nesterov2015random,duchi2015optimal,ghadimi2016mini,shamir2017optimal,Balasubramanian2018nips,liu2018_NIPS,liu2018signsgd} suggested that ZO methods for solving \textit{single-objective} optimization problems typically agree with the iteration complexity of FO
methods but encounter a slowdown factor up to a small-degree polynomial of the problem dimensionality.
To the best of our knowledge,    it was an open question whether any convergence rate analysis   can be established for {black-box min-max} (bi-level) optimization. \textcolor{black}{In this paper, we develop a  
  \emph{provable} and \emph{scalable} 
black-box
min-max stochastic optimization method by integrating a \emph{query-efficient} randomized ZO gradient estimator with a \emph{computation-efficient} alternating  gradient descent-ascent framework. Here the former  requires a small number of function queries, and the latter  
needs just one-step descent/ascent update.}

\paragraph{Contribution.} 
We summarize our  contributions as follows.
\textbf{(\emph{i})} We  identify a class of black-box attack problems which turn out to be min-max black-box optimization problems. 
\textbf{(\emph{ii})} We propose
 a scalable and   principled framework  (ZO-Min-Max) for solving constrained min-max saddle point problems under both one-sided and two-sided black-box objective functions. Here 
the one-sided   setting refers to the scenario where only the outer minimization problem is black-box.
\textbf{(\emph{iii})} We provide a novel  convergence analysis characterizing 
the number of objective function evaluations required
to attain locally robust solution to   black-box min-max problems (structured by nonconvex outer minimization and strongly concave inner maximization).
Our analysis    handles stochasticity  in both objective function and ZO gradient estimator, 
and shows that ZO-Min-Max  yields $\mathcal{O}(1/T + 1/b + d/q)$ convergence rate,  
where 
$T$ is number of iterations, $b$ is mini-batch size, 
 $q$ is number of random direction vectors used in ZO gradient estimation, and $d$ is number of optimization variables.
\textbf{(\emph{iv})} We demonstrate 
the effectiveness  of our proposal in practical data poisoning and evasion attack generation problems.\footnote{Source code will be released.}
\section{Related Work}
\paragraph{FO min-max optimization.}
Gradient-based methods have been applied with celebrated success to solve min-max problems such as robust learning \citep{qian2019robust}, generative adversarial networks (GANs) \citep{Sanjabi18}, adversarial training \citep{al2018adversarial,madry2017towards}, and robust adversarial attack generation \citep{wang2019unified}.
Some of FO methods are motivated by theoretical justifications based on Danskin's theorem~\citep{danskin1966theory}, which implies that the negative of the gradient of the outer minimization problem at inner maximizer is a descent direction~\citep{madry2017towards}. Convergence analysis of other FO min-max methods has been studied under  different problem settings, e.g., \citep{lu2019hybrid,qian2019robust,rafique2018non,sanjabi2018convergence,nouiehed2019solving}.
It was shown in \citep{lu2019hybrid} that a deterministic FO min-max algorithm has $\mathcal{O}(1/T)$ convergence rate. In  \citep{qian2019robust,rafique2018non},
stochastic FO min-max methods have also been  proposed, which yield the convergence rate  in the order of $\mathcal O ( 1/\sqrt{T})$ and  $\mathcal O (1/T^{1/4})$, respectively. However, these works were restricted to unconstrained optimization at the minimization side.
In \citep{sanjabi2018convergence},     noncovnex-concave min-max problems were studied, but 
the proposed analysis requires solving  the maximization problem only up to some small error.
In \citep{nouiehed2019solving}, the $\mathcal O(1/T)$ convergence rate was proved for nonconvex-nonconcave min-max problems under  Polyak- {\L}ojasiewicz conditions. Different from the aforementioned FO settings,  ZO min-max stochastic optimization   suffers  randomness from both stochastic sampling in objective function and ZO gradient estimation, and this randomness   would be    coupled in alternating gradient descent-descent steps and thus make it more challenging in convergence analysis.

\paragraph{Gradient-free min-max optimization.}
In the black-box setup, {coevolutionary algorithms} were used extensively to solve min-max problems~\citep{herrmann1999genetic,schmiedlechner2018towards}. 
However, they may oscillate and never converge to a solution due to pathological behaviors such as \emph{focusing} and \emph{relativism}~\citep{watson2001coevolutionary}. Fixes to these issues have been proposed and analyzed---e.g., {asymmetric fitness}~\citep{jensen2003new, branke2008new}. In \citep{al2018application}, the authors employed an evolution strategy as an unbiased approximate for the descent direction of the outer minimization problem and showed empirical gains over coevlutionary techniques, albeit without any theoretical guarantees. 
{Min-max black-box problems can also be addressed by resorting to 
direct search and  model-based descent  and  trust region methods \citep{audet2017derivative,larson2019derivative,rios2013derivative}. However, these methods lack convergence rate analysis and are difficult to scale to high-dimensional problems. For example, the   off-the-shelf model-based solver  \textsf{COBYLA} only supports problems with  $2^{16}$ variables at maximum in SciPy  Python library \citep{scipy2001}, which is even smaller than the size of a single ImageNet image.
}

The recent work \citep{bogunovic2018adversarially} proposed a robust  Bayesian optimization (BO) algorithm and established a theoretical lower bound on the required number of the min-max objective evaluations to find a near-optimal point. 
However, BO approaches are often tailored to low-dimensional problems and its computational complexity prohibits scalable application. From a game theory perspective, the min-max solution for some problems correspond to the Nash equilibrium between the outer minimizer and the inner maximizer, and hence black-box Nash equilibria solvers can be used~\citep{picheny2019bayesian,al2018approximating}. This setup, however, does not always hold in general. Our work contrasts with the above lines of work in designing and analyzing black-box min-max techniques that are \emph{both} scalable and theoretically grounded.

\section{Problem Setup}
In this section, we define the black-box min-max problem and briefly motivate its applications. By \emph{min-max}, we mean that the problem is a composition of  inner maximization and outer minimization of the objective function $f$. By \emph{black-box}, we mean that the \textit{objective function} $f$ is only accessible via {functional} evaluations. Mathematically, we have
{\small
\begin{align}\label{eq: prob}
    \begin{array}{ll}
\displaystyle \min_{\mathbf x \in \mathcal X} \max_{\mathbf y \in \mathcal Y}         &  f(\mathbf x, \mathbf y)
    \end{array}
\end{align}}%
where $\mathbf x$ and $\mathbf y$ are optimization variables, $f$ is a differentiable objective function, 
and  $\mathcal X \subset \mathbb{R}^{d_x}$ and $\mathcal Y \subset \mathbb{R}^{d_y}$ are 
compact  
convex sets.  \textcolor{black}{For ease of notation,  let $d_x = d_y = d$.}
In \eqref{eq: prob}, the objective function $f$ could represent either a deterministic loss or stochastic loss  $f(\mathbf x, \mathbf y) = \mathbb E_{\boldsymbol{\xi}\sim p} \left [f(\mathbf x, \mathbf y; \boldsymbol{\xi}) \right ]$, where $\boldsymbol{\xi}$   is a random variable following the distribution $p$. In this paper, we cover the   stochastic variant in \eqref{eq: prob}.
 We focus on two \textit{black-box} scenarios. 
    \textit{\textbf{(a)} One-sided black-box}: $f$ is   a white box w.r.t. $\mathbf y$ but a black box w.r.t. $\mathbf x$.
 \textit{\textbf{(b)} Two-sided black-box}: $f$ is    a black box w.r.t. both $\mathbf x$ and $\mathbf y$.

\paragraph{Motivation of setup (a) and (b).}

The formulation of the \textit{one-sided black-box} min-max problem can be used to design the
\textit{black-box ensemble evasion attack},  
where the attacker 
generates   adversarial  examples (i.e., crafted examples with slight perturbations for misclassification at the \textit{testing} phase) and optimizes its worst-case performance against an \textit{ensemble}
of  black-box classifiers and/or example classes.
The formulation of \textit{two-sided black-box} min-max problem represents another type of attack at the \textit{training} phase, known as \textit{poisoning attack}, where the attacker deliberately influences the training data (by injecting  poisoned samples) to manipulate the results of a black-box predictive model. 
{Although 
problems of designing ensemble evasion attack \citep{liu2016delving,liu2018caad,wang2019unified} and 
data poisoning attack \citep{jagielski2018manipulating,wang2019neural} have been studied in the literature, most of them assumed that the adversary has the \textit{full} knowledge of the target ML model, leading to an impractical \textit{white-box} attack setting. By contrast, we provide  a solution to min-max attack generation   under   \textit{black-box} ML models.
We refer readers to Section~\ref{sec: exp} for further discussion and demonstration of our framework on these problems.
}

\section{ZO-Min-Max: A Framework for Black-Box Min-Max Optimization}
Our interest  is in a scalable and theoretically principled framework for solving black-box min-max problems of the form \eqref{eq: prob}. To this end, 
we first introduce a randomized  gradient estimator  that only requires a few number of point-wise  function evaluations.
Based on that, we then propose a ZO
alternating projected  gradient method to solve~\eqref{eq: prob} under both one-sided and two-sided black-box setups.

\paragraph{Randomized gradient estimator.}
\textcolor{black}{In the ZO setting, we adopt
a randomized gradient estimator to estimate the gradient
of a function with the \textit{generic} form $h(\mathbf x) \Def \mathbb E_{\boldsymbol{\xi}} [ h(\mathbf x; \boldsymbol{\xi}) ] $ \citep{gao2014information,berahas2019theoretical}, 
{\small \begin{align}\label{eq: grad_rand_ave}
    \widehat{\nabla}_{\mathbf x}h  (\mathbf x ) =
    {   \frac{1}{bq  } \sum_{j \in \mathcal I} \sum_{i=1}^q  \frac{d [h ( \mathbf x + \mu \mathbf u_i; \boldsymbol{\xi}_j ) - h ( \mathbf x  ; \boldsymbol{\xi}_j) ] }{\mu} }  \mathbf u_i,
\end{align}}%
where  
$d$ is number of  variables,
$\mathcal I$ denotes the mini-batch set of $b$ \emph{i.i.d.} stochastic samples $ \{ \boldsymbol{\xi}_j \}_{j=1}^b$, 
$\{ \mathbf u_i \}_{i=1}^q$ are $q$ \emph{i.i.d.} random direction vectors drawn uniformly from the unit sphere, and $\mu > 0$ is a  smoothing parameter.  We note that the ZO gradient estimator \eqref{eq: grad_rand_ave} involves randomness from both stochastic sampling w.r.t. $\mathbf u_i$ as well as the random direction sampling w.r.t. $\boldsymbol{\xi}_j$.
It is known from \citep[Lemma\,2]{gao2014information} that    $\widehat{\nabla}_{\mathbf x}h  (\mathbf x ) $ provides an unbiased estimate of the gradient of the smoothing function of $h$ rather than 
the true gradient of $h$. Here   the smoothing function  of $h$ is defined by
$h_{\mu}(\mathbf x) = \mathbb E_{\mathbf v} [ h(\mathbf x + \mu \mathbf v) ]$,  where
 $\mathbf v$ follows the uniform distribution over the unit Euclidean ball. Besides the bias, we provide an upper bound on the variance of    \eqref{eq: grad_rand_ave} in Lemma\,\ref{lemma: smooth_f_random_stochastic}.
}

\begin{lemma}
\label{lemma: smooth_f_random_stochastic}
\textcolor{black}{Suppose that for all $\boldsymbol{\xi}$, $h(\mathbf x; \boldsymbol{\xi})$ has $L_h$ Lipschitz continuous gradients and the gradient of $h(\mathbf x; \boldsymbol{\xi})$ is upper bounded as $\| \nabla_{\mathbf x}  h(\mathbf x; \boldsymbol{\xi}) \|_2^2 \leq \eta^2 $ } at     $\mathbf x \in \mathbb R^d$. Then $\mathbb E \left [  \widehat{\nabla}_{\mathbf x}h  (\mathbf x ) \right ] = \nabla_{\mathbf x} h_\mu (\mathbf x)$, and  
{\small 
\begin{align}
& \mathbb E \left [
\| 
 \widehat{\nabla}_{\mathbf x}h  (\mathbf x ) - \nabla_{\mathbf x} h_\mu (\mathbf x)
\|_2^2
\right ]  \leq 
\sigma^2(L_h,\mu, b, q, d),
\label{eq: second_moment_grad_random}
\end{align}%
}where the expectation is taken over all randomness, and $\sigma^2(L_h,\mu, b, q, d) = \frac{2 \eta^2}{b} + \frac{4d \eta^2 + \mu^2 L_h^2 d^2}{q} $. 
\end{lemma}
\textbf{Proof:} See Appendix\,\ref{app: proof_variance}. 
\hfill $\square$

In Lemma\,\ref{lemma: smooth_f_random_stochastic}, if we choose $\mu \leq 1/\sqrt{d}$, then the variance bound is given by $\mathcal O(1/b + d/q)$.
In our problem setting \eqref{eq: prob}, the ZO gradients $\widehat \nabla_{\mathbf x} f(\mathbf x, \mathbf y) $ and $\widehat \nabla_{\mathbf y} f (\mathbf x, \mathbf y) $  follow the   generic form of \eqref{eq: grad_rand_ave} by fixing $\mathbf y$ and letting $h(\cdot) \Def f(\cdot, \mathbf y)$ or by fixing $\mathbf x$ and letting $h(\cdot) \Def f(\mathbf x, \cdot )$, respectively. 

\paragraph{Algorithmic framework.}
To solve problem \eqref{eq: prob}, we alternatingly perform  ZO projected gradient descent/ascent method for updating $\mathbf x$ and $\mathbf y$.  
Specifically, 
the ZO projected gradient descent (ZO-PGD) is conducted over $\mathbf x$ 
{\small \begin{align}\label{eq: pgd_out_min}
    \mathbf {x}^{(t)} = \mathrm{proj}_{\mathcal X} \left  ( \mathbf {x}^{(t-1)}- \alpha  \widehat {\nabla}_{\mathbf x} f \left (\mathbf {x}^{(t-1)},\mathbf y^{(t-1)} \right ) \right ),  
\end{align}}%
where $t$ is the iteration index, $\widehat{\nabla}_{\mathbf x} f$ denotes the ZO gradient estimate of $f$ w.r.t. $\mathbf x$, $\alpha > 0$ is the learning rate at the $\mathbf x$-minimization step, and 
$\mathrm{proj}_{\mathcal X} (\mathbf a)$ signifies the projection of $\mathbf a$ onto $\mathcal X$,   given by the solution to the   problem $\min_{\mathbf x \in \mathcal X}  \| \mathbf x - \mathbf a \|_2^2$. For one-sided ZO min-max optimization, besides \eqref{eq: pgd_out_min}, we perform FO projected gradient ascent (PGA) over $\mathbf y$. And 
for two-sided ZO min-max optimization,   our update on $\mathbf y$ obeys  ZO-PGA
{\small \begin{align}\label{eq: pga_in_max}
    \mathbf {y}^{(t)} = \mathrm{proj}_{\mathcal Y} \left  ( \mathbf {y}^{(t-1)} + \beta  \widehat {\nabla}_{\mathbf y} f \left (\mathbf {x}^{(t)},\mathbf y^{(t-1)} \right ) \right ),  
\end{align}}%
where $\beta > 0$ is the learning rate at the $\mathbf   y$-maximization step, and 
$\widehat{\nabla}_{\mathbf y} f$ denotes the ZO gradient estimate of $f$ w.r.t. $\mathbf y$. The proposed  method is named as \textit{ZO-Min-Max}; see the pseudocode in Appendix\,\ref{app: alg_code}.

\paragraph{Why estimates gradient rather than  distribution of function values?}
Besides ZO optimization using random gradient estimates, the black-box min-max  problem \eqref{eq: prob} can also be solved using the Bayesian optimization (BO) approach, e.g., \citep{bogunovic2018adversarially,al2018approximating}. The core idea of BO is to approximate the objective function as a Gaussian process (GP) learnt from the history of function values at queried points. Based on GP, the solution to  problem \eqref{eq: prob} is then updated by maximizing  a certain reward function, known as acquisition function. The advantage of BO is its mild requirements on the setting of black-box problems, e.g., at the absence of differentiability. However, 
BO
usually does not scale beyond low-dimensional problems
since learning the accurate GP model and solving the acquisition problem takes intensive  computation cost per iteration. 
By contrast, our proposed method is more  efficient, and mimics the first-order method by just using the  random gradient estimate \eqref{eq: grad_rand_ave}  
as the descent/ascent direction. In Figure\,\ref{fig: ZOMinMax_BO}, we compare ZO-Min-Max with the BO based  STABLEOPT algorithm proposed by \citep{bogunovic2018adversarially} through a toy    example shown in Appendix\,\ref{app: toy} and a  poisoning attack generation example in Sec.\,\ref{sec: exp_poison}. 
We can see that ZO-Min-Max not only achieves more accurate solution
but also requires less computation time.

\paragraph{Why is difficult to analyze the convergence of ZO-Min-Max?}
The convergence analysis of ZO-Min-Max is more challenging than the case of FO min-max algorithms. 
The stochasticity of the gradient estimator makes the convergence analysis sufficiently different from the \textcolor{black}{FO} deterministic case \citep{lu2019hybrid,qian2019robust}, since the errors in minimization and maximization are coupled as the algorithm proceeds. 

{Moreover, the conventioanl analysis 
of ZO optimization for single-objective problems cannot directly be applied to ZO-Min-Max. Even at the one-sided black-box setting, ZO-Min-Max conducts alternating optimization using one-step ZO-PGD and PGA with respect to $\mathbf x$ and $\mathbf y$ respectively. This is different from a reduced ZO optimization problem with respect to $\mathbf x$ only  by solving  problem $\min_{\mathbf x \in \mathcal X} h(\mathbf x)$,
where $h(\mathbf x) \Def \max_{\mathbf y \in \mathcal Y} f( \mathbf x, \mathbf y)$. Here acquiring the solution to the inner maximization   problem   
could be
non-trivial and computationally intensive.
Furthermore,   the alternating algorithmic structure  leads to  opposite optimization directions (minimization vs maximization) over variables $\mathbf x$ and $\mathbf y$ respectively.  Even applying ZO optimization only to one side, it needs to quantify the effect of ZO gradient estimation on the descent over both $\mathbf x$ and $\mathbf y$. We provide a detailed convergence analysis  of ZO-Min-Max in Sec.\,\ref{sec: conv}.
}

\section{Convergence Rate Analysis}\label{sec: conv}
We first elaborate on assumptions and notations used in analyzing the convergence of ZO-Min-Max (Algorithm \ref{alg: ZO_2side}). 

\textbf{A1}: In \eqref{eq: prob}, $f(\mathbf x, \mathbf y)$ is continuously differentiable, and is strongly concave w.r.t. $\mathbf y$ with parameter {$\gamma > 0$, namely, given $\mathbf x \in \mathcal X$, $ f(\mathbf x, \mathbf y_1) \leq  f(\mathbf x, \mathbf y_2) + \nabla_{\mathbf y} f(\mathbf x , \mathbf y_2)^T (\mathbf y_1 - \mathbf y_2) - \frac{\gamma}{2} \| \mathbf y_1 - \mathbf y_2 \|^2 $ for all points $\mathbf y_1, \mathbf y_2 \in \mathcal Y$.  And $f$   is lower bounded  {by a finite number $f^*$}
and  has bounded gradients $\| \nabla_{\mathbf x} f(\mathbf x, \mathbf y; \boldsymbol{\xi }) \| \leq \eta^2$ and $\| \nabla_{\mathbf y} f(\mathbf x, \mathbf y; \boldsymbol{\xi }) \| \leq \eta^2$  for stochastic optimization with $\boldsymbol{\xi} \sim p$.} Here $\| \cdot \|$ denotes the $\ell_2$ norm. The constraint sets $\mathcal{X},\mathcal{Y}$ are  convex and bounded with diameter $R$.

{\textbf{A2}: $f(\bx,\by)$  has  Lipschitz continuous gradients, i.e., there exists $L_x, L_y > 0$ such that    $\|\nabla_{\bx}f(\bx_1,\by)-\nabla_{\bx}f(\bx_2,\by)\|\le L_x\|\bx_1-\bx_2\|$ for $\forall\bx_1,\bx_2\in\mathcal{X}$, and 
$\|\nabla_{\by} f(\bx_1,\by)-\nabla_{\by}f(\bx_2,\by)\|\le L_{y}\|\bx_1-\bx_2\|$ and $\|\nabla_{\by} f(\bx,\by_1)-\nabla_{\by}f(\bx,\by_2)\|\le L_{y}\|\by_1-\by_2\|$ for $\forall\by_1,\by_2\in\mathcal{Y}$. 
}

We remark that \textbf{A1} and \textbf{A2}   are required for analyzing the convergence of ZO-Min-Max. They were used even for the analysis of 
first-order  optimization methods  with  single rather than bi-level objective functions \citep{chen2018convergence,pmlr-v97-ward19a}. In \textbf{A1}, the strongly concavity of $f(\mathbf x, \mathbf y)$ with respect to $\mathbf y$ holds for applications such as robust learning over multiple domains \citep{qian2019robust}, and adversarial attack generation shown in Section\,\ref{sec: exp}. In \textbf{A2},  the assumption of smoothness (namely, Lipschitz continuous gradient) is required to quantify the descent of the  alternating projected stochastic gradient descent-ascent method.
For clarity, we also summarize  the problem and algorithmic parameters used in our convergence analysis in
 Table\,\ref{table: parameter} of Appendix\,\ref{app: table_para}.

\textcolor{black}{We measure the convergence of ZO-Min-Max by
 the proximal gradient
\citep{lu2019hybrid,ghadimi2016mini}, 
{\small\begin{align}\label{eq.optgap}
    \mathcal{G}(\bx,\by)=\left[\begin{array}{c}
         (1/\alpha)\left ( \bx-\textrm{proj}_{\mathcal{X}}(\bx-\alpha\nabla_{\bx} f(\bx,\by))\right ) \\
        (1/\beta)\left ( \by-\textrm{proj}_{\mathcal{Y}}(\by+\beta\nabla_{\by} f(\bx,\by))\right )
    \end{array}\right],
\end{align}}%
where $(\bx,\by)$ is a first-order stationary point of \eqref{eq: prob} iff $\|\mathcal{G}(\bx,\by)\|=0$.
}

\paragraph{Descent property in $\mathbf x$-minimization.}
{In what follows, we delve into our convergence analysis.
Since ZO-Min-Max (Algorithm \ref{alg: ZO_2side}) calls for ZO-PGD for solving both  one-sided and two-sided black-box optimization problems, we first show   the descent property   at the $\mathbf x$-minimization step of ZO-Min-Max in   Lemma\,\ref{le.common}.
}

\begin{lemma}\label{le.common}
(Descent lemma in minimization) Under \textbf{A1}-\textbf{A2}, let $(\bx^{(t)},\by^{(t)})$ be a sequence generated by ZO-Min-Max. When $f(\bx,\by)$ is
black-box {w.r.t. $\mathbf x$}, 
then we have following descent property w.r.t. $\mathbf x$:
{\small \begin{align}\label{eq: inequality_lemma}
\mathbb{E}[ {f}(\bx^{(t+1)}, \by^{(t)}) ]
\leq &  \mathbb {E}[{f}(\bx^{(t)}, \by^{(t)})] 
- \left(\frac{1}{\alpha}-\frac{L_x}{2}\right) \mathbb{E}\|\Delta^{(t+1)}_{\bx}\|^2\nonumber \\
&+ \alpha \sigma_x^2 +L_x\mu^2
\end{align}}%
where $\Delta^{(t)}_{\bx}\Def\bx^{(t)}-\bx^{(t-1)}$,  and \textcolor{black}{$\sigma_x^2 \Def \sigma^2(L_x, \mu,  b, q, d)$ defined in  \eqref{eq: second_moment_grad_random}}.
\end{lemma}
{\textbf{Proof}: See Appendix\,\ref{app: zomin}. \hfill $\square$}
\textcolor{black}{It is clear from
\leref{le.common}  that updating $\bx$ leads to the reduced  objective value when choosing a small learning rate $\alpha$. However, ZO  gradient estimation brings in additional errors in terms of   $\alpha \sigma_x^2$ and $L_x \mu^2$, where the former is induced by the variance of gradient estimates in \eqref{eq: second_moment_grad_random} and the latter is originated from bounding the distance between $f$ and its smoothing version; see \eqref{eq.xuppersmo}  in Appendix\,\ref{app: ZOMinMaxPGA}.}

\paragraph{Convergence rate of ZO-Min-Max by performing PGA.}
We next investigate the convergence   of ZO-Min-Max when  PGA is
used at the $\mathbf y$-maximization step (Line 8 of Algorithm\,\ref{alg: ZO_2side}) for solving one-sided black-box optimization problems. 

\begin{lemma}\label{le.le2}
(Descent lemma in maximization) Under \textbf{A1}-\textbf{A2}, let $(\bx^{(t)},\by^{(t)})$ be a sequence generated by Algorithm \ref{alg: ZO_2side} and define the potential function  as
{\small\begin{align}\label{eq: potential_1side}
\mathcal{P}(\bx^{(t)},\by^{(t)},\Delta^{(t)}_{\by})=
& \mathbb{E}[f(\bx^{(t)},\by^{(t)})] \nonumber \\
& +
\frac{4 + 4 \beta^2 L_y^2 - 7 \beta \gamma }{2\beta^2 \gamma}
\mathbb{E}\|\Delta^{(t)}_{\by}\|^2,
\end{align}}%
where $\Delta^{(t)}_{\by}\Def\by^{(t)}-\by^{(t-1)}$.  
When $f(\bx,\by)$ is   {black-box w.r.t. $\mathbf x$ and}  white-box w.r.t. $\mathbf y$, then we have the following descent property w.r.t. $\by$:
{\small \begin{align}
& \mathcal{P}(\bx^{(t+1)},\by^{(t+1)},\Delta^{(t+1)}_{\by})
\le \mathcal{P}(\bx^{(t+1)},\by^{(t)},\Delta^{(t)}_{\by})
\nonumber \\
& -\left(\frac{1}{2\beta}-\frac{2L^2_y}{\gamma}\right) \mathbb{E}\|\Delta^{(t+1)}_{\by}\|^2+\left(\frac{2}{\gamma^2\beta}+\frac{\beta}{2}\right)  L^2_x \mathbb{E}\|\Delta^{(t+1)}_{\bx}\|^2.
\label{eq.edesofy}
\end{align}}%
\end{lemma}
\textbf{Proof}: See Appendix\,\ref{sec:zofomax}. \hfill $\square$

It is shown from  \eqref{eq.edesofy} that when $\beta$ is small enough,   then the
term $(1/(2\beta)-2L^2_y/\gamma)\mathbb{E}\|\Delta^{(t+1)}_{\by}\|^2$ will give some descent of  the potential function  after  PGA, while the last term in \eqref{eq.edesofy} will give some ascent to the potential function. However, such a   quantity will be compensated by the descent of the objective function in the minimization step shown by \leref{le.common}.
Combining  \leref{le.common} and Lemma\,\ref{le.le2}, we obtain the   convergence rate of ZO-Min-Max in Theorem\,\ref{th.main1}.

\begin{theorem}\label{th.main1}
Suppose that \textbf{A1}-\textbf{A2} hold, the sequence $(\bx^{(t)},\by^{(t)})$ over $T$ iterations is generated by  Algorithm \ref{alg: ZO_2side} in which learning rates satisfy $\beta<1/(4L^2_y)$ and  $\alpha\le\min\{1/L_x,1/(L_x/2+2L^2_x/(\gamma^2\beta)+\beta L^2_x/2)\}$. 
\textcolor{black}{When $f(\bx,\by)$ is {black-box w.r.t. $\mathbf x$ and}  white-box w.r.t. $\mathbf y$,  the convergence rate of ZO-Min-Max under a uniformly and randomly picked $(\mathbf x^{(r)}, \mathbf y^{(r)})$ from  $\{ (\bx^{(t)},\by^{(t)}) \}_{t=1}^T$ is given by}
{\small \begin{equation}\label{eq: thr1_rate}
 \mathbb{E}\|\mathcal{G}(\bx^{(r)},\by^{(r)})\|^2
\le\frac{{c}}{\zeta}\frac{ (
\mathcal{P}_1 -
\textcolor{black}{f^* - \nu R^2 }
 )}{T}+ \frac{{c \alpha}\sigma^2_x}{\zeta}   + \frac{{c}L_x\mu^2}{\zeta} 
\end{equation} }%
where \textcolor{black}{$\zeta$  is a constant independent on the parameters $\mu$, $b$, $q$, $d$ and $T$,}
{$\mathcal P_t \Def  \mathcal{P}(\bx^{(t)},\by^{(t)},\Delta^{(t)}_{\by})$ given by \eqref{eq: potential_1side},
$c= \max\{L_x+3/\alpha,3/\beta\}$, \textcolor{black}{$\nu = \min \{  4 + 4 \beta^2 L_y^2 - 7 \beta \gamma , 0\}  /(2\beta^2 \gamma)$},  
  $\sigma_x^2$ is variance bound of ZO gradient estimate given in    \eqref{eq: inequality_lemma}, and $f^*$, $R$, $\gamma$, $L_x$ and $L_y$ have been defined in \textbf{A1}-\textbf{A2}.
  }

\end{theorem}
\textbf{Proof}: See Appendix\,\ref{sec:th1}. \hfill $\square$

{
To better interpret  Theorem\,\ref{th.main1},
we begin by clarifying the parameters involved in our convergence rate \eqref{eq: thr1_rate}. 
First,  the parameter $\zeta$  in the denominator of the derived convergence error  is      non-trivially lower bounded given appropriate   learning rates $\alpha$ and $\beta$ (as will be evident in \textbf{Remark\,1}). 
Second, the parameter  $c$ is inversely proportional to   $\alpha$ and $\beta$. Thus,  to  guarantee the constant effect of the ratio $c/\xi$, it is better not to set these learning rates too small; see a specification in \textbf{Remark\,1-2}.
Third, the parameter $\nu$ is non-negative and appears in terms of $-\nu R^2$, thus, it will not make convergence rate worse. 
Fourth, $\mathcal P_1$ is the initial value of the potential function \eqref{eq: potential_1side}. By setting an appropriate learning rate $\beta$ (e.g., following \textbf{Remark\,2}),  $\mathcal P_1$ is then upper bounded by a constant determined by the initial value of the objective function, the distance of the first two updates, Lipschitz constant $L_y$ and strongly concave parameter $\gamma$.
We next provide \textbf{Remarks\,1-3} on Theorem\,\ref{th.main1}. 
}

{
\textbf{Remark\,1.} Recall that  $\zeta = \min \{ c_1, c_2\}$ (Appendix\,B.2.3), where $c_1 = {1}/{(2\beta)}-{2L^2_y}/{\gamma}$ and $c_2 = \frac{1}{\alpha}-(\frac{L_x}{2}+\frac{2L^2_x}{\gamma^2\beta}+\frac{\beta L^2_x}{2})$.  Given the fact that $L_x$ and $L_y$ are Lipschitz  constants and $\gamma$ is the strongly concavity constant,
a proper lower bound of  $\zeta$ thus relies on the choice of the learning rates $\alpha$ and $\beta$.
By setting $\beta \leq \frac{\gamma}{8 L_y^2}$ and $ \alpha \leq 1/({L_x}+\frac{4L^2_x}{\gamma^2\beta}+ {\beta L^2_x})$, it is easy to verify that $c_1 \geq {\frac{2L_y^2}{\gamma}}
$ and $c_2 \geq \frac{L_x}{2}+\frac{2L^2_x}{\gamma^2\beta}+\frac{\beta L^2_x}{2} \geq \textcolor{black}{\frac{L_x}{2} + \frac{2L_x^2}{\gamma}}
$. Thus, we obtain that   $\zeta \geq \min \{  \frac{2L_y^2}{\gamma} , \frac{2 L_x^2}{\gamma} + \frac{L_x}{2}\}$.
This  justifies that  $\zeta$ has a non-trivial lower bound, which    will not make the convergence error  bound    \eqref{eq: thr1_rate} vacuous (although the bound has  not been optimized over $\alpha$ and $\beta$). }

{
\textbf{Remark\,2.} It is not wise to set   learning rates $\alpha$ and $\beta$ to  extremely small values  
 since $c$ is \textit{inversely} 
 proportional to $\alpha$ and $\beta$. We could choose $\beta = \frac{\gamma}{8 L_y^2}$ and $ \alpha = 1/({L_x}+\frac{4L^2_x}{\gamma^2\beta}+ {\beta L^2_x})$ in  Remark\,1 to guarantee the constant effect of $c/\zeta$. }
 
{
\textbf{Remark\,3.} By setting $\mu \leq \min \{ 1/\sqrt{d}, 1/\sqrt{T}\}$, we obtain   $\sigma_x^2 = \mathcal O(1/b + d/q)$ from  Lemma\,\ref{lemma: smooth_f_random_stochastic}, and Theorem\,\ref{th.main1} implies that 
ZO-Min-Max yields $ \mathcal O(1/T + 1/b + d/q   ) $ convergence rate for     one-sided black-box optimization. Compared to the FO rate $\mathcal O(1/T)$ \citep{lu2019hybrid,Sanjabi18}, ZO-Min-Max   converges only to
a neighborhood of stationary points  with $\mathcal O(1/T)$ rate, where the size of the neighborhood is determined by the mini-batch size $b$ and the number of random direction vectors $q$ used in ZO gradient estimation.  It is also worth mentioning that such a stationary gap may exist even in the FO/ZO projected stochastic gradient descent for solving single-objective minimization problems \citep{ghadimi2016mini}.
 }

{
As shown in \textbf{Remark\,3}, ZO-Min-Max could result in a stationary gap. A large   mini-batch size $b$ or   number of random direction vectors $q$ can improve its iteration complexity. However, this requires $O(bq)$ times more function queries per iteration from  \eqref{eq: grad_rand_ave}. It implies   the tradeoff between iteration complexity and function query complexity in ZO optimization.
}

\paragraph{Convergence rate of ZO-Min-Max by performing ZO-PGA.}
The previous convergence analysis of ZO-Min-Max 
is served as a basis for analyzing the more general two-sided black-box case, where  ZO-PGA is
used at  the $\mathbf y$-maximization step (Line 6 of Algorithm\,\ref{alg: ZO_2side}). In Lemma\,\ref{le.le3}, we examine  the   descent property in maximization by using   ZO gradient estimation.

\begin{lemma}\label{le.le3}
(Descent lemma in maximization) Under \textbf{A1}-\textbf{A2}, let $(\bx^{(t)},\by^{(t)})$ be a sequence generated by Algorithm \ref{alg: ZO_2side} and define the potential function as
{\small\begin{align}\label{eq: potential_2side}
\mathcal{P}'(\bx^{(t)},\by^{(t)},\Delta^{(t)}_{\by})= & \mathbb{E}[f(\bx^{(t)},\by^{(t)})] \nonumber \\
& \hspace*{-0.5in} +
\frac{4 + 4(3  L_y^2 + 2) \beta^2 - 7 \beta \gamma }{\beta^2\gamma}
\mathbb{E}\|\Delta^{(t)}_{\by}\|^2.
\end{align}}%
When function $f(\bx,\by)$ is black-box w.r.t. both $\mathbf x$ and $\mathbf y$, we have the following descent w.r.t. $\mathbf y$:
{\small \begin{align}
&\mathcal{P}'(\bx^{(t+1)},\by^{(t+1)},\Delta^{(t+1)}_{\by})\le\mathcal{P}'(\bx^{(t+1)},\by^{(t)},\Delta^{(t)}_{\by})
\nonumber \\
& -\left(\frac{1}{2\beta}-\frac{6L^2_y+4}{\gamma}\right)\mathbb{E}\|\Delta^{(t+1)}_{\by}\|^2  \nonumber \\ & +\left(\frac{6L^2_x}{\gamma^2\beta}+\frac{3\beta L^2_x}{2}\right)\mathbb{E}\|\Delta^{(t+1)}_{\bx}\|^2
+
\frac{7 \beta^2 \gamma^2+ 28 \beta \gamma +12  }{\beta \gamma^2}
\sigma^2_y \nonumber \\
& + \frac{\beta \gamma + 4}{4\beta^2 \gamma }  \mu^2 d^2  L_y^2,
\label{eq.zdesofy}
\end{align}}%
where {$\sigma_y^2 \Def \sigma^2(L_y,\mu, b, q, d)$ given in  \eqref{eq: second_moment_grad_random}}. 
\end{lemma}
\textbf{Proof}: See Appendix\,\ref{sec:zozomax}. \hfill $\square$

Lemma\,\ref{le.le3} is analogous to Lemma\,\ref{le.le2} by taking into account the effect of ZO gradient estimate $\widehat{\nabla}_{\mathbf y} f(\mathbf x, \mathbf y)$ on the potential function \eqref{eq: potential_2side}. {Such an effect is characterized by the terms related to $\sigma_y^2$ and $\mu^2 d^2 L_y^2$  in \eqref{eq.zdesofy}.}

\begin{theorem}\label{th.main2}
{Suppose that \textbf{A1}-\textbf{A2} hold, the sequence $(\bx^{(t)},\by^{(t)})$ over $T$ iterations is generated by  Algorithm \ref{alg: ZO_2side} in which learning rates satisfy}
$\beta<\gamma/(4(3L^2_y+2))$ and  $\alpha\le\min\{L_x, 1/(L_x/2+6L^2_x/(\gamma^2\beta)+3\beta L^2_x/2)\}$. 
\textcolor{black}{When $f(\bx,\by)$ is {black-box w.r.t. both $\mathbf x$ and $\mathbf y$},  the convergence rate of ZO-Min-Max under a uniformly and randomly picked $(\mathbf x^{(r)}, \mathbf y^{(r)})$ from  $\{ (\bx^{(t)},\by^{(t)}) \}_{t=1}^T$ is given by}
{\small \begin{align}
\mathbb{E}\|\mathcal{G}(\bx^{(r)},\by^{(r)})\|^2\le & \frac{c}{\zeta'}\frac{
\mathcal{P}^{\prime}_1
- f^* - \nu^\prime R^2
}{T}  +\frac{c\alpha }{\zeta'} \sigma^2_x \nonumber \\
& + \left ( \frac{cb_1 }{\zeta'} +   d^2 L_y^2  \right ) \mu^2 + \left (  \frac{cb_2 }{\zeta'} + 2 \right )
\sigma^2_y ,
\nonumber
\end{align}}%
where 
$\zeta'$ is a constant independent on the parameters $\mu$, $b$, $q$, $d$ and $T$,
{$\mathcal{P}^{\prime}_t \Def \mathcal{P}^\prime(\bx^{(t)},\by^{(t)},\Delta^{(t)}_{\by})$ in \eqref{eq: potential_2side},
$c$ has been defined in \eqref{eq: thr1_rate}
},  \textcolor{black}{$\nu^\prime = \frac{\min \{ 4 + 4(3  L_y^2 + 2) \beta^2 - 7 \beta \gamma , 0 \} }{\beta^2\gamma}$,}
$  b_1 = L_x + \frac{d^2 L_y^2 (4 + \beta \gamma )}{4\beta^2 \gamma}$ and $ b_2 =\frac{7 \beta^2 \gamma^2 + 28 \beta \gamma + 12 }{\beta \gamma^2 }$,   $\sigma_x^2$ and $\sigma_y^2$ have been introduced in \eqref{eq: inequality_lemma} and \eqref{eq.zdesofy}, and $f^*$, $R$, $\gamma$, $L_x$ and $L_y$ have been defined in \textbf{A1}-\textbf{A2}.
\end{theorem} 
\textbf{Proof}: See Appendix\,\ref{sec:th2}. \hfill $\square$  

{Following the similar argument in Remark\,1 of \thref{th.main1}, one can choose proper learning rates $\alpha$ and $\beta$ to obtain valid lower bound on $\zeta^\prime$. However, different from  \thref{th.main1},
the convergence error  shown by \thref{th.main2} involves
an additional error term related to $\sigma_y^2$ and has worse dimension-dependence on the term related to  $\mu^2$. The latter yields a more restricted choice of the smoothing parameter $\mu$: we   obtain $\mathcal O(1/T + 1/b + d/q)$  convergence  rate when   $\mu \leq  1/(d\sqrt{T})$. 
}

\section{Applications \& Experiment Results
}
\label{sec: exp}
In what follows, we   
evaluate the empirical performance of ZO-Min-Max on two applications of adversarial exploration:
a) design of black-box  ensemble evasion attack against deep neural networks (DNNs), 
and b) design of
black-box poisoning attack against  a logistic regression model.

\subsection{Ensemble   attack via universal perturbation}
\paragraph{A black-box min-max problem formulation.}
We consider the scenario in which the attacker generates a \textit{universal}   adversarial  perturbation over multiple input images     against an \textit{ensemble}
of  multiple classifiers  \citep{liu2016delving,liu2018caad}. 
Considering $I$ classes of images (the     group of images within a class $i$  is denoted by $\Omega_i$)  and $J$ network models, the goal of the adversary is to
find  the  \textit{universal perturbation} $\mathbf x$ additive to
$I$   classes of images against $J$  models. 
The proposed black-box  attack formulation mimics the white-box attack formulation \cite{wang2019unified}
{\small \begin{align}\label{eq: attack_general_minmax}
  \displaystyle \minimize_{ \mathbf x \in \mathcal X } \maximize_{\mathbf w \in \mathcal W} ~~ f(\mathbf x, \mathbf w) \Def  & {\sum_{j=1}^J \sum_{i=1}^I \left [  w_{ij} F_{ij} \left ( \mathbf x;   \Omega_i  \right ) \right ]} \nonumber \\
    & - \lambda \| \mathbf w - \mathbf 1/(IJ) \|_2^2,
\end{align}}%
 where $ \mathbf x \in \mathbb R^d$ and $\mathbf w \in \mathbb R^{IJ}$ are optimization variables,   $w_{ij}$ denotes the $(i,j)$th entry of $\mathbf w$ corresponding to the importance weight of attacking  the set of images at class $i$ under the model $j$, $\mathcal X$ denotes the perturbation constraint, e.g., $\mathcal X = \{  \mathbf x  \, | \, \|  \mathbf x  \|_\infty \leq \epsilon,      \forall \mathbf z \in \cup_i \Omega_i \}$, and $\mathcal W $ is the probabilistic simplex $\mathcal W = \{ \mathbf w ~ | ~ \mathbf 1^T \mathbf w = 1, \mathbf w \geq 0 \}$. In problem \eqref{eq: attack_general_minmax},
  $ F_{ij} \left ( \mathbf x ; \Omega_i \right )$ is the attack loss for attacking  images in $\Omega_i$ under  model $j$,
 and $\lambda > 0$
is a regularization parameter to strike a balance between the worse-case attack loss and the average loss  \cite{wang2019unified}.
We note that 
 $\{ F_{ij} \}$     are   black-box functions w.r.t. $\mathbf x$ since   the adversary  has no access to the  internal configuration s of   DNN models to be attacked. Accordingly, the input gradient cannot be computed by back-propagation.
This implies that problem \eqref{eq: attack_general_minmax} belongs to the  \textit{one-sided black-box} optimization problem (white-box objective w.r.t.  $\mathbf w$). {We refer readers to Appendix\,\ref{app: ensemble_evasion} for more details on \eqref{eq: attack_general_minmax}.}

 \paragraph{Implementation details.} 
We consider $J = 2$ DNN-based classifiers, \textcolor{black}{{Inception-V3} \citep{szegedy2016rethinking}} and \textcolor{black}{{ResNet-50} \citep{he2016deep}}, and $I = 2$ image classes, each of which contains $20$ images  randomly selected from ImageNet \citep{deng2009imagenet}. 

In  \eqref{eq: attack_general_minmax}, we  choose $\lambda = 5$.  
In Algorithm\,\ref{alg: ZO_2side}, we set the learning rates by    $\alpha = 0.05$ and $\beta = 0.01$. Also, we  use the full batch of image samples and set $q = 10$   in   gradient estimation \eqref{eq: grad_rand_ave}, where we set $\mu = 5 \times 10^{-3}$. 
The function query complexity is thus $q$ ($=10$) times more than the number of iterations.  

\paragraph{Baseline methods for comparison.}
In experiments, we compare {ZO-Min-Max} with   (a) \textit{FO-Min-Max}, (b)   \textit{ZO-PGD}  \citep{ghadimi2016mini}, and (c) \textit{ZO-Finite-Sum}, where  the    method (a) is  the FO counterpart of Algorithm\,\ref{alg: ZO_2side}, 
the  method (b)  performs single-objective ZO minimization under the equivalent form of  \eqref{eq: attack_general_minmax}, $\min_{\mathbf x \in \mathcal X} h(\mathbf x)$ where $h(\mathbf x) = \max_{\mathbf w \in \mathcal W} f(\mathbf x, \mathbf w)$,
and the method  (c) performs   ZO-PGD  to minimize  the finite-sum (average) loss 
rather than the worst-case (min-max) loss. We remark that the baseline method (b) calls for the solution to the inner maximization problem $\max_{\mathbf w \in \mathcal W} f(\mathbf x, \mathbf w)$, which is elaborated on {Appendix\,\ref{app: ensemble_evasion}}.
It is also worth mentioning that  although {ZO-Finite-Sum} tackles an objective function different from  \eqref{eq: attack_general_minmax}, it is  motivated by the previous work  on designing the adversarial attack against model ensembles  \citep{liu2018caad}, and we can fairly compare ZO-Min-Max with ZO-Finite-Sum in terms of the attack performance of obtained universal adversarial perturbations. 

  \begin{figure}[htb]
\centerline{ 
\begin{tabular}{c}
\includegraphics[width=.45\textwidth,height=!]{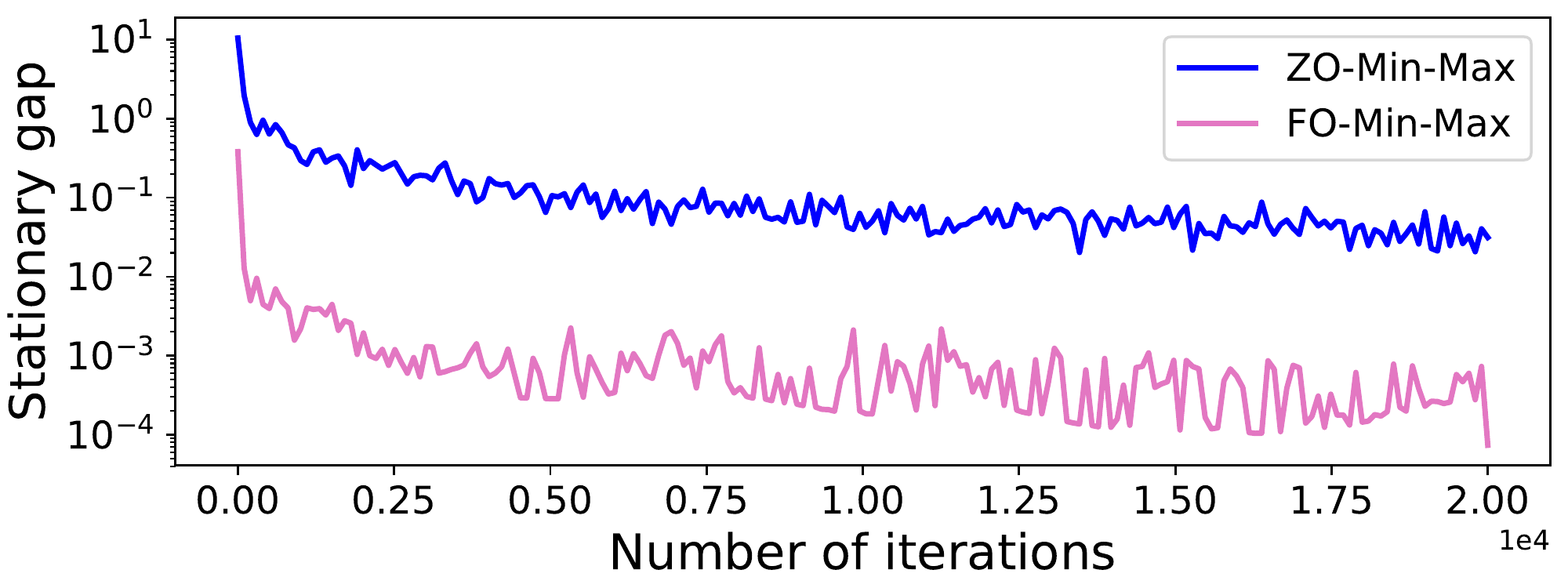}  \\
(a)\\
\includegraphics[width=.45\textwidth,height=!]{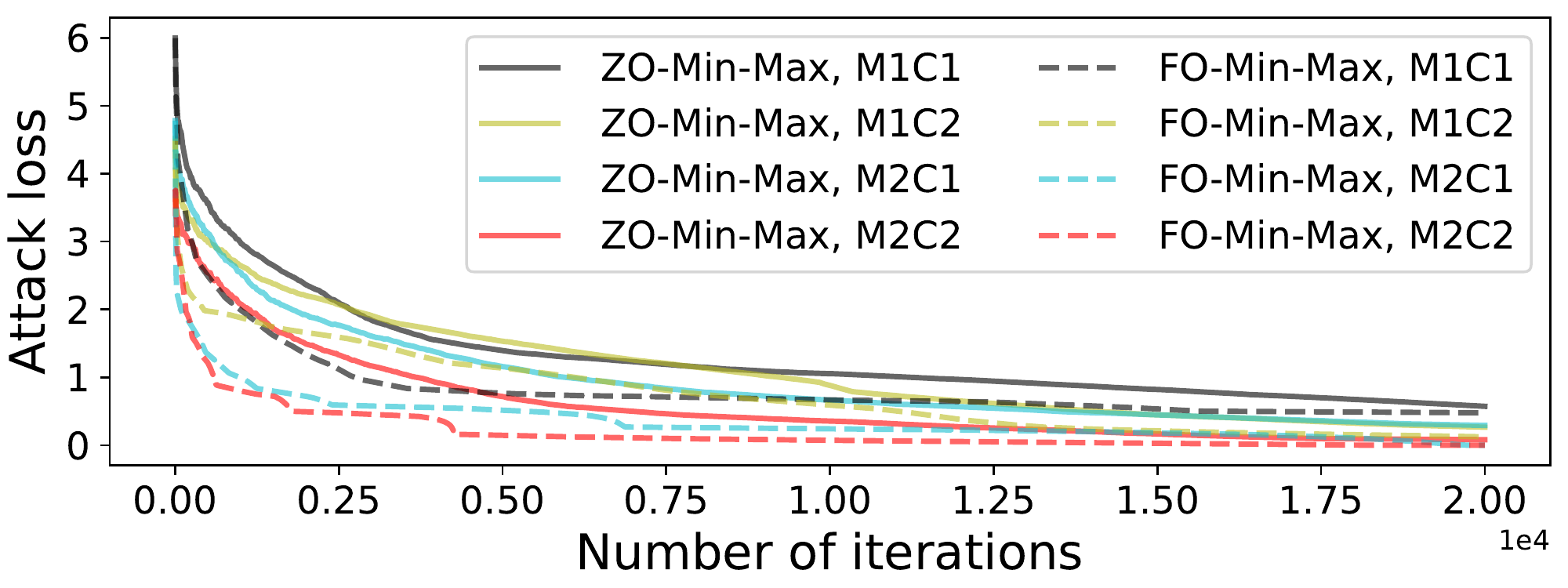}\\
(b)
\end{tabular} 
}
\caption{\footnotesize{ZO-Min-Max vs. FO-Min-Max  in attack generation. (a) Stationary gap;  (b)  attack loss at each model-class pair.
}}
  \label{fig: convergence_gap}
\end{figure}

  \begin{figure}[htb]
\centerline{ 
\begin{tabular}{c}
\includegraphics[width=.4\textwidth,height=!]{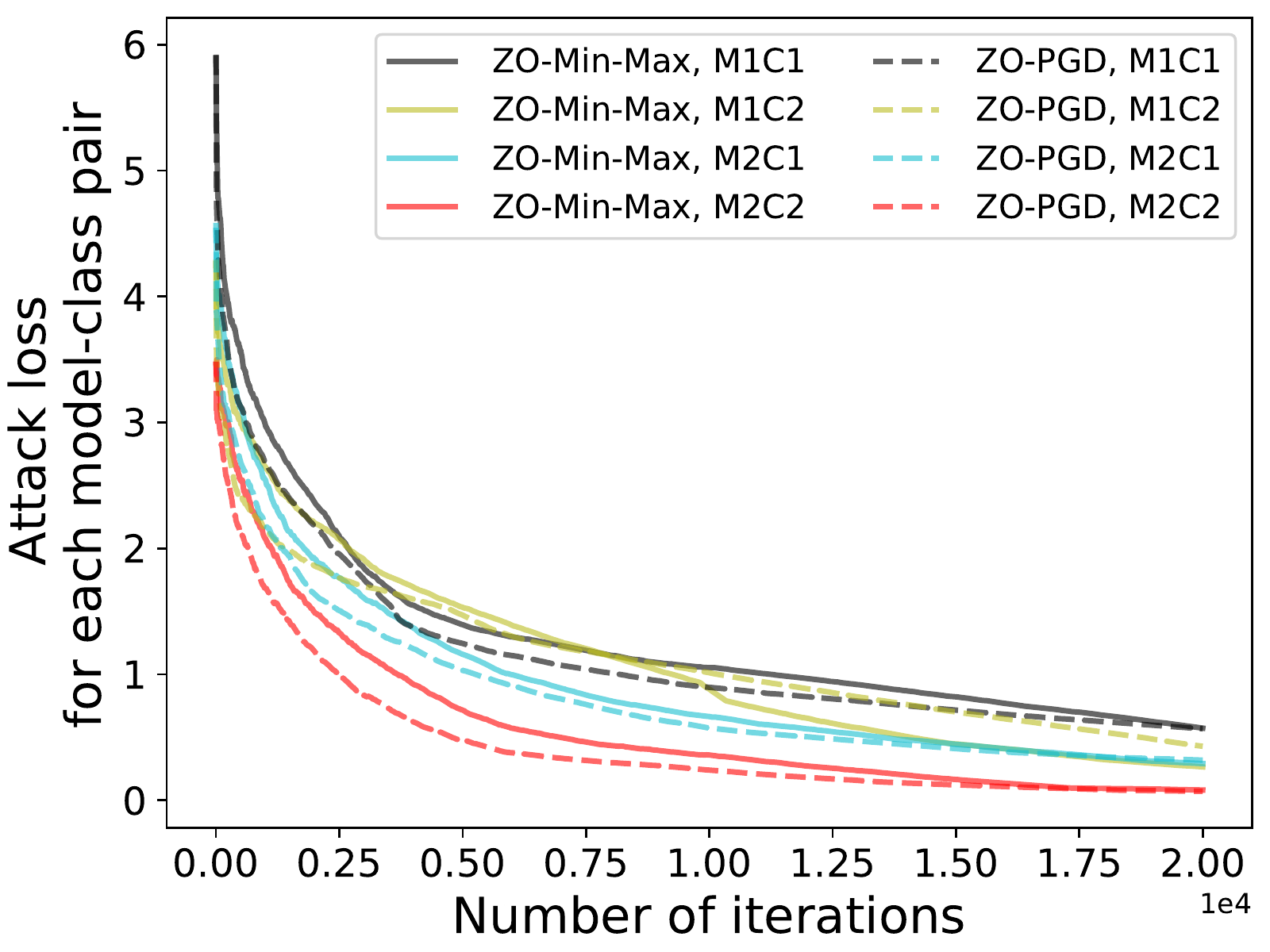} 
\end{tabular} 
}
\caption{\footnotesize{{
Comparison of attack losses achieved by ZO-Min-Max and  ZO-PGD.
}
}}
  \label{fig: ensemble_attack}
\end{figure}

\paragraph{Results.}
{In Figure\,\ref{fig: convergence_gap}, we demonstrate the empirical convergence  of ZO-Min-Max, in terms of the stationary gap $\| \mathcal G(\mathbf x, \mathbf y) \|_2$ given in \eqref{eq.optgap} and the attack loss corresponding to each model-class pair M$j$C$i$.
Here  M and C represents network \underline{m}odel and image \underline{c}lass, respectively.
For comparison, we also present the convergence of FO-Min-Max.  Figure\,\ref{fig: convergence_gap}-(a) shows that  the stationary gap decreases as the iteration increases, and converges to an iteration-independent   bias compared with FO-Min-Max. In Figure\,\ref{fig: convergence_gap}-(b), we see that ZO-Min-Max yields  slightly worse   attack performance (in terms of higher attack loss at each model-class pair) than FO-Min-Max. However, it does not need to have access to the configuration of the victim neural network models.
}

\begin{figure*}[h]
\centerline{ 
\begin{tabular}{ccc}
\includegraphics[width=.33\textwidth,height=!]{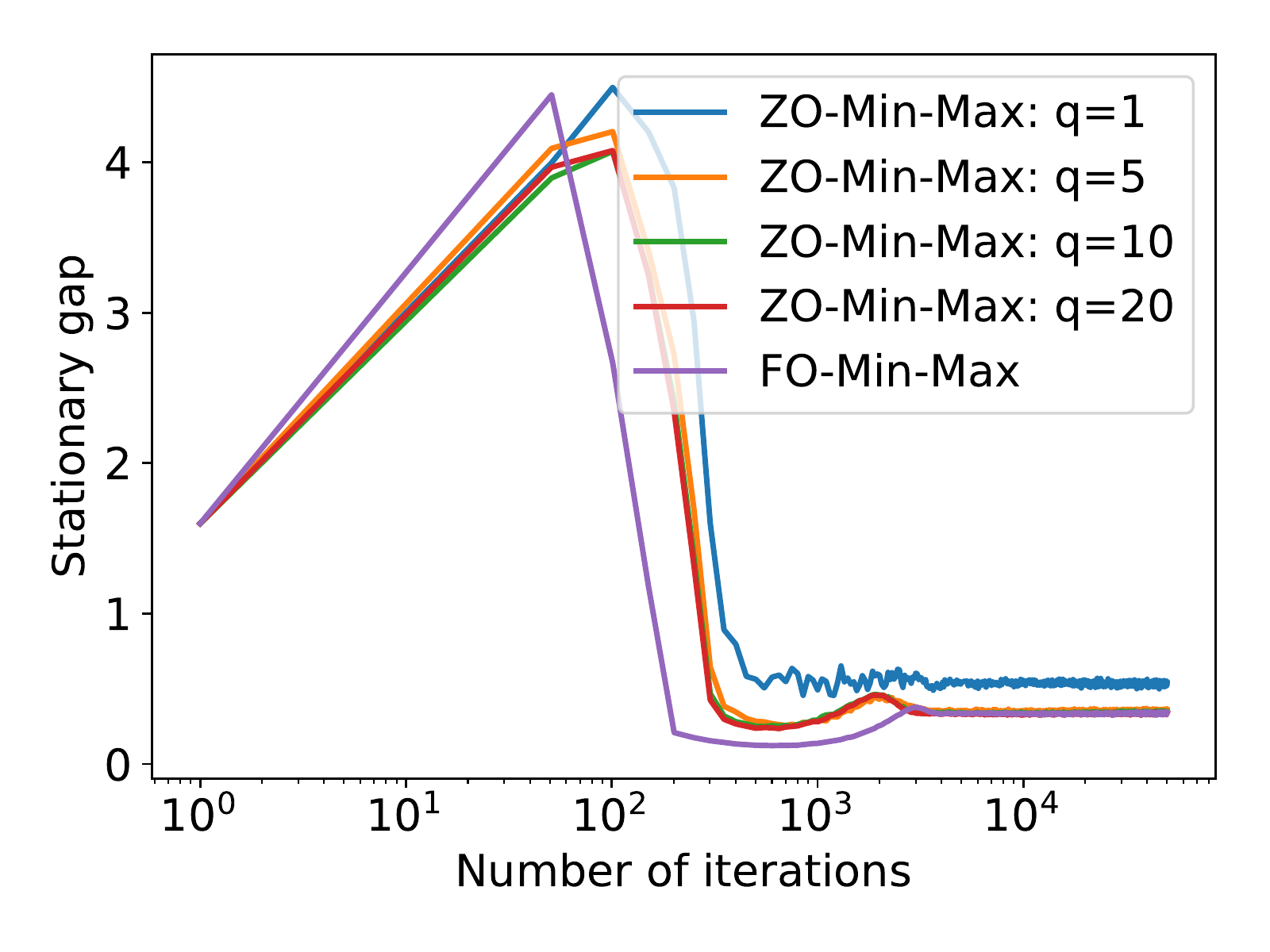} \hspace*{-0.15in} &
\includegraphics[width=.33\textwidth,height=!]{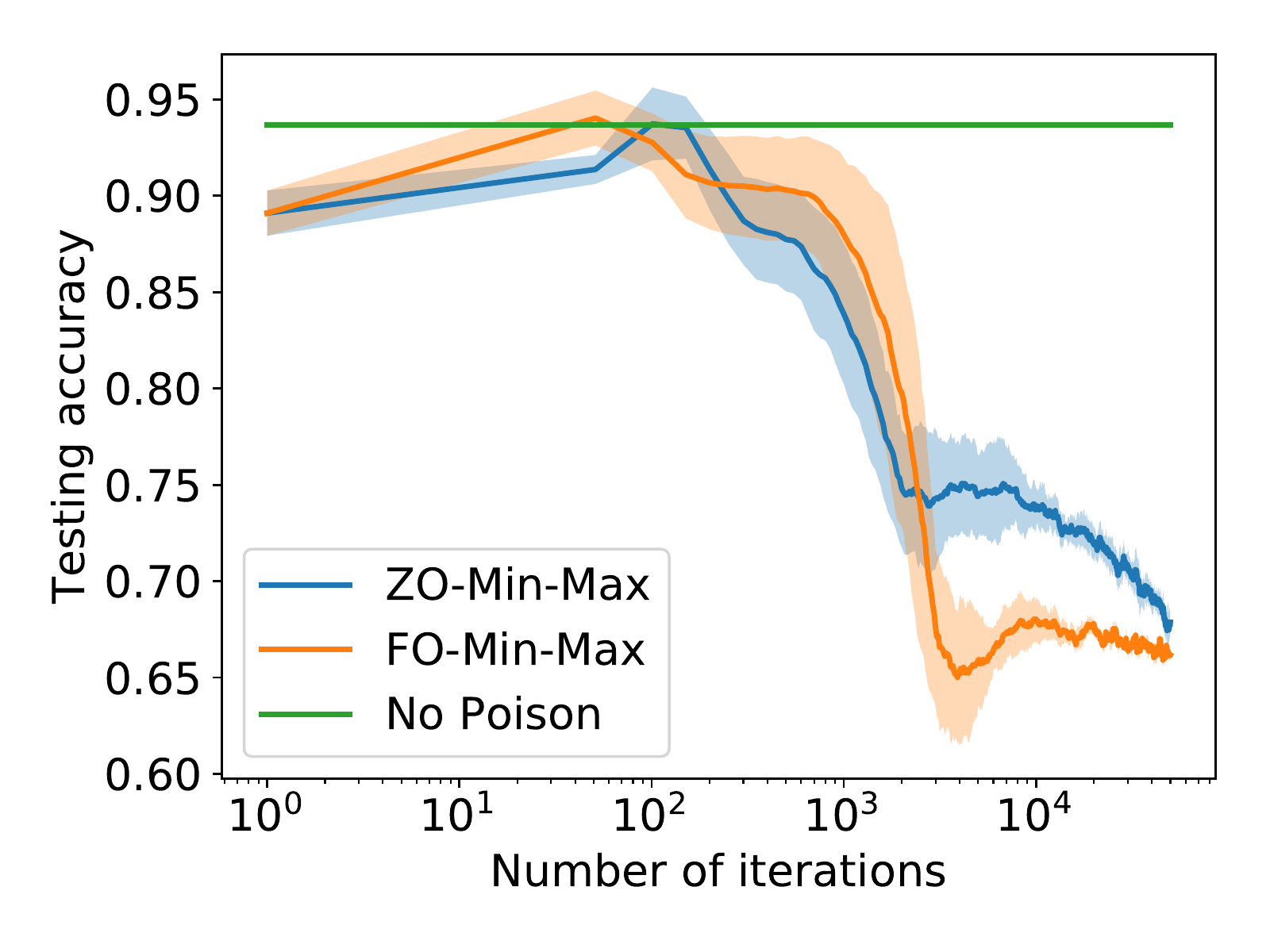}  &\hspace*{-0.15in} 
    \includegraphics[width=0.33\textwidth]{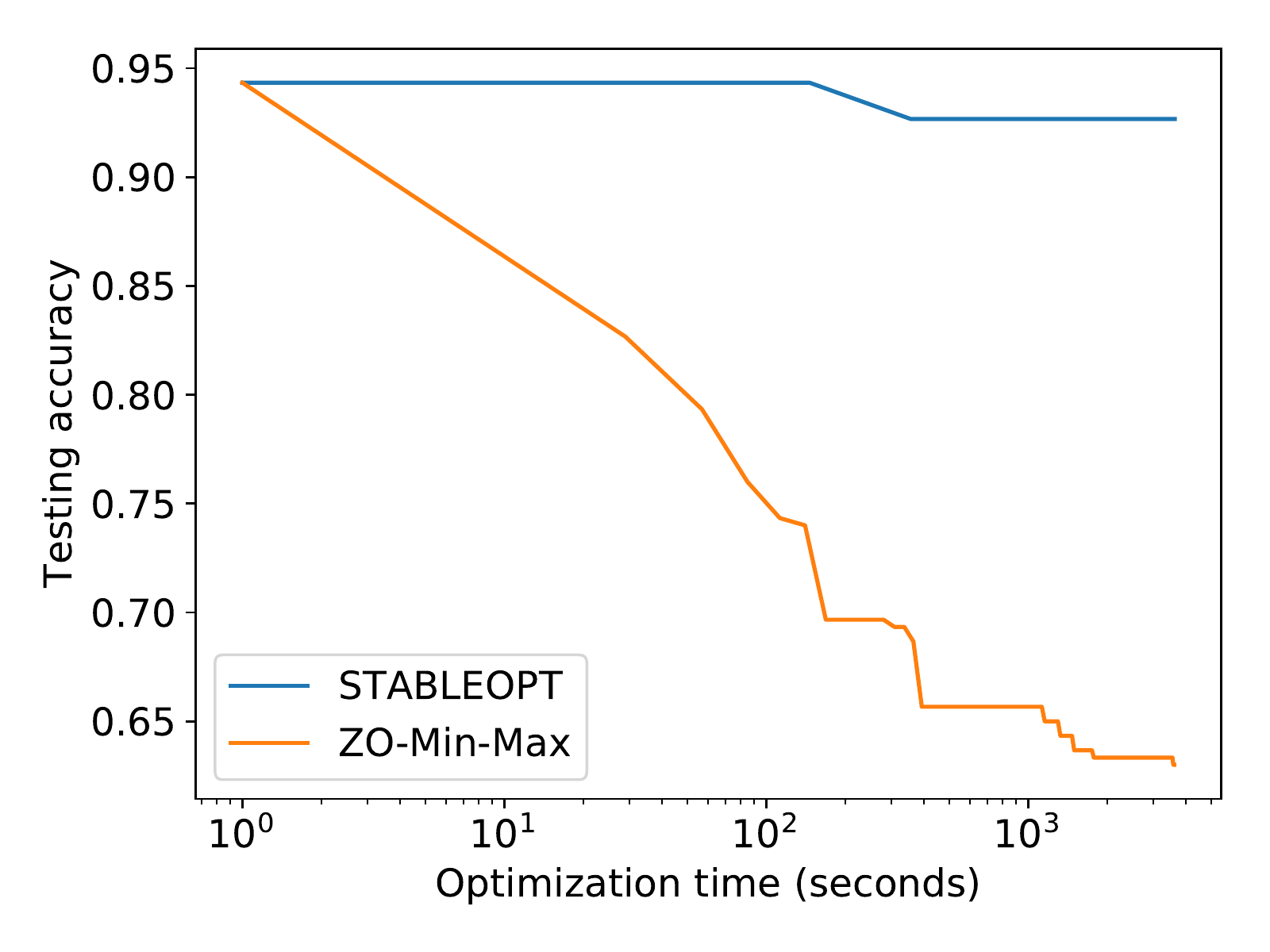} 
\\
(c)\hspace*{-0.05in} & (b) & (c)
\end{tabular} 
}
\caption{\footnotesize{{Empirical performance of ZO-Min-Max in design of poisoning attack: a) stationary gap  versus iterations  b) testing accuracy  versus iterations (the shaded region represents variance of $10$ random trials), and
c) comparison between ZO-Min-Max and STABLEOPT on testing
accuracy versus optimization time.
}}}
  \label{fig: poison_attack}
\end{figure*}

In Figure\,\ref{fig: ensemble_attack},
we compare the attack loss of using ZO-Min-Max with that of using  ZO-PGD, which solves problem  \eqref{eq: attack_general_minmax} by calling for an internal maximization oracle in $\mathbf w$ (see Appendix\,\ref{app: ensemble_evasion}). 
As we can see, ZO-Min-Max converges slower than ZO-PGD at early iterations. That is because the former only performs one-step PGA to update $\mathbf w$, while the latter solves the $\mathbf w$-maximization problem analytically. However, as the number of iterations increases,   ZO-Min-Max achieves almost the same performance as ZO-PGD.

In Appendix\,\ref{app: ensemble_evasion}, we have provided additional comparisons on ZO-Min-Max versus ZO-Finite-Sum, and  versus per-image   PGD attack  \citep{madry2017towards}.

\subsection{Black-box data poisoning attack}\label{sec: exp_poison}
\paragraph{Data poisoning against logistic regression.}
Let $\mathcal D = \{ \mathbf z_i, t_i \}_{i=1}^n $ denote   the training   dataset, among which $n^\prime \ll n$   samples are corrupted by a perturbation vector $\mathbf x$, leading to poisoned training data $\mathbf z_i + \mathbf x$ towards  breaking the training process and thus the prediction accuracy. The poisoning attack problem is then formulated as 
{\small \begin{align}\label{eq: poison_robust_attack}
\hspace*{-0.03in}\displaystyle \maximize_{ \| \mathbf x \|_\infty \leq \epsilon } \minimize_{ \boldsymbol{\theta}  }        ~g(\mathbf x, \boldsymbol{\theta}) \Def \{ F_{\mathrm{tr}}(\mathbf x , \boldsymbol{\theta}; \mathcal D_0) + \lambda \|  \boldsymbol{\theta} \|_2^2 \},
\hspace*{-0.03in}
\end{align}}%
where $\mathbf x $ and $\boldsymbol{\theta}$ are optimization variables,     $F_{\mathrm{tr}}(\mathbf x , \boldsymbol{\theta}; \mathcal D_0)$ denotes the training loss over model parameters $\boldsymbol{\theta}$ at the presence of data poison $\mathbf x$, and  $\lambda > 0$ is a regularization parameter. Note that problem \eqref{eq: poison_robust_attack} can be written  in  the form of \eqref{eq: prob},  $\min_{\mathbf x} \max_{\boldsymbol{\theta}}-g(\mathbf x, \boldsymbol{\theta})$. Clearly, if $F_{\mathrm{tr}}$ is a convex loss, e.g., logistic regression or linear regression \citep{jagielski2018manipulating}), then $-g$ is strongly concave in $\boldsymbol{\theta}$. 
Since  the adversary has no knowledge on the training objective and the benign training data, $g(\mathbf x, \boldsymbol{\theta})$ is a \textit{two-sided black-box} function in both $\mathbf x$ and $\boldsymbol{\theta}$ from the adversary's perspective.

\vspace{-0.1in}
\paragraph{Implementation \& Baseline.}
In problem \eqref{eq: poison_robust_attack},  we set the poisoning ratio $n^\prime / n = 15\%$  and $\lambda = 10^{-3}$ for problem \eqref{eq: poison_robust_attack}, {where the sensitivity of   $\lambda$ is studied in  Figure\,\ref{fig: poison_attack_reg}.}
More details on the specification of problem  \eqref{eq: poison_robust_attack} are provided in Appendix\,\ref{app: poison}.
 In 
  Algorithm\,\ref{alg: ZO_2side}, unless specified otherwise we choose  $b = 100$,  $q =5$, $\alpha = 0.02$, $\beta = 0.05$, and $T = 50000$.
  We report the empirical results \textit{averaged} over $10$ independent trials with random initialization. We   compare 
our method with   FO-Min-Max and the state-of-the-art BO-based  method {STABLEOPT}  \citep{bogunovic2018adversarially}.

In Figure\,\ref{fig: poison_attack}, we present the    convergence of ZO-Min-Max to generate data poisoning attack and validate the resulting attack performance in terms of testing accuracy of the logistic regression model trained on 
the poisoned    dataset. 
Figure\,\ref{fig: poison_attack}-(a) shows the stationary gap of ZO-Min-Max under  different number of random direction vectors in gradient estimation   \eqref{eq: grad_rand_ave}. As we can see, a moderate choice of $q$ (e.g., $q \geq 5$ in our example) is sufficient to achieve near-optimal solution compared with FO-Min-Max. 
However, it suffers from a convergence bias, consistent with Theorem\,\ref{th.main2}.
Figure\,\ref{fig: poison_attack}-(b) demonstrates the testing accuracy (against iterations) of the  model learnt from   poisoned training  data, where the poisoning attack is generated by ZO-Min-Max (black-box attack) and FO-Min-Max (white-box attack). As we can see,  ZO-Min-Max yields promising attacking performance comparable to  FO-Min-Max. We can also see that by contrast with the  testing accuracy  of the clean model ($94\%$ without poison), the poisoning attack eventually  reduces the testing accuracy   (below $70\%$).
Furthermore
In Figure\,\ref{fig: poison_attack}-(c), we compare  ZO-Min-Max with STABLEOPT \citep{bogunovic2018adversarially} in terms of testing accuracy  versus computation time, where the lower the accuracy is, the stronger the generated attack is. We observe that STABLEOPT   has a poorer scalability while our method reaches a data poisoning attack that induces much lower testing accuracy   within $500$ seconds. 
In Appendix\,\ref{app: poison}, we   provide additional results on the   model learnt  under different data poisoning ratios.
\vspace{-0.1in}
\section{Conclusion}
\vspace{-0.1in}
This paper addresses black-box robust optimization problems given a finite number of function evaluations. In particular, we present ZO-Min-Max: a framework of alternating, randomized gradient estimation based ZO optimization algorithm to find a first-order stationary solution to   the black-box min-max problem. Under mild assumptions, ZO-Min-Max enjoys a sub-linear convergence rate. It scales to dimensions that are infeasible for recent robust solvers based on Bayesian optimization. Furthermore, we experimentally demonstrate the potential application of the framework on generating black-box evasion and data poisoning attacks. 
\bibliography{reference,Ref_ZO}
\bibliographystyle{icml2020}
\newpage
\clearpage
\appendix
\input{SupplementaryMaterial.tex}

\end{document}

%% file: SupplementaryMaterial.tex
\setcounter{section}{0}
\setcounter{figure}{0}
\makeatletter 
\renewcommand{\thefigure}{A\@arabic\c@figure}
\makeatother
\setcounter{table}{0}
\renewcommand{\thetable}{A\arabic{table}}
\setcounter{mylemma}{0}
\renewcommand{\themylemma}{A\arabic{mylemma}}
\setcounter{algorithm}{0}
\renewcommand{\thealgorithm}{A\arabic{algorithm}}
\onecolumn
\section*{Appendix}

\section{The ZO-Min-Max Algorithm}
\label{app: alg_code}
      \begin{algorithm}[H]
        \caption{ZO-Min-Max to solve problem \eqref{eq: prob}}
        \begin{algorithmic}[1]
            \State Input: given $\mathbf x^{(0)}$ and $\mathbf y^{(0)}$,   learning rates $\alpha$ and $\beta$,   the number of random directions $q$, and the possible mini-batch size $b$ for stochastic optimization
            \For{$t =  1,2,\ldots, T$}
            \State  $\mathbf x$-step: perform ZO-PGD  \eqref{eq: pgd_out_min}  
            \State $\mathbf y$-step: 
            \If {$f(\mathbf x^{(t)}, \mathbf y)$ is black box w.r.t. $\mathbf y$}
            \State perform ZO-PGA  \eqref{eq: pga_in_max}
            \Else  
            \State perform PGA  using $\nabla_{\mathbf y}f(\mathbf x^{(t)}, \mathbf y^{(t-1)})$ 
             like  \eqref{eq: pga_in_max}
            \EndIf
            \EndFor  
            \end{algorithmic}\label{alg: ZO_2side}
      \end{algorithm}

\section{Detailed Convergence Analysis}
\subsection{Table of Parameters}
\label{app: table_para}

In Table\,\ref{table: parameter}, we summarize the problem and algorithmic parameters used in our convergence analysis.

\begin{table*}[htb]
\centering
\caption{Summary of problem and algorithmic parameters and their descriptions. 
}
\label{table: parameter}
\begin{adjustbox}{max width=1\textwidth }
\begin{threeparttable}
\begin{tabular}{|c|c|}
\hline
parameter          & \begin{tabular}[c]{@{}c@{}} description 
\end{tabular}        
\\ \hline  
$d$ & \# of optimization variables \\ \hline 
$b$ & mini-batch size \\ \hline
$q$ & \# of random direction vectors used in ZO gradient estimation \\ \hline
$\alpha$ & learning rate for ZO-PGD \\ \hline
$\beta$ & learning rate for ZO-PGA \\ \hline
$\gamma$ & strongly concavity parameter of $f(\mathbf x, \mathbf y)$ with respect to $\mathbf y$ \\ \hline
$\eta$ & upper bound on the gradient norm, implying Lipschitz continuity \\ \hline
$L_x$, $L_y$ &  Lipschitz continuous gradient constant of $f(\mathbf x, \mathbf y)$ with respect to $\mathbf x$ and $\mathbf y$ respectively  \\ \hline
$R$ & diameter of the compact convex  set $\mathcal X$ or $\mathcal Y$ \\ \hline
$f^*$ & lower bound on the function value, implying feasibility \\ \hline 
$\sigma_x^2$, $\sigma_y^2$ & variances of ZO gradient estimator for variables $\mathbf x$ and $\mathbf y$, bounded by \eqref{eq: second_moment_grad_random} \\ \hline 
 \end{tabular}
\end{threeparttable}
\end{adjustbox}
\vspace*{-0.1in}
\end{table*}

\subsection{Proof of Lemma \ref{lemma: smooth_f_random_stochastic} } \label{app: proof_variance}

Before going into the proof, let's review some preliminaries and give some definitions. Define $h_{\mu}(\bx,\bxi)$ to be the smoothed version of $h(\bx,\bxi)$ and since $\bxi$ models a subsampling process over a finite number of candidate functions, we can further have $h_{\mu}(\bx) \triangleq \mathbb E_{\bxi}[h_{\mu}(\bx,\boldsymbol{\xi})]$ and $\nabla_{\bx} h_{\mu}(\bx) = \mathbb E_{\bxi}[\nabla_{\mathbf x} h_{\mu}(\bx,\boldsymbol{\xi})]$

Recall that {in the finite sum setting when $\boldsymbol \xi_j$ parameterizes the $j$th function}, the gradient estimator is given by
{\small \begin{align}\label{eq: grad_rand_ave_repeat}
    \widehat{\nabla}_{\mathbf x}h  (\mathbf x ) =
    {   \frac{1}{bq  } \sum_{j \in \mathcal I} \sum_{i=1}^q  \frac{d [h ( \mathbf x + \mu \mathbf u_i; \boldsymbol{\xi}_j ) - h ( \mathbf x  ; \boldsymbol{\xi}_j) ] }{\mu} }  \mathbf u_i.
\end{align}}%
{where $\mathcal I$ is a set with $b$ elements, containing the indices of functions selected for gradient evaluation.}

From standard result of the zeroth order gradient estimator, we know
\begin{align}
    \mathbb E_{\mathcal I}\left[\mathbb E_{\mathbf u_i, i \in [q]}\left[\widehat{\nabla}_{\mathbf x}h  (\mathbf x )\right]\big |\mathcal I\right] 
    = \mathbb E_{\mathcal I}\left[\frac{1}{b} \sum_{j \in \mathcal I}\nabla_{\bx} f_{\mu}(\bx, \bxi_j)\right] =  \nabla_{\bx} h_{\mu}(\bx).
\end{align}

Now let's go into the proof. First, we have
\begin{align}\label{eq: split_variance}
   & \mathbb E \left [
\| 
 \widehat{\nabla}_{\mathbf x}h  (\mathbf x ) - \nabla_{\mathbf x} h_\mu (\mathbf x)
\|_2^2\right] \nonumber \\
 = & \mathbb E_{\mathcal I} \left [ \mathbb E_{\mathbf u_i, i \in [q]} \left[
\left \| 
 \widehat{\nabla}_{\mathbf x}h  (\mathbf x ) - \frac{1}{b} \sum_{j \in \mathcal I}\nabla_{\bx} f_{\mu}(\bx, \bxi_j)  + \frac{1}{b} \sum_{j \in \mathcal I}\nabla_{\bx} f_{\mu}(\bx, \bxi_j) - \nabla_{\mathbf x} h_\mu (\mathbf x)
\right \|_2^2 \bigg |\mathcal I \right]\right] \nonumber \\
\leq & 2 \mathbb E_{\mathcal I} \left [ \mathbb E_{\mathbf u_i, i \in [q]} \left [ 
\left \| 
 \widehat{\nabla}_{\mathbf x}h  (\mathbf x ) - \frac{1}{b} \sum_{j \in \mathcal I}\nabla_{\bx} f_{\mu}(\bx, \bxi_j) \right\|_2^2 +  \left \| \frac{1}{b} \sum_{j \in \mathcal I}\nabla_{\bx} f_{\mu}(\bx, \bxi_j) - \nabla_{\mathbf x} h_\mu (\mathbf x)
\right \|_2^2 \bigg |\mathcal I\right]\right]. 
\end{align}

Further, by definition, given $\mathcal I$, $\widehat{\nabla}_{\mathbf x}h  (\mathbf x )$ is the average of  ZO gradient estimates   under $q$ \emph{i.i.d.} random directions, each of which has the 
 mean $\frac{1}{b} \sum_{j \in \mathcal I}\nabla_{\bx} f_{\mu}(\bx, \bxi_j)$. 
 
Thus  for the first term at the right-hand-side (RHS) of the above inequality, we have
\begin{align}\label{eq: variance_first}
   \mathbb E_{\mathbf u_i, i \in [q]} \left [ 
\left \| 
 \widehat{\nabla}_{\mathbf x}h  (\mathbf x ) - \frac{1}{b} \sum_{j \in \mathcal I}\nabla_{\bx} f_{\mu}(\bx, \bxi_j) \right\|_2^2 \bigg |\mathcal I\right]
 \leq & \frac{1}{q} \left (2d \left\|\frac{1}{b} \sum_{j \in \mathcal I}\nabla_{\bx} f(\bx, \bxi_j)\right\|^2  + \frac{\mu^2 L_h^2 d^2}{2}\right)\nonumber \\
 \leq & \frac{1}{q} \left (2d \eta^2 + \frac{\mu^2 L_h^2 d^2}{2}\right)
\end{align}
where the first inequality is by the standard bound of the variance of zeroth order estimator and the second inequality is by the assumption that $\|\nabla_{\bx} h(x; \boldsymbol{\xi})\|^2 \leq \eta^2$ and thus $ \|\frac{1}{b} \sum_{j \in \mathcal I}\nabla_{\bx} f(\bx, \bxi_j)\|^2 \leq \eta^2$.

 In addition, we have
\begin{align} \label{eq: variance_second}
    & \mathbb E_{\mathcal I} \left [ \mathbb E_{\mathbf u_i, i \in [q]} \left [ 
  \left \| \frac{1}{b} \sum_{j \in \mathcal I}\nabla_{\bx} f_{\mu}(\bx, \bxi_j) - \nabla_{\mathbf x} h_\mu (\mathbf x)
\right \|_2^2 \bigg |\mathcal I\right]\right] \nonumber \\
=& \mathbb E_{\mathcal I} \left [  
  \left \| \frac{1}{b} \sum_{j \in \mathcal I}\nabla_{\bx} f_{\mu}(\bx, \bxi_j) - \nabla_{\mathbf x} h_\mu (\mathbf x)
\right \|_2^2 \right] \nonumber \\
= & \frac{1}{b} \mathbb E_{\xi}  \left [  
  \left \| \nabla_{\bx} f_{\mu}(\bx, \bxi) - \nabla_{\mathbf x} h_\mu (\mathbf x)
\right \|_2^2 \right]  
\leq  \frac{\eta^2}{b}
\end{align}
where the second equality is because $\xi_j$ are i.i.d. draws from the same distribution as $\xi$ and $\mathbb E[\nabla_{\bx} f_{\mu}(\bx, \bxi)]= \nabla_{\mathbf x} h_\mu (\mathbf x) $, the last inequality is because $\| \nabla_{\bx} f_{\mu}(\bx, \bxi) \|_2^2 \leq \eta^2$ by assumption. Substituting \eqref{eq: variance_first} and \eqref{eq: variance_second} into \eqref{eq: split_variance} finishes the proof. 
\hfill $\square$

\subsection{Convergence Analysis of ZO-Min-Max by Performing PGA}\label{app: ZOMinMaxPGA}
In this section, we will provide the details of the proofs. Before proceeding, we have the following illustration, which will be useful in the proof.

\paragraph{The order of taking expectation:}
Since iterates $\bx^{(t)},\by^{(t)},\forall t$ are random variables, we need to define
\begin{equation}
\mathcal{F}^{(t)}=\{\bx^{(t)},\by^{(t)},\bx^{(t-1)},\by^{(t-1)},\ldots,\bx^{(1)},\by^{(1)}\}
\end{equation}
as the history of the iterates. Throughout the theoretical analysis, taking expectation means that we take expectation over random variable at the $t$th iteration conditioned on $\mathcal{F}^{(t-1)}$ and then take expectation over $\mathcal{F}^{(t-1)}$.

\paragraph{Subproblem:}
Also, it is worthy noting that  performing \eqref{eq: pgd_out_min} and \eqref{eq: pga_in_max} are equivalent to the following optimization problem:
\begin{align}
    \bx^{(t)}=&\min_{\bx\in\mathcal{X}}\left\langle\widehat{\nabla}_{\bx}f(\bx^{(t-1)},\by^{(t-1)}),\bx-\bx^{(t-1)}\right\rangle+\frac{1}{2\alpha}\|\bx-\bx^{(t-1)}\|^2,\label{optx}
    \\
    \by^{(t)}=&\max_{\by\in\mathcal{Y}}\left\langle\widehat{\nabla}_{\by}f(\bx^{(t)},\by^{(t-1)}),\by-\by^{(t-1)}\right\rangle-\frac{1}{2\beta}\|\by-\by^{(t-1)}\|^2.\label{opty}
\end{align}
When $f(\bx,\by)$ is white-box {w.r.t. $\mathbf y$}, \eqref{opty} becomes
\begin{equation}\label{bopty}
     \by^{(t)}=\max_{\by\in\mathcal{Y}}\left\langle{\nabla}_{\by}f(\bx^{(t)},\by^{(t-1)}),\by-\by^{(t-1)}\right\rangle-\frac{1}{2\beta}\|\by-\by^{(t-1)}\|^2.
\end{equation}

In the proof of ZO-Min-Max, we will use the optimality condition of these two problems to derive the descent lemmas.

\paragraph{Relationship with smoothing function}

We denote by $f_{\mu, x} (\mathbf x, \mathbf y)$ the smoothing version of $f$ w.r.t. $\mathbf x$ with parameter $\mu > 0$. The similar definition holds for $f_{\mu, y} (\mathbf x, \mathbf y)$. 
By taking $f_{\mu, x} (\mathbf x, \mathbf y)$ as an example, under \textbf{A2} $f$ and $f_{\mu,x}$ has the following relationship \cite[Lemma\,4.1]{gao2014information}: 
\begin{align}
&|{f}_{\mu,\bx}(\bx,\by) - f(\bx,\by) )| \leq \frac{L_x \mu^2}{2} \quad\textrm{and}\quad \|\nabla_{\bx}{f}_{\mu,\bx}(\bx,\by) - \nabla_{\bx}{f}(\bx,\by)  \|_2^2 \leq \frac{\mu^2d^2 L_x^2}{4}\label{eq.xuppersmo}, 
\\
&|{f}_{\mu,\by}(\bx,\by) - f(\bx,\by) )| \leq \frac{L_y \mu^2}{2} \quad\textrm{and}\quad \|\nabla_{\by}{f}_{\mu,\by}(\bx,\by) - \nabla_{\by}{f}(\bx,\by)  \|_2^2 \leq \frac{\mu^2d^2 L_y^2}{4}.\label{eq.yuppersmo}
\end{align}

First, we will show the descent lemma in minimization as follows.
\subsubsection{Proof of \leref{le.common}}\label{app: zomin}

{\bf Proof}: {Since $f(\mathbf x, \mathbf y)$ has $L_x$ Lipschtiz continuous  gradients with respect to $\mathbf x$}, we have 
\begin{align}
    f_{\mu}(\bx^{(t+1)}, \by^{(t)}) 
    \leq &  f_{\mu}(\bx^{(t)}, \by^{(t)}) + \langle \nabla_{\bx} f_{\mu}(\bx^{(t)},\by^{(t)}), \bx^{(t+1)} - \bx^{(t)}\rangle + \frac{L_x}{2} \|\bx^{(t+1)} - \bx^{(t)}\|^2 \nonumber  
    \\
    = & f_{\mu}(\bx^{(t)}, \by^{(t)}) + \langle \widehat{\nabla}_{\bx}  {f}(\bx^{(t)},\by^{(t)}), \bx^{(t+1)} - \bx^{(t)}\rangle +\frac{L_x}{2} \|\bx^{(t+1)} - \bx^{(t)}\|^2 \nonumber 
    \\
    & + \langle {\nabla}_{\bx}  f_{\mu}(\bx^{(t)},\by^{(t)}) - \widehat{\nabla}_{\bx}  {f}(\bx^{(t)},\by^{(t)}), \bx^{(t+1)} - \bx^{(t)}\rangle.
\end{align}

Recall that 
\begin{align}
    \mathbf {x}^{(t+1)} = \mathrm{proj}_{\mathcal X} ( \mathbf {x}^{(t)}- \alpha  \widehat {\nabla}_{\mathbf x} f(\mathbf {x}^{(t)},\mathbf y^{(t)} ) ),
\end{align}

From the optimality condition of $\bx$-subproblem \eqref{optx}, we have
\begin{align}\label{eq: opt_temp}
    \langle \widehat{\nabla}_{\mathbf x} f(\mathbf  {x}^{(t)},\by^{(t)}),   \bx^{(t+1)} - \bx^{(t)}  \rangle \leq -\frac{1}{\alpha} \|\bx^{(t+1)} - \bx^{(t)} \|^2.
\end{align}
Here we use the fact that 
the optimality condition of problem  \eqref{optx} at the solution $\bx^{(t+1)}$   yields
$\langle{\widehat{\nabla}_{\bx}f(\bx^{(t)},\by^{(t)}) + (\bx^{(t+1)}-\bx^{(t)})/\alpha , \bx^{(t+1)} - \mathbf x }\rangle \leq 0$ for any $\mathbf x \in \mathcal X$. By setting $\mathbf x = \mathbf x^{(t)}$, we obtain \eqref{eq: opt_temp}.

In addition, we define another iterate generated by $\nabla_{\bx} f_{\mu}(\bx^{(t)},\by^{(t)})$
\begin{align}
    \mathbf {\widehat x}^{(t+1)} = \mathrm{proj}_{\mathcal X} ( \mathbf {x}^{(t)}- \alpha   {\nabla}_{\mathbf x} f_{\mu}(\mathbf {x}^{(t)},\mathbf y^{(t)} ) ). 
\end{align}
Then, we can have
\begin{align}
     &\langle {\nabla}_{\bx}  f_{\mu}(\bx^{(t)},\by^{(t)}) - \widehat{\nabla}_{\bx}  {f}(\bx^{(t)},\by^{(t)}), \bx^{(t+1)} - \bx^{(t)}\rangle \nonumber \\
     = & \langle {\nabla}_{\bx}  f_{\mu}(\bx^{(t)},\by^{(t)}) - \widehat{\nabla}_{\bx}  {f}(\bx^{(t)},\by^{(t)}), \bx^{(t+1)} - \bx^{(t)} - (\mathbf {\widehat x}^{(t+1)} - \bx^{(t)})\rangle \nonumber \\
     & + \langle {\nabla}_{\bx}  f_{\mu}(\bx^{(t)},\by^{(t)}) - \widehat{\nabla}_{\bx}  {f}(\bx^{(t)},\by^{(t)}),  \mathbf {\widehat x}^{(t+1)} - \bx^{(t)}\rangle.
\end{align}

Due to the fact that 
$\mathbb{E}_{\mathbf u}[\widehat{\nabla}_{\bx}  {f}(\bx^{(t)},\by^{(t)})] = \nabla_{\bx}  f_{\mu}(\bx^{(t)},\by^{(t)}), $
we further have
\begin{align}
    \mathbb{E}_{\mathbf u}[\langle {\nabla}_{\bx}  f_{\mu}(\bx^{(t)},\by^{(t)}) - \widehat{\nabla}_{\bx}  {f}(\bx^{(t)},\by^{(t)}),  \mathbf {\widehat x}^{(t+1)} - \bx^{(t)}\rangle] = 0.
\end{align}

Finally, we also have 
\begin{align}
    &\langle {\nabla}_{\bx}  f_{\mu}(\bx^{(t)},\by^{(t)}) - \widehat{\nabla}_{\bx}  {f}(\bx^{(t)},\by^{(t)}), \bx^{(t+1)} - \bx^{(t)} - (\mathbf {\widehat x}^{(t+1)} - \bx^{(t)})\rangle \nonumber \\
    \leq& \frac{\alpha}{2}\|{\nabla}_{\bx}  f_{\mu}(\bx^{(t)},\by^{(t)}) - \widehat{\nabla}_{\bx}  {f}(\bx^{(t)},\by^{(t)})\|^2 + \frac{1}{2\alpha} \| \bx^{(t+1)} - \bx^{(t)} - (\mathbf {\widehat x}^{(t+1)} - \bx^{(t)}) \|^2 \nonumber \\
    \leq&  \alpha \|{\nabla}_{\bx}  f_{\mu}(\bx^{(t)},\by^{(t)}) - \widehat{\nabla}_{\bx}  {f}(\bx^{(t)},\by^{(t)})\|^2
\end{align}
where the first inequality is due to Young's inequality, the second inequality is due to non-expansiveness of  the projection operator. Thus
\begin{align}
    &\mathbb{E}_{\mathbf u}[\langle {\nabla}_{\bx}  f_{\mu}(\bx^{(t)},\by^{(t)}) - \widehat{\nabla}_{\bx}  {f}(\bx^{(t)},\by^{(t)}), \bx^{(t+1)} - \bx^{(t)} - (\mathbf {\widehat x}^{(t+1)} - \bx^{(t)})\rangle] \nonumber \\
    \leq&  \mathbb{E}_{\mathbf u}[\alpha \|{\nabla}_{\bx}  f_{\mu}(\bx^{(t)},\by^{(t)}) - \widehat{\nabla}_{\bx}  {f}(\bx^{(t)},\by^{(t)})\|^2] \leq \alpha \sigma_x^2
\end{align}
where $\sigma_x^2 \Def \sigma^2(L_x, b, q, d)$ which was defined in \eqref{eq: second_moment_grad_random}.

Combining all above, we have
\begin{align}
   \mathbb{E}[f_{\mu}(\bx^{(t+1)}, \by^{(t)}) ]
    \leq &  \mathbb {E}[f_{\mu}(\bx^{(t)}, \by^{(t)})] - \left(\frac{1}{\alpha}-\frac{L_x}{2}\right) \|\bx^{(t+1)} - \bx^{(t)}\|^2 + \alpha \sigma^2,
\end{align}
and we request $\alpha\le 1/L_x$, which completes the proof. 

Using $ |{f}_{\mu,x}(\bx,\by) - f(\bx,\by) )| \leq \frac{L_x \mu^2}{2}$, we can get
\begin{equation}
\mathbb{E}[f(\bx^{(t+1)},\by^{(t)})]-\frac{L_x\mu^2}{2}\le\mathbb{E}[f_{\mu}(\bx^{(t+1)},\by^{(t)})]\le \mathbb{E}[f(\bx^{(t+1)},\by^{(t)})]+\frac{L_x\mu^2}{2},
\end{equation}
so  we are able to obtain \textcolor{black}{from \eqref{eq: second_moment_grad_random}}
\begin{equation}\label{eq.desofx}
\mathbb{E}[f(\bx^{(t+1)},\by^{(t)})]\le\mathbb{E}[f(\bx^{(t)},\by^{(t)})]-\left(\frac{1}{\alpha}-\frac{L_x}{2}\right)\|\bx^{(t+1)}-\bx^{(t)}\|^2+\alpha \sigma^2_x+L_x\mu^2.
\end{equation}$\square$

\begin{corollary}\label{co:yup}
\begin{equation}
\mathbb{E}    \left\langle\widehat{\nabla}f(\bx^{(t)},\by^{(t-1)})-\nabla f_{\mu}(\bx^{(t)},\by^{(t-1)}),\by^{(t)}-\by^{(t-1)}\right\rangle\le \beta \sigma_y^2
\end{equation}
$\sigma_y^2 \Def \sigma^2(L_y, b, q, d)$ which was defined in \eqref{eq: second_moment_grad_random}.
\end{corollary}
\textbf{Proof:}

Define 
\begin{align}
    \mathbf {\widetilde y}^{(t)} = \mathrm{proj}_{\mathcal Y} ( \mathbf {y}^{(t)}- \beta   {\nabla}_{\mathbf y} f_{\mu}(\mathbf {x}^{(t)},\mathbf y^{(t-1)} ) ), 
\end{align}
we have 
\begin{align}
     &\langle {\nabla}_{\by}  f_{\mu}(\bx^{(t)},\by^{(t-1)}) - \widehat{\nabla}_{\bx}  {f}(\bx^{(t)},\by^{(t-1)}), \by^{(t)} - \by^{(t-1)}\rangle \nonumber \\
     = & \langle {\nabla}_{\by}  f_{\mu}(\bx^{(t)},\by^{(t-1)}) - \widehat{\nabla}_{\by}  {f}(\bx^{(t)},\by^{(t-1)}), \by^{(t)} - \by^{(t-1)} - (\mathbf {\widetilde y}^{(t)} - \by^{(t-1)})\rangle \nonumber \\
     & + \langle {\nabla}_{\by}  f_{\mu}(\bx^{(t)},\by^{(t-1)}) - \widehat{\nabla}_{\by}  {f}(\bx^{(t)},\by^{(t-1)}),  \mathbf {\widetilde y}^{(t)} - \by^{(t-1)}\rangle.
\end{align}

Due to the fact that 
$\mathbb{E}_{\mathbf u}[\widehat{\nabla}_{\by}  {f}(\bx^{(t)},\by^{(t-1)})] = \nabla_{\by}  f_{\mu}(\bx^{(t)},\by^{(t-1)}), $
we further have
\begin{align}
    \mathbb{E}_{\mathbf u}[\langle {\nabla}_{\by}  f_{\mu}(\bx^{(t)},\by^{(t-1)}) - \widehat{\nabla}_{\by}  {f}(\bx^{(t)},\by^{(t-1)}),  \mathbf {\widetilde y}^{(t)} - \by^{(t-1)}\rangle] = 0.
\end{align}

Finally, we also have 
\begin{align}
    &\mathbb{E}_{\mathbf u}[\langle {\nabla}_{\by}  f_{\mu}(\bx^{(t)},\by^{(t-1)}) - \widehat{\nabla}_{\by}  {f}(\bx^{(t)},\by^{(t-1)}), \by^{(t)} - \by^{(t-1)} - (\mathbf {\widetilde y}^{(t)} - \by^{(t-1)})\rangle] \nonumber \\
    \leq & \mathbb{E}_{\mathbf u}[\frac{\beta }{2} \|\langle {\nabla}_{\by}  f_{\mu}(\bx^{(t)},\by^{(t-1)}) - \widehat{\nabla}_{\by}  {f}(\bx^{(t)},\by^{(t-1)})\|^2 + \frac{1 }{2\beta} \|\by^{(t)} - \by^{(t-1)} - (\mathbf {\widetilde y}^{(t)} - \by^{(t-1)})\|^2 ]  \nonumber \\
    \leq &  \mathbb{E}_{\mathbf u}[\beta \|{\nabla}_{\by}  f_{\mu}(\bx^{(t)},\by^{(t-1)}) - \widehat{\nabla}_{\by}  {f}(\bx^{(t)},\by^{(t-1)})\|^2] \leq \beta \sigma_y^2
\end{align}
where $\sigma_y^2 \Def \sigma^2(L_y, b, q, d)$ which was defined in \eqref{eq: second_moment_grad_random}.

\hfill $\square$

Next, before showing the proof of \leref{le.le2}, we need the following lemma to show the recurrence of the size of the successive difference between two iterations.

\begin{lemma}\label{le.le1}
Under assumption 1, assume iterates $\bx^{(t)},\by^{(t)}$ generated by algorithm \ref{alg: ZO_2side}. When $f(\bx^{(t)},\by)$ is white-box, we have
\begin{multline}
\frac{2}{\beta^2\gamma}\mathbb{E}\|\by^{(t+1)}-\by^{(t)}\|^2 -  \frac{2}{\beta^2\gamma}\mathbb{E}\|\by^{(t)}-\by^{(t-1)}\|^2 \le \frac{ 2L^2_x}{\beta\gamma^2}\mathbb{E}\|\bx^{(t+1)}-\bx^{(t)}\|^2
\\
+\frac{2}{\beta}\mathbb{E}\|\by^{(t+1)}-\by^{(t)}\|^2-\left(\frac{4}{\beta}-\frac{2 L^2_y}{\gamma}\right)\mathbb{E}\|\by^{(t)}-\by^{(t-1)}\|^2.\label{eq.condes}
\end{multline}
\end{lemma}

{\bf Proof}:
from the optimality condition of $\by$-subproblem \eqref{bopty}  at iteration $t$ and $t-1$, we have the following two inequalities:
\begin{align}
-\langle\nabla_{\by} f(\bx^{(t+1)},\by^{(t)})-\frac{1}{\beta}(\by^{(t+1)}-\by^{(t)}), \by^{(t+1)}-\by^{(t)}\rangle\le &0, \label{eq: temp_ineq_1}
\\
\langle\nabla_{\by} f(\bx^{(t)},\by^{(t-1)})-\frac{1}{\beta}(\by^{(t)}-\by^{(t-1)}), \by^{(t+1)}-\by^{(t)}\rangle\le & 0. \label{eq: temp_ineq_2}
\end{align}

Adding the above inequalities, we can get
\begin{align}
\notag
\frac{1}{\beta}\langle\bv^{(t+1)},\by^{(t+1)}-\by^{(t)}\rangle\le \left\langle\nabla_{\by} f(\bx^{(t+1)},\by^{(t)})-\nabla_{\by} f(\bx^{(t)},\by^{(t)}),\by^{(t+1)}-\by^{(t)}\right\rangle
\\
+\left\langle\nabla_{\by} f(\bx^{(t)},\by^{(t)})-\nabla_{\by} f(\bx^{(t)},\by^{(t-1)}),\by^{(t+1)}-\by^{(t)}\right\rangle\label{eq:le3eq}
\end{align}
where $\bv^{(t+1)}=\by^{(t+1)}-\by^{(t)}-(\by^{(t)}-\by^{(t-1)})$.

According to the quadrilateral indentity, we know
\begin{equation}\label{eq.quadin}
\left\langle\bv^{(t+1)},\by^{(t+1)}-\by^{(t)}\right\rangle=\frac{1}{2}\left(\|\by^{(t+1)}-\by^{(t)}\|^2+\|\bv^{(t+1)}\|^2-\|\by^{(t)}-\by^{(t-1)}\|^2\right).
\end{equation}
 Based on the definition of $\mathbf v^{(t+1)}$, we substituting \eqref{eq.quadin} into \eqref{eq:le3eq}, which gives
\begin{align}
\notag
\frac{1}{2\beta}\|\by^{(t+1)}-\by^{(t)}\|^2\le&\frac{1}{2\beta}\|\by^{(t)}-\by^{(t-1)}\|^2-\frac{1}{2\beta}\|\bv^{(t+1)}\|^2
\\\notag
&+\left\langle\nabla_{\by} f(\bx^{(t+1)},\by^{(t)})-\nabla_{\by} f(\bx^{(t)},\by^{(t)}),\by^{(t+1)}-\by^{(t)}\right\rangle
\\
&+\left\langle\nabla_{\by} f(\bx^{(t)},\by^{(t)})-\nabla_{\by} f(\bx^{(t)},\by^{(t-1)}),\by^{(t+1)}-\by^{(t)}\right\rangle\label{eq.devyy}
\\\notag
\mathop{\le}\limits^{(a)} &\frac{1}{2\beta}\|\by^{(t)}-\by^{(t-1)}\|^2+\left\langle\nabla_{\by} f(\bx^{(t+1)},\by^{(t)})-\nabla_{\by} f(\bx^{(t)},\by^{(t)}),\by^{(t+1)}-\by^{(t)}\right\rangle
\\\notag
&+\frac{\beta L^2_y}{2}\|\by^{(t)}-\by^{(t-1)}\|^2-\gamma\|\by^{(t)}-\by^{(t-1)}\|^2
\\\notag
\mathop{\le}\limits^{(b)} &\frac{1}{2\beta}\|\by^{(t)}-\by^{(t-1)}\|^2+\frac{\gamma}{2}\|\by^{(t+1)}-\by^{(t)}\|^2
\\
&+\frac{L^2_x}{2\gamma}\|\bx^{(t+1)}-\bx^{(t)}\|^2-(\gamma-\frac{\beta L^2_y}{2})\|\by^{(t)}-\by^{(t-1)}\|^2\label{eq.devyy2}
\end{align}
where in $(a)$ we use the strong concavity of function $f(\bx,\by)$  in $\by$ {(with parameter $\gamma > 0$)} and Young's inequality, i.e.,
\begin{align}
&\langle \nabla_{\by} f(\bx^{(t)},\by^{(t)})-\nabla_{\by} f(\bx^{(t)},\by^{(t-1)}),\by^{(t+1)}-\by^{(t)}\rangle \nonumber
\\
=   &\langle\nabla_{\by} f(\bx^{(t)},\by^{(t)})-\nabla_{\by} f(\bx^{(t)},\by^{(t-1)}),\bv^{(t+1)}+\by^{(t)}-\by^{(t-1)}\rangle  \nonumber
\\
\le & \frac{\beta L^2_y}{2}\|\by^{(t)}-\by^{(t-1)}\|^2+\frac{1}{2\beta}\|\bv^{(t+1)}\|^2-\gamma\|\by^{(t)}-\by^{(t-1)}\|^2
\label{eq: strong_concavity_young}
\end{align} 
and in $(b)$ we apply the Young's inequality, i.e.,
\begin{equation}
    \left \langle\nabla_{\by} f(\bx^{(t+1)},\by^{(t)})-\nabla_{\by} f(\bx^{(t)},\by^{(t)}),\by^{(t+1)}-\by^{(t)}\right\rangle\le\frac{ L^2_x}{2\gamma}\|\bx^{(t+1)}-\bx^{(t)}\|^2+\frac{\gamma}{2}\|\by^{(t+1)}-\by^{(t)}\|^2.\label{eq.youdiff}
\end{equation}
Therefore, we have
\begin{align}\notag
\frac{1}{2\beta}\|\by^{(t+1)}-\by^{(t)}\|^2\le & \frac{1}{2\beta}\|\by^{(t)}-\by^{(t-1)}\|^2+\frac{ L^2_x}{2\gamma}\|\bx^{(t+1)}-\bx^{(t)}\|^2
\\
&+\frac{\gamma}{2}\|\by^{(t+1)}-\by^{(t)}\|^2-\left(\gamma-\frac{\beta L^2_y}{2}\right)\|\by^{(t)}-\by^{(t-1)}\|^2,
\end{align}
which implies
\begin{align}
\notag
\frac{2}{\beta^2\gamma}\|\by^{(t+1)}-\by^{(t)}\|^2\le & \frac{2}{\beta^2\gamma}\|\by^{(t)}-\by^{(t-1)}\|^2+\frac{ 2L^2_x}{\beta\gamma^2}\|\bx^{(t+1)}-\bx^{(t)}\|^2
\\
&+\frac{2}{\beta}\|\by^{(t+1)}-\by^{(t)}\|^2-\left(\frac{4}{\beta}-\frac{2 L^2_y}{\gamma}\right)\|\by^{(t)}-\by^{(t-1)}\|^2.\label{eq.condes}
\end{align}
By taking the expectation on both sides of \eqref{eq.condes}, we can get the results of \leref{le.le1}.
$\square$

\leref{le.le1} basically gives the recursion of $\|\Delta^{(t)}_y\|^2$. It can be observed that term $(4/\beta-2L^2_y/\gamma)\|\Delta^{(t)}_{\by}\|$ provides the descent of the recursion when $\beta$ is small enough, which will take an important role in the proof of \leref{le.le2} when we quantify the descent in maximization.

Then, we can quantify the descent of the objective value by the following descent lemma.

\subsubsection{Proof of \leref{le.le2}}\label{sec:zofomax}
{\bf Proof}: let $f'(\bx^{(t+1)},\by^{(t+1)})=f(\bx^{(t+1)},\by^{(t+1)})-1(\by^{(t+1)})$ and $1(\by)$ denote the indicator function with respect  to the constraint of $\mathbf y$. From the optimality condition of sub-problem $\by$ in \eqref{opty}, we have
\begin{equation}
    \nabla_{\by}f(\bx^{(t+1)},\by^{(t)})-\frac{1}{\beta}(\by^{(t+1)}-\by^{(t)})-\xi^{(t+1)}=0
\end{equation}
where $\xi^{(t)}$ denote the subgradient of $1(\by^{(t)})$. Since function $f'(\bx,\by)$ is concave with respect to $\by$, we have 
\begin{align}
\notag
&f'(\bx^{(t+1)},\by^{(t+1)})-f'(\bx^{(t+1)},\by^{(t)})\le\langle\nabla_{\by}  f(\bx^{(t+1)},\by^{(t)}),\by^{(t+1)}-\by^{(t)}\rangle -\langle \xi^{(t)},\by^{(t+1)}-\by^{(t)}\rangle
\\\notag
\mathop{=}\limits^{(a)}&\frac{1}{\beta}\|\by^{(t+1)}-\by^{(t)}\|^2-\langle\xi^{(t)}-\xi^{(t+1)},\by^{(t+1)}-\by^{(t)}\rangle
\\\notag
=&\frac{1}{\beta}\|\by^{(t+1)}-\by^{(t)}\|^2+\left\langle\nabla_{\by} f(\bx^{(t+1)},\by^{(t)})-\nabla_{\by} f(\bx^{(t)},\by^{(t-1)}),\by^{(t+1)}-\by^{(t)}\right\rangle
\\
&-\frac{1}{\beta}\left\langle\bv^{(t+1)},\by^{(t+1)}-\by^{(t)}\right\rangle\label{eq.le2qe}
\end{align}
where in $(a)$ we use $\xi^{(t+1)}=\nabla_{\by} f(\bx^{(t+1)},\by^{(t)})-\frac{1}{\beta}(\by^{(t+1)}-\by^{(t)})$ .

The last two terms of \eqref{eq.le2qe} is the same as the RHS of \eqref{eq:le3eq}. We can apply the similar steps from \eqref{eq.devyy} to \eqref{eq.devyy2}. To be more specific, the derivations are shown  as follows:
First, we know
\begin{multline}
f'(\bx^{(t+1)},\by^{(t+1)})-f'(\bx^{(t+1)},\by^{(t)})\le
\frac{1}{\beta}\|\by^{(t+1)}-\by^{(t)}\|^2
\\
+\left\langle\nabla_{\by} f(\bx^{(t+1)},\by^{(t)})-\nabla_{\by} f(\bx^{(t)},\by^{(t-1)}),\by^{(t+1)}-\by^{(t)}\right\rangle
-\frac{1}{\beta}\left\langle\bv^{(t+1)},\by^{(t+1)}-\by^{(t)}\right\rangle.
\end{multline}

Then, we move term $1/\beta\langle\bv^{(t+1)},\by^{(t+1)}-\by^{(t)}\rangle$ to RHS of \eqref{eq.le2qe} and have
\begin{align}
\notag
&f(\bx^{(t+1)},\by^{(t+1)})-f(\bx^{(t+1)},\by^{(t)})
\\\notag
\le&\frac{1}{2\beta}\|\by^{(t+1)}-\by^{(t)}\|^2+\frac{1}{2\beta}\|\by^{(t)}-\by^{(t-1)}\|^2-\frac{1}{2\beta}\|\bv^{(t+1)}\|^2
\\\notag
&+\left\langle\nabla_{\by} f(\bx^{(t+1)},\by^{(t)})-\nabla_{\by} f(\bx^{(t)},\by^{(t)}),\by^{(t+1)}-\by^{(t)}\right\rangle
\\\notag
&+\left\langle\nabla_{\by} f(\bx^{(t)},\by^{(t)})-\nabla_{\by} f(\bx^{(t)},\by^{(t-1)}),\by^{(t+1)}-\by^{(t)}\right\rangle
\\\notag
\le &\frac{1}{2\beta}\|\by^{(t+1)}-\by^{(t)}\|^2+\left\langle\nabla_{\by} f(\bx^{(t+1)},\by^{(t)})-\nabla_{\by} f(\bx^{(t)},\by^{(t)}),\by^{(t+1)}-\by^{(t)}\right\rangle
\\\notag
&+\frac{\beta L^2_y}{2}\|\by^{(t)}-\by^{(t-1)}\|^2-\gamma\|\by^{(t)}-\by^{(t-1)}\|^2
\\\notag
\mathop{\le}\limits^{(a)} &\frac{1}{\beta}\|\by^{(t+1)}-\by^{(t)}\|^2+\frac{1}{2\beta}\|\by^{(t)}-\by^{(t-1)}\|^2
\\
&+\frac{\beta L^2_x}{2}\|\bx^{(t+1)}-\bx^{(t)}\|^2-(\gamma-\frac{\beta L^2_y}{2})\|\by^{(t)}-\by^{(t-1)}\|^2\label{eq.devyy3}
\end{align}
where in $(a)$ we use 
\begin{equation}
    \langle\nabla_{\by}f(\bx^{(t+1)},\by^{(t)})-\nabla_{\by}f(\bx^{(t)},\by^{(t)})\rangle\le\frac{\beta L^2_x}{2}\|\bx^{(t+1)}-\bx^{(t)}\|^2+\frac{1}{2\beta}\|\by^{(t+1)}-\by^{(t)}\|^2
\end{equation}
which is different from \eqref{eq.youdiff}; also $\by^{(t)},\by^{(t+1)}\in\mathcal{Y}$ so have $f'(\bx^{(t+1)},\by^{(t+1)})=f(\bx^{(t+1)},\by^{(t+1)})$ and  $f'(\bx^{(t+1)},\by^{(t)})=f(\bx^{(t+1)},\by^{(t)})$.

 Combing \eqref{eq.condes}, we have
\begin{align}
\notag
&f(\bx^{(t+1)},\by^{(t+1)})
+\left(\frac{2}{\beta^2\gamma}+\frac{1}{2\beta}\right)\|\by^{(t+1)}-\by^{(t)}\|^2-4\left(\frac{1}{\beta}-\frac{L^2_y}{2\gamma}\right)\|\by^{(t+1)}-\by^{(t)}\|^2
\\\notag
\le & f(\bx^{(t+1)},\by^{(t)})+\left(\frac{2}{\beta^2\gamma}+\frac{1}{2\beta}\right)\|\by^{(t)}-\by^{(t-1)}\|^2-4\left(\frac{1}{\beta}-\frac{L^2_y}{2\gamma}\right)\|\by^{(t)}-\by^{(t-1)}\|^2
\\
&-\left(\frac{1}{2\beta}-\frac{2L^2_y}{\gamma}\right)\|\by^{(t+1)}-\by^{(t)}\|^2+\left(\frac{2L^2_x}{\gamma^2\beta}+\frac{\beta L^2_x}{2}\right)\|\bx^{(t+1)}-\bx^{(t)}\|^2.\label{eq.desofy}
\end{align}
By taking the expectation on both sides of \eqref{eq.condes}, we can get the results of \leref{le.le2}. 
$\square$

Next, we use the following lemma to show the  descent of the objective value after solving $\bx$-subproblem by \eqref{eq: pgd_out_min}.

\subsubsection{Proof of \thref{th.main1}}\label{sec:th1}
{\bf Proof}:

From \leref{le.le2}, we know
\begin{align}
&\mathbb{E}[f(\bx^{(t+1)},\by^{(t+1)})]
+\left(\frac{2}{\beta^2\gamma}+\frac{1}{2\beta}\right)\mathbb{E}[\|\by^{(t+1)}-\by^{(t)}\|^2] \nonumber \\
& -4\left(\frac{1}{\beta}-\frac{L^2_y}{2\gamma}\right)\mathbb{E}[\|\by^{(t+1)}-\by^{(t)}\|^2] 
\le   \mathbb{E}[f(\bx^{(t+1)},\by^{(t)})] \nonumber \\
& \quad +\left(\frac{2}{\beta^2\gamma}+\frac{1}{2\beta}\right)\mathbb{E}[\|\by^{(t)}-\by^{(t-1)}\|^2]-4\left(\frac{1}{\beta}-\frac{L^2_y}{2\gamma}\right)\mathbb{E}[\|\by^{(t)}-\by^{(t-1)}\|^2] \nonumber 
\\
& \quad -\left(\frac{1}{2\beta}-\frac{2L^2_y}{\gamma}\right)\mathbb{E}[\|\by^{(t+1)}-\by^{(t)}\|^2]+\left(\frac{2L^2_x}{\gamma^2\beta}+\frac{\beta L^2_x}{2}\right)\mathbb{E}[\|\bx^{(t+1)}-\bx^{(t)}\|^2].\label{eq.desofy3}
\end{align}

Combining \leref{le.common}, we have
\begin{align}
&\mathbb{E}[f(\bx^{(t+1)},\by^{(t+1)})]
+\left(\frac{2}{\beta^2\gamma}+\frac{1}{2\beta}\right)\mathbb{E}\left[\|\by^{(t+1)}-\by^{(t)}\|^2\right] \nonumber \\
& -4\left(\frac{1}{\beta}-\frac{L^2_y}{2\gamma}\right)\mathbb{E}\left[\|\by^{(t+1)}-\by^{(t)}\|^2\right]  
\le \mathbb{E}[f(\bx^{(t)},\by^{(t)})]   +\left(\frac{2}{\beta^2\gamma}+\frac{1}{2\beta}\right)\mathbb{E}\left[\|\by^{(t)}-\by^{(t-1)}\|^2\right] \nonumber \\
& \quad -4\left(\frac{1}{\beta}-\frac{L^2_y}{2\gamma}\right)\mathbb{E}\left[\|\by^{(t)}-\by^{(t-1)}\|^2\right]    -\underbrace{\left(\frac{1}{2\beta}-\frac{2L^2_y}{\gamma}\right)}_{c_1}\mathbb{E}\left[\|\by^{(t+1)}-\by^{(t)}\|^2\right] \nonumber \\
& \quad -\underbrace{\left(\frac{1}{\alpha}-\left(\frac{L_x}{2}+\frac{2L^2_x}{\gamma^2\beta}+\frac{\beta L^2_x}{2}\right)\right)}_{c_2}\mathbb{E}\left[\|\bx^{(t+1)}-\bx^{(t)}\|^2\right]
  + \alpha \sigma^2_x+L_x\mu^2.\label{eq.despoten}
\end{align}

If
\begin{equation}
\beta<\frac{\gamma}{4L^2_y}\quad\textrm{and}\quad \alpha<\frac{1}{\frac{L_x}{2}+\frac{2L^2_x}{\gamma^2\beta}+\frac{\beta L^2_x}{2}},
\end{equation}
then we have that there exist {positive} constants $c_1$ and $c_2$ such that
\begin{align}
&\mathcal{P}(\bx^{(t+1)},\by^{(t+1)},\Delta^{(t+1)}_{\by})-\mathcal{P}(\bx^{(t)},\by^{(t)},\Delta^{(t)}_{\by}) \nonumber 
\\
\le&-c_1\mathbb{E}\left[\|\by^{(t+1)}-\by^{(t)}\|^2\right]-c_2\mathbb{E}\left[\|\bx^{(t+1)}-\bx^{(t)}\|^2\right]+ \alpha \nonumber  \sigma^2_x+L_x\mu^2
\\
\le&-\zeta\left(\mathbb{E}\left[\|\by^{(t+1)}-\by^{(t)}\|^2\right]+\mathbb{E}\left[\|\bx^{(t+1)}-\bx^{(t)}\|^2\right]\right)+ \alpha \sigma^2_x+L_x\mu^2\label{eq.deslemma}
\end{align}
where $\zeta=\min\{c_1,c_2\}$.

From \eqref{eq.optgap}, we can have
\begin{align}
\notag
&\|\mathcal{G}(\bx^{(t)},\by^{(t)})\|
\\\notag
\le&\frac{1}{\alpha}\|\bx^{(t+1)}-\bx^{(t)}\|
+\frac{1}{\alpha}\|\bx^{(t+1)}-\textrm{proj}_{\mathcal{X}}(\bx^{(t)}-\alpha\nabla_{\bx} f(\bx^{(t)},\by^{(t)}))\|
+\frac{1}{\beta}\|\by^{(t+1)}-\by^{(t)}\| \\
& +\frac{1}{\beta}\|\by^{(t+1)}-\textrm{proj}_{\mathcal{Y}}(\by^{(t)}+\beta\nabla_{\by} f(\bx^{(t)},\by^{(t)})\| \nonumber
\\\notag
\mathop{\le}\limits^{(a)}&\frac{1}{\alpha}\|\bx^{(t+1)}-\bx^{(t)}\|
\\\notag
&+\frac{1}{\alpha}\|\textrm{proj}_{\mathcal{X}}(\bx^{(t+1)}-\alpha(\nabla_{\bx} f(\bx^{(t)},\by^{(t)})+\frac{1}{\alpha}(\bx^{(t+1)}-\bx^{(t)}))-\textrm{proj}_{\mathcal{X}}(\bx^{(t)}-\alpha\nabla_{\bx} f(\bx^{(t)},\by^{(t)}))\|
\\\notag
&+\frac{1}{\beta}\|\by^{(t+1)}-\by^{(t)}\|
\\\notag
&+\frac{1}{\beta}\|\textrm{proj}_{\mathcal{Y}}(\by^{(t+1)}+\beta(\nabla_{\by} f(\bx^{(t+1)},\by^{(t)})-\frac{1}{\beta}(\by^{(t+1)}-\by^{(t)}))-\textrm{proj}_{\mathcal{Y}}(\by^{(t)}+\beta\nabla_{\by} f(\bx^{(t)},\by^{(t)}))\|
\\\notag
\mathop{\le}\limits^{(b)}&\frac{3}{\alpha}\|\bx^{(t+1)}-\bx^{(t)}\|+\|\nabla_{\by} f(\bx^{(t+1)},\by^{(t)}))-\nabla_{\by} f(\bx^{(t)},\by^{(t)}))\|+\frac{3}{\beta}\|\by^{(t+1)}-\by^{(t)}\|
\\\notag
\mathop{\le}\limits^{(c)}&\left(\frac{3}{\alpha}+L_x\right)\|\bx^{(t+1)}-\bx^{(t)}\|+\frac{3}{\beta}\|\by^{(t+1)}-\by^{(t)}\|
\end{align}
where in $(a)$ we use $\bx^{(t+1)}=\textrm{proj}_{\mathcal{X}}(\bx^{(t+1)}-\alpha\nabla f(\bx^{(t+1)},\by^{(t)})-(\bx^{(t+1)}-\bx^{(t)}))$; 
in $(b)$ we use nonexpansiveness of the projection operator; in $(c)$ we apply the Lipschitz continuous of function $f(\bx,\by)$ with respect to $\bx$ and $\by$ under assumption \textbf{A2}.

Therefore, we can know that there exist a constant $c=\max\{L_x+\frac{3}{\alpha},\frac{3}{\beta}\}$ such that
\begin{equation}\label{eq.optgapi}
    \|\mathcal{G}(\bx^{(t)},\by^{(t)})\|^2\le c\left(\|\bx^{(t+1)}-\bx^{(t)}\|^2+\|\by^{(t+1)}-\by^{(t)}\|^2\right).
\end{equation}
After applying the telescope sum on \eqref{eq.deslemma} and taking expectation over \eqref{eq.optgapi}, we have 
\begin{equation}
    \frac{1}{T}\sum^{T}_{t=1}\mathbb{E}\|\mathcal{G}(\bx^{(t)},\by^{(t)})\|^2\le\frac{c}{\zeta}\left(\frac{\mathcal{P}_1-\mathcal{P}_{T+1}}{T}+ \alpha \sigma^2_x+L_x\mu^2\right).\label{eq.gdes_v0}
\end{equation}
{
Recall from \textbf{A1} that $f \geq f^*$ and $\mathcal Y$ is bounded with diameter $R$, therefore, $\mathcal{P}_t$ given by \eqref{eq: potential_1side} yields
\begin{align}\label{eq: PT_lb}
    \mathcal P_t \geq  
f^* +  \left (\frac{  \min \{  4 + 4 \beta^2 L_y^2 - 7 \beta \gamma , 0\}  }{2\beta^2 \gamma} \right) R^2, \quad\forall t.
\end{align}
And let $(\mathbf x^{(r)}, \mathbf y^{(r)})$ be uniformly and randomly picked from $\{ (\mathbf x^{(t)}, \mathbf y^{(t)}) \}_{t=1}^T$, based on \eqref{eq.gdes_v0} and \eqref{eq: PT_lb}, we obtain
\begin{align}\label{eq.gdes_v1}
   \mathbb E_r [ \mathbb{E}\|\mathcal{G}(\bx^{(r)},\by^{(r)})\|^2  ]  =   \frac{1}{T}\sum^T_{t=1}\mathbb{E}\|\mathcal{G}(\bx^{(t)},\by^{(t)})\|^2\le\frac{c}{\zeta}\left(\frac{\mathcal{P}_1- f^* -\nu R^2 }{T}+ \alpha \sigma^2_x+L_x\mu^2\right),
\end{align}
where recall that $\zeta=\min\{c_1,c_2\}$, $c=\max\{L_x+\frac{3}{\alpha},\frac{3}{\beta}\}$ and $\nu = \frac{  \min \{  4 + 4 \beta^2 L_y^2 - 7 \beta \gamma , 0\}  }{2\beta^2 \gamma}$.   
}

The proof is now complete. 
$\square$

\subsection{Convergence Analysis of ZO-Min-Max by Performing ZO-PGA}

Before showing the proof of \leref{le.le3}, we first give the following lemma regarding to recursion of the difference between two successive iterates of variable $\by$.
\begin{lemma}\label{le.zle1}
Under assumption 1, assume iterates $\bx^{(t)},\by^{(t)}$ generated by algorithm \ref{alg: ZO_2side}. When function $f(\bx^{(t)},\by)$ is black-box, we have
\begin{align}\notag
\frac{2}{\beta^2\gamma}\mathbb{E}\|\by^{(t+1)}-\by^{(t)}\|^2
    \le&\frac{2}{\beta^2\gamma}\mathbb{E}\|\by^{(t)}-\by^{(t-1)}\|^2+\frac{2}{\beta}\mathbb{E}\|\by^{(t+1)}-\by^{(t)}\|^2
\\\notag
&+\frac{6L^2_y}{\beta\gamma^2}\mathbb{E}\|\bx^{(t+1)}-\bx^{(t)}\|^2-\left(\frac{4}{\beta}-\frac{6 L^2_y+4}{\gamma}\right)\mathbb{E}\|\by^{(t)}-\by^{(t-1)}\|^2
\\
&+\frac{4\sigma^2_y}{\beta\gamma}\left(\frac{3}{\gamma}+4\beta\right)+\frac{\mu^2d^2L^2_y}{\beta^2\gamma}.
\end{align}
\end{lemma}
From the optimality condition of $\by$-subproblem in \eqref{opty} at iteration $t$ and $t-1$, we have
\begin{align}
-\left\langle\widehat{\nabla}_{\by} f(\bx^{(t+1)},\by^{(t)})-\frac{1}{\beta}(\by^{(t+1)}-\by^{(t)}), \by^{(t+1)}-\by^{(t)}\right\rangle\le &0,
\\
\left\langle\widehat{\nabla}_{\by} f(\bx^{(t)},\by^{(t-1)})-\frac{1}{\beta}(\by^{(t)}-\by^{(t-1)}), \by^{(t+1)}-\by^{(t)}\right\rangle\le & 0.
\end{align}

Adding the above inequalities and applying the definition of $\bv^{(t+1)}$, we can get
\begin{align}
\notag
\frac{1}{\beta}\langle\bv^{(t+1)},\by^{(t+1)}-\by^{(t)}\rangle\le \underbrace{\left\langle\widehat{\nabla}_{\by} f(\bx^{(t+1)},\by^{(t)})-\widehat{\nabla}_{\by} f(\bx^{(t)},\by^{(t)}),\by^{(t+1)}-\by^{(t)}\right\rangle}_{\textbf{I}}
\\
+\underbrace{\left\langle\widehat{\nabla}_{\by} f(\bx^{(t)},\by^{(t)})-\widehat{\nabla}_{\by} f(\bx^{(t)},\by^{(t-1)}),\by^{(t+1)}-\by^{(t)}\right\rangle}_{\textbf{II}}.\label{eq.zdevyy}
\end{align}

Next, we will bound $\mathbb{E}[\textbf{I}]$ and $\mathbb{E}[\textbf{II}]$ separably as follows.

First, we give an upper bound of $\mathbb{E}[\textbf{I}]$ as the following,
\begin{align}\notag
   &\mathbb{E}\left\langle\widehat{\nabla}_{\by} f(\bx^{(t+1)},\by^{(t)})-\widehat{\nabla}_{\by} f(\bx^{(t)},\by^{(t)}),\by^{(t+1)}-\by^{(t)}\right\rangle 
  \\\notag
  \le&\frac{3}{2\gamma}\mathbb{E}\|\widehat{\nabla}_{\by} f(\bx^{(t+1)},\by^{(t)})-\nabla_{\by} {f}_{\mu,\by}(\bx^{(t+1)},\by^{(t)})\|^2+\frac{\gamma}{6}\mathbb{E}\|\by^{(t+1)}-\by^{(t)}\|^2
  \\\notag
  &+\frac{3}{2\gamma}\mathbb{E}\|{\nabla}_{\by} {f}_{\mu,\by}(\bx^{(t+1)},\by^{(t)})-\nabla_{\by} {f}_{\mu,\by}(\bx^{(t)},\by^{(t)})\|^2+\frac{\gamma}{6}\mathbb{E}\|\by^{(t+1)}-\by^{(t)}\|^2
  \\\notag
  &+\frac{3}{2\gamma}\mathbb{E}\|{\nabla}_{\by} {f}_{\mu,\by}(\bx^{(t)},\by^{(t)})-\widehat{\nabla} {f}_{\by}(\bx^{(t)},\by^{(t)})\|^2+\frac{\gamma}{6}\mathbb{E}\|\by^{(t+1)}-\by^{(t)}\|^2
  \\
  \le& \frac{3\sigma^2_y}{\gamma}+\frac{3L^2_x}{2\gamma}\mathbb{E}\|\bx^{(t+1)}-\bx^{(t)}\|^2+\frac{\gamma}{2}\mathbb{E}\|\by^{(t+1)}-\by^{(t)}\|^2
\end{align}
where \leref{lemma: smooth_f_random_stochastic} is used.

Second, we need to give an upper bound of $\mathbb{E}[\textbf{II}]$ as follows:
\begin{align}
\notag
&\langle \widehat{\nabla} f(\bx^{(t)},\by^{(t)})-\widehat{\nabla} f(\bx^{(t)},\by^{(t-1)}),\by^{(t+1)}-\by^{(t)}\rangle
\\\notag
= &\langle\widehat{\nabla} f(\bx^{(t)},\by^{(t)})-\widehat{\nabla} f(\bx^{(t)},\by^{(t-1)}),\bv^{(t+1)}+\by^{(t)}-\by^{(t-1)}\rangle
\\\notag
=&\left\langle\nabla f(\bx^{(t)},\by^{(t)})-\nabla f(\bx^{(t)},\by^{(t-1)}),\by^{(t)}-\by^{(t-1)}\right\rangle
\\\notag
&+\left\langle\nabla f_{\mu,\by}(\bx^{(t)},\by^{(t)})-\nabla f(\bx^{(t)},\by^{(t)}),\by^{(t)}-\by^{(t-1)}\right\rangle
\\\notag
&+\left\langle\widehat{\nabla}f(\bx^{(t)},\by^{(t)})-\nabla f_{\mu,\by}(\bx^{(t)},\by^{(t)}),\by^{(t)}-\by^{(t-1)}\right\rangle
\\\notag
&-\left\langle\nabla f_{\mu,\by}(\bx^{(t)},\by^{(t-1)})-\nabla f(\bx^{(t)},\by^{(t-1)}),\by^{(t)}-\by^{(t-1)}\right\rangle
\\\notag
&-\left\langle\widehat{\nabla}f(\bx^{(t)},\by^{(t-1)})-\nabla f_{\mu,\by}(\bx^{(t)},\by^{(t-1)}),\by^{(t)}-\by^{(t-1)}\right\rangle
\\\notag
&+\langle\widehat{\nabla} f(\bx^{(t)},\by^{(t)})-\widehat{\nabla} f(\bx^{(t)},\by^{(t-1)}),\bv^{(t+1)}\rangle.
\end{align}
Next, we take expectation on both sides of the above equality and obtain
\begin{align}
\notag
&\mathbb{E}\langle \widehat{\nabla} f(\bx^{(t)},\by^{(t)})-\widehat{\nabla} f(\bx^{(t)},\by^{(t-1)}),\by^{(t+1)}-\by^{(t)}\rangle
\\\notag
\mathop{\le}\limits^{(a)}&\left(\frac{3\beta L^2_y}{2}+\beta\right)\|\by^{(t)}-\by^{(t-1)}\|^2+\frac{1}{2\beta}\|\bv^{(t+1)}\|^2-\gamma\|\by^{(t)}-\by^{(t-1)}\|^2
\\
&+\frac{\mu^2d^2 L^2_y}{4\beta}+4\beta\sigma^2_y\label{eq.bdtermii}
\end{align}
where in $(a)$ we use the fact that {1) $\gamma$-strong concavity of $f$ with respect to $\mathbf y$}:
\begin{equation}
    \left\langle\nabla f(\bx^{(t)},\by^{(t)})-\nabla f(\bx^{(t)},\by^{(t-1)}),\by^{(t)}-\by^{(t-1)}\right\rangle\le -\gamma\|\by^{(t)}-\by^{(t-1)}\|^2;
\end{equation}
and {the facts that 2) smoothing property \eqref{eq.yuppersmo} and Young's inequality }

\begin{equation}
    \mathbb{E}\left\langle\nabla f_{\mu,\by}(\bx^{(t)},\by^{(t)})-\nabla f(\bx^{(t)},\by^{(t)}),\by^{(t)}-\by^{(t-1)}\right\rangle\le\frac{\mu^2 d^2 L^2_y}{8\beta}+\frac{\beta}{2}\|\by^{(t)}-\by^{(t-1)}\|^2;
\end{equation}
and {the fact that 3) the ZO estimator is unbiased according to \leref{lemma: smooth_f_random_stochastic}}
\begin{equation}
    \mathbb{E}\left\langle\widehat{\nabla}f(\bx^{(t)},\by^{(t)})-\nabla f_{\mu,\by}(\bx^{(t)},\by^{(t)}),\by^{(t)}-\by^{(t-1)}\right\rangle=0;
\end{equation}
and
\begin{equation}
    \mathbb{E}\left\langle\nabla f_{\mu,\by}(\bx^{(t)},\by^{(t-1)})-\nabla f(\bx^{(t)},\by^{(t-1)}),\by^{(t)}-\by^{(t-1)}\right\rangle\le\frac{\mu^2 d^2 L^2_y}{8\beta}+\frac{\beta}{2}\|\by^{(t)}-\by^{(t-1)}\|^2;
\end{equation}
and from \coref{co:yup} we have

\begin{equation}
\mathbb{E}    \left\langle\widehat{\nabla}f(\bx^{(t)},\by^{(t-1)})-\nabla f_{\mu,\by}(\bx^{(t)},\by^{(t-1)}),\by^{(t)}-\by^{(t-1)}\right\rangle\le \beta \sigma^2_y;
\end{equation}
and
\begin{align}
\notag
&\mathbb{E}\langle\widehat{\nabla} f(\bx^{(t)},\by^{(t)})-\widehat{\nabla} f(\bx^{(t)},\by^{(t-1)}),\bv^{(t+1)}\rangle
\\\notag
\le&\frac{3\beta}{2}\mathbb{E}\|\nabla f_{\mu,\by}(\bx^{(t)},\by^{(t)})-\widehat{\nabla}f(\bx^{(t)},\by^{(t)})\|^2+\frac{1}{6\beta}\|\bv^{(t+1)}\|^2
\\\notag
&+\frac{3\beta}{2}\mathbb{E}\|\nabla f_{\mu,\by}(\bx^{(t)},\by^{(t)})-{\nabla}f_{\mu,\by}(\bx^{(t)},\by^{(t-1)})\|^2+\frac{1}{6\beta}\|\bv^{(t+1)}\|^2
\\\notag
&+\frac{3\beta}{2}\mathbb{E}\|\nabla f_{\mu,\by}(\bx^{(t)},\by^{(t-1)})-\widehat{\nabla}f(\bx^{(t)},\by^{(t-1)})\|^2+\frac{1}{6\beta}\|\bv^{(t+1)}\|^2
\\
\le&3\beta\sigma^2_y+\frac{1}{2\beta}\|\bv^{(t+1)}\|^2+\frac{3\beta L^2_y}{2}\|\by^{(t)}-\by^{(t-1)}\|^2.
\end{align}

Then, from \eqref{eq.zdevyy}, we can have
\begin{align}\notag
\frac{1}{2\beta}\mathbb{E}\|\by^{(t+1)}-\by^{(t)}\|^2
    \le&\frac{1}{2\beta}\mathbb{E}\|\by^{(t)}-\by^{(t-1)}\|^2-\frac{1}{2\beta}\mathbb{E}\|\bv^{(t+1)}\|^2
\\\notag
&+\frac{3\sigma^2_y}{\gamma}+\frac{3L^2_x}{2\gamma}\mathbb{E}\|\bx^{(t+1)}-\bx^{(t)}\|^2+\frac{\gamma}{2}\mathbb{E}\|\by^{(t+1)}-\by^{(t)}\|^2
\\\notag
&+\left\langle\widehat{\nabla} f(\bx^{(t)},\by^{(t)})-\widehat{\nabla} f(\bx^{(t)},\by^{(t-1)}),\by^{(t+1)}-\by^{(t)}\right\rangle
\\\notag
\le&\frac{1}{2\beta}\mathbb{E}\|\by^{(t)}-\by^{(t-1)}\|^2+\frac{\gamma}{2}\mathbb{E}\|\by^{(t+1)}-\by^{(t)}\|^2
\\\notag
&+\frac{3L^2_y}{2\gamma}\mathbb{E}\|\bx^{(t+1)}-\bx^{(t)}\|^2-\left(\gamma-\left(\frac{3\beta L^2_y}{2}+\beta\right)\right)\mathbb{E}\|\by^{(t)}-\by^{(t-1)}\|^2
\\
&+\frac{3\sigma^2_y}{\gamma}+4\beta\sigma^2_y+\frac{\mu^2d^2L^2_y}{4\beta},\label{eq.zdevyy3}
\end{align}
which implies
\begin{align}\notag
\frac{2}{\beta^2\gamma}\mathbb{E}\|\by^{(t+1)}-\by^{(t)}\|^2
    \le&\frac{2}{\beta^2\gamma}\mathbb{E}\|\by^{(t)}-\by^{(t-1)}\|^2+\frac{2}{\beta}\mathbb{E}\|\by^{(t+1)}-\by^{(t)}\|^2
\\\notag
&+\frac{6L^2_y}{\beta\gamma^2}\mathbb{E}\|\bx^{(t+1)}-\bx^{(t)}\|^2-\left(\frac{4}{\beta}-\frac{6 L^2_y+4}{\gamma}\right)\mathbb{E}\|\by^{(t)}-\by^{(t-1)}\|^2
\\
&+\frac{4\sigma^2_y}{\beta\gamma}\left(\frac{3}{\gamma}+4\beta\right)+\frac{\mu^2d^2L^2_y}{\beta^2\gamma}.
\label{eq.zdevyy2}
\end{align}

\subsubsection{Proof of \leref{le.le3}}\label{sec:zozomax}
{\bf Proof}: Similarly as \ref{sec:zofomax}, let $f'(\bx^{(t+1)},\by^{(t+1)})=f(\bx^{(t+1)},\by^{(t+1)})-1(\by^{(t+1)})$, $1(\cdot)$ denotes the indicator function and $\xi^{(t)}$ denote the subgradient of $1(\by^{(t)})$.  Since function $f'(\bx,\by)$ is concave with respect to $\by$, we have
\begin{align}
\notag
&f'(\bx^{(t+1)},\by^{(t+1)})-f'(\bx^{(t+1)},\by^{(t)})\le\langle\nabla  f(\bx^{(t+1)},\by^{(t)}),\by^{(t+1)}-\by^{(t)}\rangle -\langle \xi^{(t)},\by^{(t+1)}-\by^{(t)}\rangle
\\\notag
\mathop{=}\limits^{(a)}&\frac{1}{\beta}\|\by^{(t+1)}-\by^{(t)}\|^2-\langle\xi^{(t)}-\xi^{(t+1)},\by^{(t+1)}-\by^{(t)}\rangle
\\\notag
=&\frac{1}{\beta}\|\by^{(t+1)}-\by^{(t)}\|^2+\left\langle\widehat{\nabla} f(\bx^{(t+1)},\by^{(t)})-\widehat{\nabla} f(\bx^{(t)},\by^{(t-1)}),\by^{(t+1)}-\by^{(t)}\right\rangle \nonumber \\
& -\frac{1}{\beta}\left\langle\bv^{(t+1)},\by^{(t+1)}-\by^{(t)}\right\rangle
\end{align}
where in $(a)$ we use $\xi^{(t+1)}=\widehat{\nabla} f(\bx^{(t+1)},\by^{(t)})-\frac{1}{\beta}(\by^{(t+1)}-\by^{(t)})$.
Then, we have
\begin{align}\notag
&\mathbb{E}f(\bx^{(t+1)},\by^{(t+1)})-\mathbb{E}f(\bx^{(t+1)},\by^{(t)})+\frac{1}{\beta}\left\langle\bv^{(t+1)},\by^{(t+1)}-\by^{(t)}\right\rangle
\\\notag
\le&\frac{1}{\beta}\|\by^{(t+1)}-\by^{(t)}\|^2+\left\langle\widehat{\nabla} f(\bx^{(t+1)},\by^{(t)})-\widehat{\nabla} f(\bx^{(t)},\by^{(t-1)}),\by^{(t+1)}-\by^{(t)}\right\rangle.
\end{align}

Applying the steps from \eqref{eq.bdtermii} to \eqref{eq.zdevyy3}, we can have
\begin{align}
&\mathbb{E}f(\bx^{(t+1)},\by^{(t+1)})-\mathbb{E}f(\bx^{(t+1)},\by^{(t)}) \nonumber 
\\
\leq &\frac{1}{\beta}\mathbb{E}\|\by^{(t+1)}-\by^{(t)}\|^2+\frac{1}{2\beta}\mathbb{E}\|\by^{(t)}-\by^{(t-1)}\|^2-\left(\gamma-\left(\frac{3\beta L^2_y}{2}+\beta\right)\right)\|\by^{(t)}-\by^{(t-1)}\|^2 \nonumber 
\\
& +\frac{3\beta L^2_x}{2}\mathbb{E}\|\bx^{(t+1)}-\bx^{(t)}\|^2+
7\beta\sigma^2_y+\frac{\mu^2d^2L^2_y}{4\beta}
\end{align}
where we use
\begin{align}
\notag
&\mathbb{E}\left\langle\widehat{\nabla}_{\by} f(\bx^{(t+1)},\by^{(t)})-\widehat{\nabla}_{\by} f(\bx^{(t)},\by^{(t)}),\by^{(t+1)}-\by^{(t)}\right\rangle \\
\le& 3\beta\sigma^2_y+\frac{3\beta L^2_x}{2}\mathbb{E}\|\bx^{(t+1)}-\bx^{(t)}\|^2+\frac{1}{2\beta}\mathbb{E}\|\by^{(t+1)}-\by^{(t)}\|^2.
\end{align}

 Combing \eqref{eq.zdevyy2}, we have
\begin{align}
\notag
&\mathbb{E}f(\bx^{(t+1)},\by^{(t+1)})
+\left(\frac{2}{\beta^2\gamma}+\frac{1}{2\beta}\right)\mathbb{E}\|\by^{(t+1)}-\by^{(t)}\|^2-\left(\frac{4}{\beta}-\frac{6L^2_y+4}{\gamma}\right)\mathbb{E}\|\by^{(t+1)}-\by^{(t)}\|^2
\\\notag
\le & \mathbb{E}f(\bx^{(t+1)},\by^{(t)})+\left(\frac{2}{\beta^2\gamma}+\frac{1}{2\beta}\right)\mathbb{E}\|\by^{(t)}-\by^{(t-1)}\|^2-\left(\frac{4}{\beta}-\frac{6L^2_y+4}{\gamma}\right)\mathbb{E}\|\by^{(t)}-\by^{(t-1)}\|^2
\\\notag
&-\left(\frac{1}{2\beta}-\frac{6L^2_y+4}{\gamma}\right)\mathbb{E}\|\by^{(t+1)}-\by^{(t)}\|^2+\left(\frac{6L^2_x}{\gamma^2\beta}+\frac{3\beta L^2_x}{2}\right)\mathbb{E}\|\bx^{(t+1)}-\bx^{(t)}\|^2.
\\
&+\frac{\mu^2d^2L^2_y}{\beta}\left(\frac{1}{4}+\frac{1}{\beta\gamma}\right)+\left(7\beta+\frac{4}{\beta\gamma}\left(\frac{3}{\gamma}+7\beta\right)\right)\sigma^2_y.\label{eq:desceofpp}
\end{align}
\hfill $\square$

\subsubsection{Proof of \thref{th.main2}}\label{sec:th2}
{\bf Proof}: 
From \eqref{eq.desofx}, we know the ``descent'' of the minimization step, i.e., the changes from $\mathcal{P}'(\bx^{(t)},\by^{(t)},\Delta_{\by}^{(t)})$ to $\mathcal{P}'(\bx^{(t+1)},\by^{(t)},\Delta_{\by}^{(t)})$. 

Combining the ``descent'' of the maximization step by \leref{le.le3} shown in \eqref{eq:desceofpp}, we can obtain the following:
\begin{align}
& \mathcal{P}'(\bx^{(t+1)},\by^{(t+1)},\Delta^{(t+1)}_{\by}) \nonumber \\
\le & \mathcal{P}'(\bx^{(t)},\by^{(t)},\Delta^{(t)}_{\by})
-\underbrace{\left(\frac{1}{2\beta}-\frac{6L^2_y+4}{\gamma}\right)}_{a_1}\mathbb{E}\left[\|\by^{(t+1)}-\by^{(t)}\|^2\right]
\\\notag
&-\underbrace{\left(\frac{1}{\alpha}-\left(\frac{L_x}{2}+\frac{6L^2_x}{\gamma^2\beta}+\frac{3\beta L^2_x}{2}\right)\right)}_{a_2}\mathbb{E}\left[\|\bx^{(t+1)}-\bx^{(t)}\|^2\right]
\\\notag
&+\mu^2\underbrace{\left(L_x+\frac{d^2L^2_y}{\beta}\left(\frac{1}{4}+\frac{1}{\beta\gamma}\right)\right)}_{b_1}+\alpha\sigma^2_x+\underbrace{\left(7\beta+\frac{4}{\beta\gamma}\left(\frac{3}{\gamma}+4\beta\right)\right)}_{b_2}\sigma^2_y.
\end{align}
When $\beta,\alpha$ satisfy the following conditions:
\begin{equation}
    \beta<\frac{\gamma}{4(3L^2_y+2)},\quad\textrm{and}\quad\alpha<\frac{1}{\frac{L_x}{2}+\frac{6L^2_x}{\gamma^2\beta}+\frac{3\beta L^2_x}{2}},
\end{equation}
we can conclude that there exist $b_1,b_2>0$ such that
\begin{align}
\notag
&\mathcal{P}'(\bx^{(t+1)},\by^{(t+1)},\Delta^{(t+1)}_{\by})
\\\notag
\le & \mathcal{P}'(\bx^{(t)},\by^{(t)},\Delta^{(t)}_{\by})-a_1\mathbb{E}\left[\|\by^{(t+1)}-\by^{(t)}\|^2\right] \\
& -a_2\left[\|\bx^{(t+1)}-\bx^{(t)}\|^2\right]+b_1\mu^2+\alpha\sigma^2_x+b_2\sigma^2_y \nonumber 
\\
\le&-\zeta'\mathbb{E}\left[\|\by^{(t+1)}-\by^{(t)}\|^2+\|\bx^{(t+1)}-\bx^{(t)}\|^2\right]+b_1\mu^2+\alpha\sigma^2_x+b_2\sigma^2\label{eq.zdeslemma}
\end{align}
where $\zeta'=\min\{a_1,a_2\}$.

From \eqref{eq.optgap}, we can have
\begin{align}
\notag
&\mathbb{E}\|\mathcal{G}(\bx^{(t)},\by^{(t)})\|
\\\notag
\le&
\frac{1}{\alpha}\mathbb{E}\|\bx^{(t+1)}-\bx^{(t)}\|+\frac{1}{\alpha}\mathbb{E}\|\bx^{(t+1)}-\textrm{proj}_{\mathcal{X}}(\bx^{(t)}-\alpha\nabla_{\bx} f(\bx^{(t)},\by^{(t)}))\|
\\\notag
&+\frac{1}{\beta}\mathbb{E}\|\by^{(t+1)}-\by^{(t)}\|+\frac{1}{\beta}\mathbb{E}\|\by^{(t+1)}-\textrm{proj}_{\mathcal{Y}}(\by^{(t)}+\beta\nabla_{\by} f(\bx^{(t)},\by^{(t)})\|
\\\notag
\mathop{\le}\limits^{(a)}&\frac{1}{\alpha}\mathbb{E}\|\bx^{(t+1)}-\bx^{(t)}\|
 +\frac{1}{\beta}\mathbb{E}\|\by^{(t+1)}-\by^{(t)}\|
\\\notag
&+\frac{1}{\alpha}\mathbb{E}\|\textrm{proj}_{\mathcal{X}}(\bx^{(t+1)}-\alpha(\widehat{\nabla}_{\bx} f(\bx^{(t)},\by^{(t)})+\frac{1}{\alpha}(\bx^{(t+1)}-\bx^{(t)}))-\textrm{proj}_{\mathcal{X}}(\bx^{(t)}-\alpha\nabla_{\bx} f(\bx^{(t)},\by^{(t)}))\|
\\\notag
&+\frac{1}{\beta}\mathbb{E}\|\textrm{proj}_{\mathcal{Y}}(\by^{(t+1)}+\beta(\widehat{\nabla}_{\by} f(\bx^{(t+1)},\by^{(t)})-\frac{1}{\beta}(\by^{(t+1)}-\by^{(t)}))-\textrm{proj}_{\mathcal{Y}}(\by^{(t)}+\beta\nabla_{\by} f(\bx^{(t)},\by^{(t)}))\|
\\\notag
\mathop{\le}\limits^{(b)}&\frac{3}{\alpha}\mathbb{E}\|\bx^{(t+1)}-\bx^{(t)}\|+\mathbb{E}\|\widehat{\nabla}_{\bx} f(\bx^{(t)},\by^{(t)}))-{\nabla}_{\bx} f(\bx^{(t)},\by^{(t)}))\|
\\\notag
&+\frac{3}{\beta}\mathbb{E}\|\by^{(t+1)}-\by^{(t)}\|+\mathbb{E}\|\widehat{\nabla}_{\by} f(\bx^{(t+1)},\by^{(t)})-\nabla_{\by} f(\bx^{(t)},\by^{(t)})\|
\\\notag
\le&\frac{3}{\alpha}\mathbb{E}\|\bx^{(t+1)}-\bx^{(t)}\|+\mathbb{E}\|\widehat{\nabla}_{\bx} f(\bx^{(t)},\by^{(t)}))-{\nabla}_{\bx} f_{\mu,\by}(\bx^{(t)},\by^{(t)}))\|
\\
\notag&+\mathbb{E}\|{\nabla}_{\bx} f_{\mu,\by}(\bx^{(t)},\by^{(t)}))-{\nabla}_{\bx} f(\bx^{(t)},\by^{(t)}))\|
\\\notag
&+\frac{3}{\beta}\mathbb{E}\|\by^{(t+1)}-\by^{(t)}\|+\mathbb{E}\|\widehat{\nabla}_{\by} f(\bx^{(t+1)},\by^{(t)})-\nabla_{\by} f_{\mu,\by}(\bx^{(t+1)},\by^{(t)})\|
\\
\notag&+\mathbb{E}\|{\nabla}_{\by} f_{\mu,\by}(\bx^{(t+1)},\by^{(t)})-\nabla_{\by} f_{\mu,\by}(\bx^{(t)},\by^{(t)})\|
\\\notag
&+\mathbb{E}\|{\nabla}_{\by} f_{\mu,\by}(\bx^{(t)},\by^{(t)})-\nabla_{\by} f(\bx^{(t)},\by^{(t)})\|
\\\notag
\mathop{\le}\limits^{(c)}&\left(\frac{3}{\alpha}+L_x\right)\mathbb{E}\|\bx^{(t+1)}-\bx^{(t)}\|+\frac{3}{\beta}\mathbb{E}\|\by^{(t+1)}-\by^{(t)}\|+2\sigma^2_y+\mu^2d^2L^2_y
\end{align}
where in $(a)$ we use the optimality condition of $\bx^{(t)}$-subproblem; in $(b)$ we use nonexpansiveness of the projection operator; in $(c)$ we apply the Lipschitz continuous of function $f(\bx,\by)$ under assumption \textbf{A2}.

Therefore, we can know that 
\begin{equation}\label{eq.zoptgapi}
    \mathbb{E}\left[\|\mathcal{G}(\bx^{(t)},\by^{(t)})\|^2\right]\le c\left(\|\bx^{(t+1)}-\bx^{(t)}\|^2+\|\by^{(t+1)}-\by^{(t)}\|^2\right)+2\sigma^2_y+\mu^2d^2L^2_y.
\end{equation}
After applying the telescope sum on \eqref{eq.zdeslemma} and taking expectation over \eqref{eq.zoptgapi}, we have
\begin{equation}
    \frac{1}{T}\sum^T_{t=1}\mathbb{E}\left[\|\mathcal{G}(\bx^{(t)},\by^{(t)})\|^2\right]\le\frac{c}{\zeta'}\frac{\mathcal{P}_{1}-\mathcal{P}_{T+1}}{T}+\frac{cb_1}{\zeta'}\mu^2+\frac{c\alpha \sigma^2_x}{\zeta'}+\frac{cb_2}{\zeta'}\sigma^2_y+2\sigma^2_y+\mu^2d^2L^2_y.\label{eq.gdes}
\end{equation}

{
Recall from \textbf{A1} that $f \geq f^*$ and $\mathcal Y$ is bounded with diameter $R$, therefore, $\mathcal{P}_t$ given by \eqref{eq: potential_2side} yields
\begin{align}\label{eq: PT_lb_2side}
    \mathcal P_t \geq  
f^* +  \left (\frac{\min \{ 4 + 4(3  L_y^2 + 2) \beta^2 - 7 \beta \gamma , 0 \} }{\beta^2\gamma} \right) R^2, \quad\forall t.
\end{align}
And let $(\mathbf x^{(r)}, \mathbf y^{(r)})$ be uniformly and randomly picked from $\{ (\mathbf x^{(t)}, \mathbf y^{(t)}) \}_{t=1}^T$, based on  \eqref{eq: PT_lb_2side} and \eqref{eq.gdes}, we obtain
\begin{align}\label{eq.gdes_2side}
 &  \mathbb E_r \left [ \mathbb{E} \left [ \|\mathcal{G}(\bx^{(r)},\by^{(r)})\|^2 \right ]  \right ]  =   \frac{1}{T}\sum^T_{t=1}\mathbb{E}\left [ \|\mathcal{G}(\bx^{(t)},\by^{(t)})\|^2 \right ] \nonumber \\
   \leq & \frac{c}{\zeta'}\frac{\mathcal{P}_{1}- f^* - \nu^\prime R^2}{T}+\frac{cb_1}{\zeta'}\mu^2+\frac{c\alpha \sigma^2_x}{\zeta'}+\frac{cb_2}{\zeta'}\sigma^2_y+2\sigma^2_y+\mu^2d^2L^2_y
\end{align}
where recall that $\zeta'=\min\{a_1,a_2\}$, $c=\max\{L_x+\frac{3}{\alpha},\frac{3}{\beta}\}$, and $\nu^\prime = \frac{\min \{ 4 + 4(3  L_y^2 + 2) \beta^2 - 7 \beta \gamma , 0 \} }{\beta^2\gamma}$.
}

The proof is now complete. $\square$

\clearpage
\newpage
\section{Toy Example in \cite{bogunovic2018adversarially}: ZO-Min-Max versus BO}
\label{app: toy}

We review the example in \cite{bogunovic2018adversarially} as below,
\begin{align}\label{eq: toy}
\begin{array}{ll}
  \displaystyle \maximize_{\mathbf x \in \mathcal C} \minimize_{\| \boldsymbol{\delta} \|_2 \leq 0.5}         &  f(\mathbf x- \boldsymbol{\delta}) \Def  -2 (x_1 - \delta_1)^6 + 12.2  (x_1 - \delta_1)^5  
  - 21.2 (x_1 - \delta_1)^4 \\
  & - 6.2  (x_1 - \delta_1) + 6.4 (x_1 - \delta_1)^3  
   + 4.7 (x_1 - \delta_1)^2 - (x_2 - \delta_2)^6 \\
   & + 11 (x_2 - \delta_2)^5 - 43.3 (x_2 - \delta_2)^4 + 10 (x_2 - \delta_2) + 74.8 (x_2 - \delta_2)^3 \\
   & - 56.9 (x_2 - \delta_2)^2 + 4.1 (x_1 - \delta_1) (x_2 - \delta_2) + 0.1 (x_1 - \delta_1)^2 (x_2 - \delta_2)^2 \\
   & - 0.4 (x_2 - \delta_2)^2 (x_1 - \delta_1) - 0.4 (x_1 - \delta_1)^2(x_2 - \delta_2)  ,
\end{array}
\end{align}
where $\mathbf x \in \mathbb R^2$, and 
$\mathcal C = \{  x_1 \in (-0.95,3.2), x_2 \in (-0.45,4.4)  \}
$.  

Problem \eqref{eq: toy} can be equivalently transformed to the min-max setting consistent with ours
\begin{align}\label{eq: toy_equi}
\begin{array}{ll}
  \displaystyle \minimize_{\mathbf x \in \mathcal C} \maximize_{\| \boldsymbol{\delta} \|_2 \leq 0.5}         &  - f(\mathbf x- \boldsymbol{\delta}).  
\end{array}
\end{align}

The optimality of solving problem \eqref{eq: toy} is measured by regret versus iteration $t$,
\begin{align}
    \mathrm{Regret}(t) = \minimize_{\| \boldsymbol{\delta} \|_2 \leq 0.5}   f(\mathbf x^*- \boldsymbol{\delta}) - \minimize_{\| \boldsymbol{\delta} \|_2 \leq 0.5}   f(\mathbf x^{(t)}- \boldsymbol{\delta}),
\end{align}
where $\minimize_{\| \boldsymbol{\delta} \|_2 \leq 0.5}   f(\mathbf x^*- \boldsymbol{\delta}) = -4.33$ and $\mathbf x^* = [-0.195, 0.284]^T$   \cite{bogunovic2018adversarially}. 

\begin{figure}[htb]
\centering
  \begin{tabular}{cc}
\includegraphics[width=.5\textwidth,height=!]{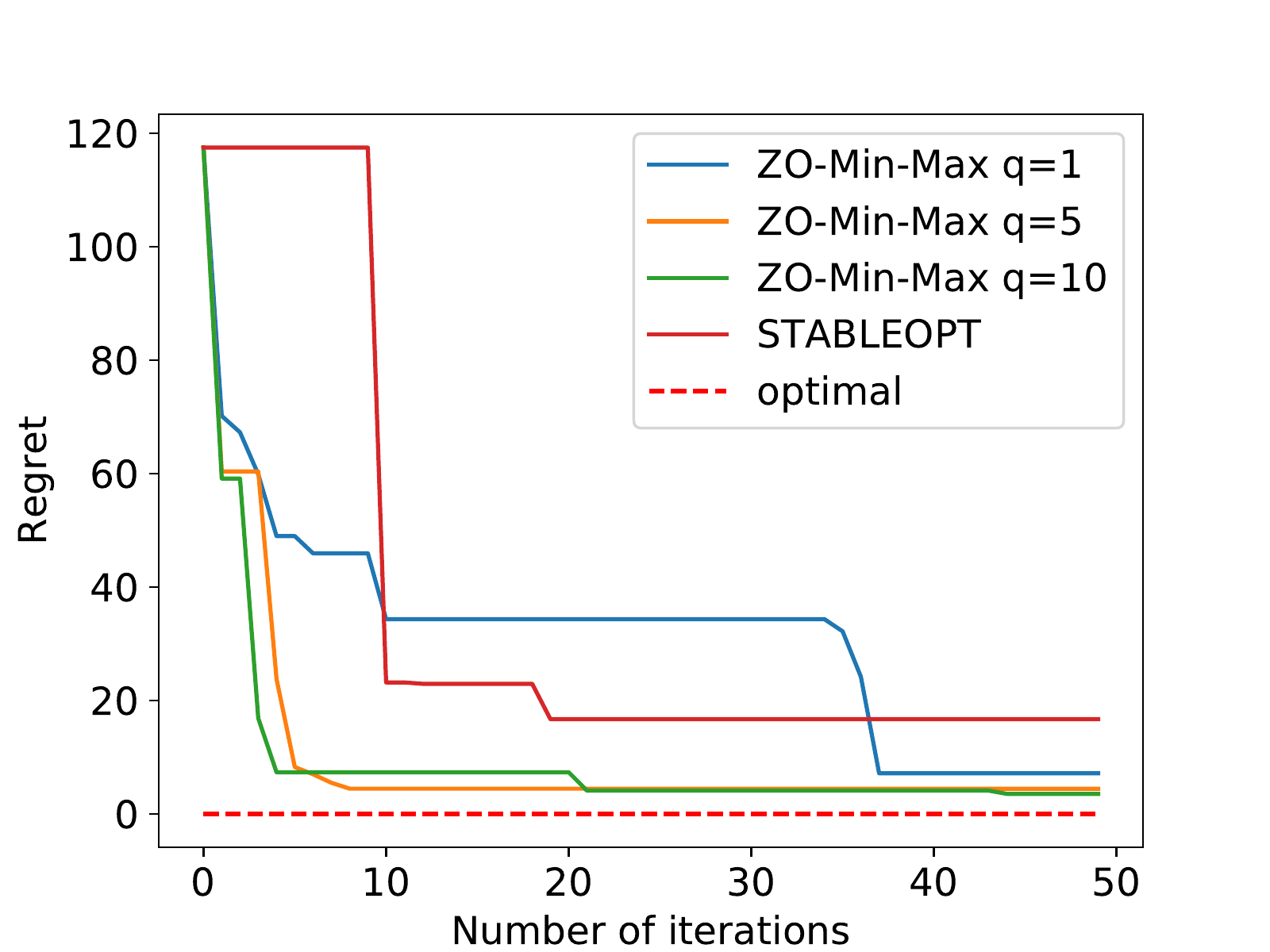} &
\includegraphics[width=.5\textwidth,height=!]{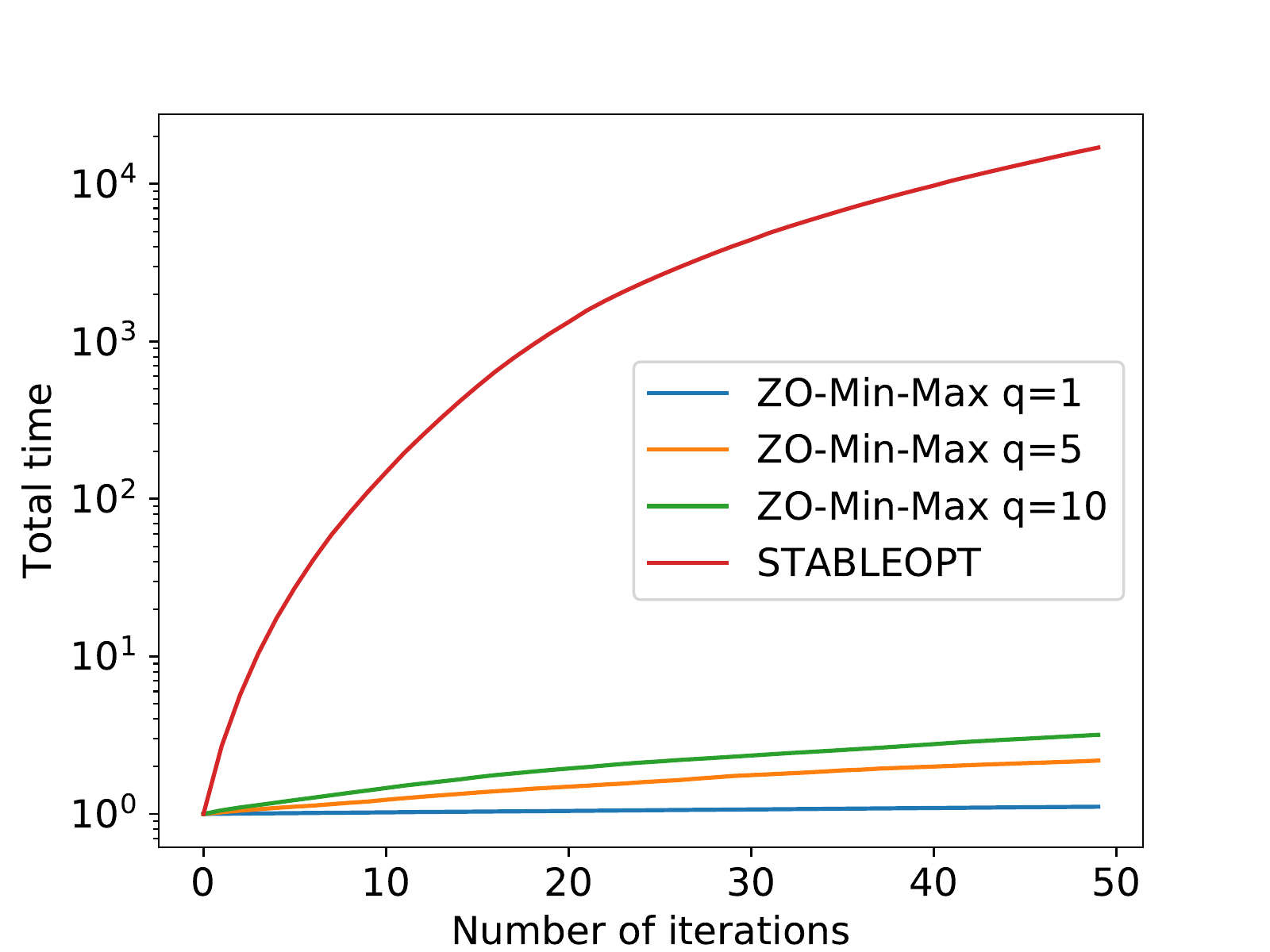} \\
(a) & (b)
\end{tabular} 
    \caption{
    \footnotesize{Comparison of ZO-Min-Max against STABLEOPT  \cite{bogunovic2018adversarially}: a) Convergence performance; b) Computation time (seconds).
  }}
  \label{fig: ZOMinMax_BO}
\end{figure}

In Figure\,\ref{fig: ZOMinMax_BO}, we compare the convergence performance and computation time of  ZO-Min-Max with the BO based approach STABLEOPT proposed in \cite{bogunovic2018adversarially}.
Here we choose the same  initial point for both ZO-Min-Max and STABLEOPT. And we   set the same number of function queries per iteration for ZO-Min-Max (with $q = 1$) and STABLEOPT. We recall from \eqref{eq: grad_rand_ave} that the larger $q$ is, the more queries ZO-Min-Max takes. In our experiments, we present the best achieved regret up to time $t$ and report the average performance of each method over $5$ random trials. As we can see,   ZO-Min-Max is more stable, with lower regret and less running time. Besides,  as $q$ becomes larger, ZO-Min-Max has a faster convergence rate. We remark that 
BO is slow since learning the accurate GP model and solving the acquisition problem takes intensive  computation cost. 

\clearpage
\newpage

\section{Additional Details on Ensemble Evasion Attack}\label{app: ensemble_evasion}

\paragraph{Experiment setup.}
We   specify the   attack loss $F_{ij}$ in \eqref{eq: attack_general_minmax} as the C\&W  untargeted   attack  loss \citep{carlini2017towards},
{\small  \begin{align}\label{eq: loss_attack_ij}
    F_{ij} \left ( \mathbf x;   \Omega_i  \right )= ({1}/{|\Omega_i|}) \sum_{\mathbf z \in \Omega_i}  
     \max \{    g_j(\mathbf z + \mathbf x )_{i} - \max_{k \neq i}  g_j(\mathbf z + \mathbf x )_k, 0 \},  
 \end{align}}%
 where $|\Omega_i|$ is the cardinality of the set $\Omega_i$,
$g_j(\mathbf z + \mathbf x )_k$ denotes the prediction score of class $k$ given the input $\mathbf z + \mathbf x$ using model $j$. 
In  \eqref{eq: attack_general_minmax}, the regularization parameter $\lambda$
trikes a balance between the worse-case attack loss and the average loss  \cite{wang2019unified}. The rationale behind that is from two manifolds. First, as $\gamma = 0$, then $ \maximize_{\mathbf w \in \mathcal W} \,    {\sum_{j=1}^J \sum_{i=1}^I \left [  w_{ij} F_{ij} \left ( \mathbf x;   \Omega_i  \right ) \right ]} = F_{i^*j^*} \left ( \mathbf x;   \Omega_i  \right )$, where $w_{i^* j^*} = 1$ and $0$s for $(i,j) \neq (i^*, j^*)$, and $(i^*, j^*) = \argmax_{i,j} F_{ij} \left ( \mathbf x;   \Omega_i  \right )$ given $\mathbf x$. On the other hand, as $\gamma \to \infty$, then $\mathbf w \to \mathbf 1/(IJ)$.

 \paragraph{Implementation of ZO-PGD for solving  problem   \eqref{eq: attack_general_minmax}.}
 To solve problem \eqref{eq: attack_general_minmax}, the baseline method ZO-PGD
   performs single-objective ZO minimization under the equivalent form of  \eqref{eq: attack_general_minmax}, $\min_{\mathbf x \in \mathcal X} h(\mathbf x)$, where $h(\mathbf x) = \max_{\mathbf w \in \mathcal W} f(\mathbf x, \mathbf w)$. 
   It is   worth noting that   we report the best convergence performance of ZO-PGD by   searching   its learning rate over $5$ grid points in $[0.01, 0.05]$.
   Also, when querying the function value of $h$ (at a given point $\mathbf x$), we need  the solution to the inner maximization problem 
    {\small \begin{align}\label{eq: attack_general_minmax_inner}
  \displaystyle   \maximize_{\mathbf w \in \mathcal W} ~~  {\sum_{j=1}^J \sum_{i=1}^I \left [  w_{ij} F_{ij} \left ( \mathbf x;   \Omega_i  \right ) \right ]}   - \lambda \| \mathbf w - \mathbf 1/(IJ) \|_2^2.
\end{align}}%
Problem \eqref{eq: attack_general_minmax_inner} is equivalent to 
   \begin{align}
    \begin{array}{ll}
\displaystyle \minimize_{\mathbf w}          &    \lambda \| \mathbf w - \mathbf 1/(IJ) - (1/(2\lambda))\mathbf f(\mathbf x) \|_2^2 \\
    \st      &  \mathbf 1^T \mathbf w = 1, \mathbf w \geq 0
    \end{array}
\end{align}
where $\mathbf f(\mathbf x) \Def [F_{11} (\mathbf x), \ldots, f_{IJ}(\mathbf x)]^T$.
The solution is given by the projection of the point $ \mathbf 1/(IJ) + (1/(2\lambda))\mathbf f(\mathbf x)$  on the the probabilistic simplex \citep{parikh2014proximal}
\begin{align}
    \mathbf w^* = \left [  \mathbf 1/(IJ) + (1/(2\lambda))\mathbf f(\mathbf x)- \mu \mathbf 1 \right ]_+,
\end{align}
where $[\cdot]_+$ is element-wise non-negative operator, and
$\mu$ is the root of the equation
\begin{align}
    \mathbf 1^T \left [ \mathbf 1/(IJ) + (1/(2\lambda))\mathbf f(\mathbf x) - \mu \mathbf 1 \right ]_+ = \sum_{i} \max \{ 0,  1/(IJ) +  f_i(\mathbf x) /(2\lambda) - \mu \} = 1.
\end{align}
The above equation in $\mu$ can be solved using the bisection method at a given $\mathbf x$ \citep{boyd2004convex}.

\begin{figure}[htb]
\centerline{ 
\begin{tabular}{cc}
\includegraphics[width=.45\textwidth,height=!]{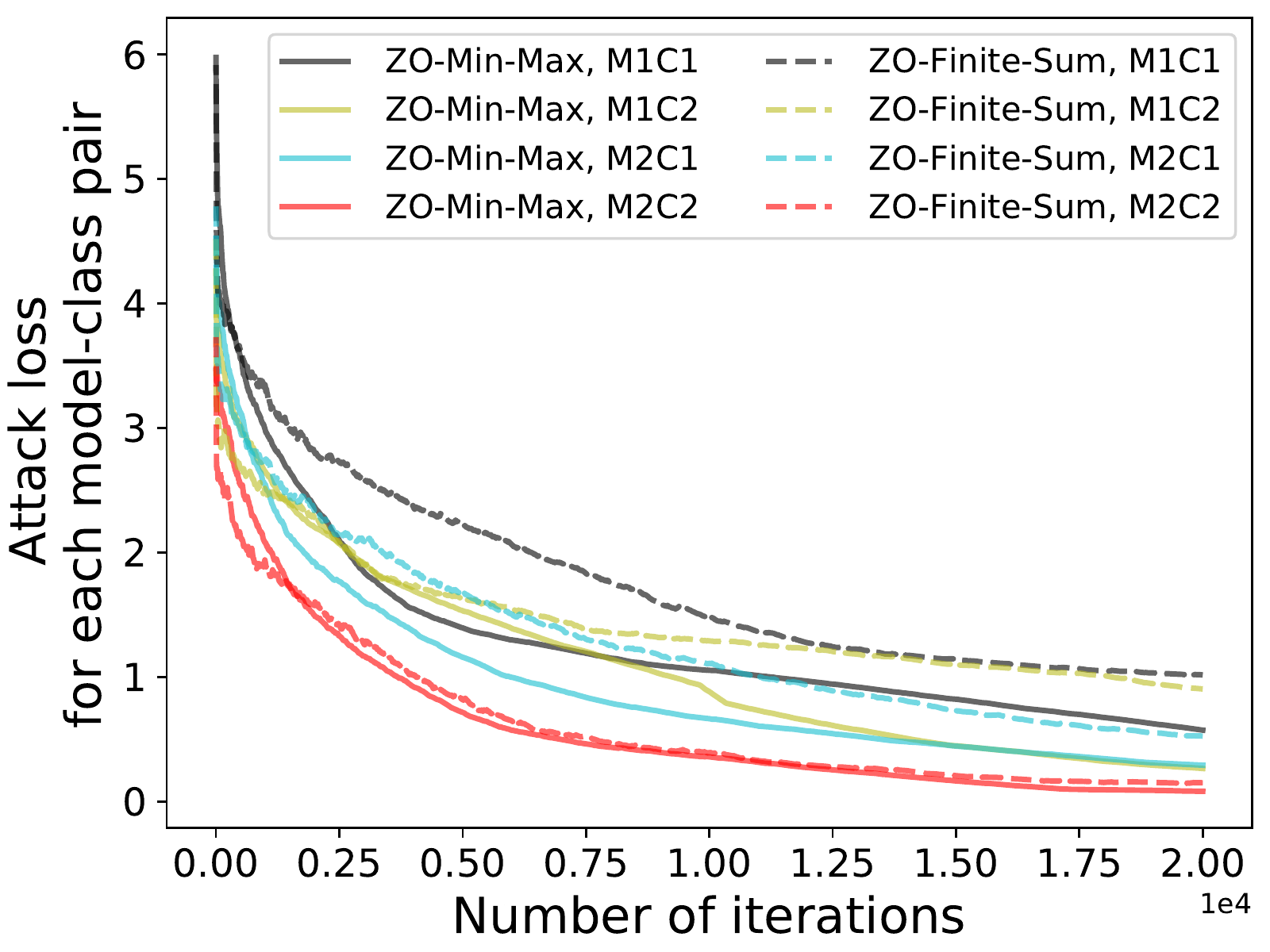} &
\includegraphics[width=.45\textwidth,height=!]{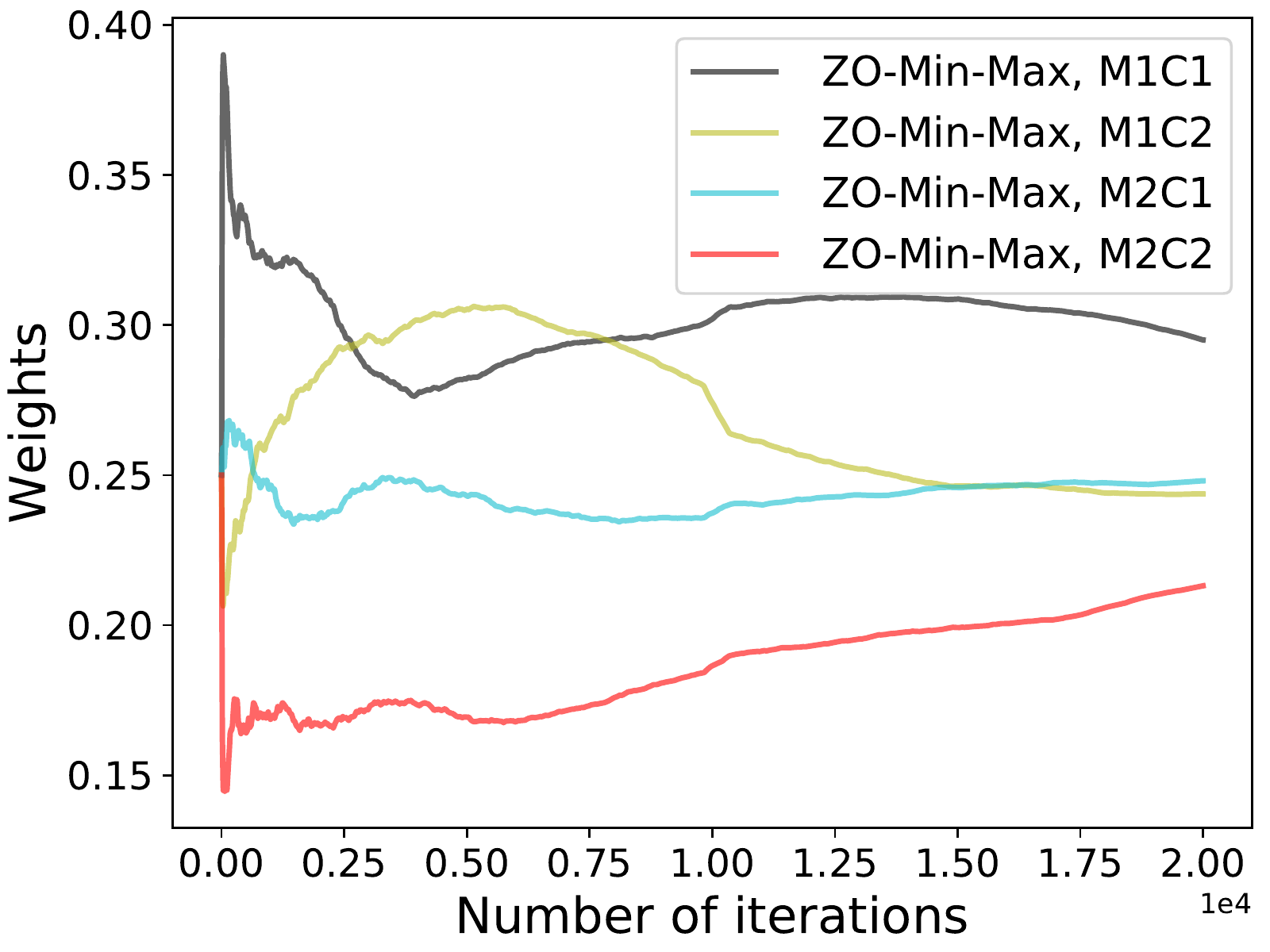}
\\
  (a) & (b)
\end{tabular} 
}
\caption{\footnotesize{Convergence performance of ZO-Min-Max in design of black-box ensemble attack. a)  Attack loss of using ZO-Min-Max vs. ZO-Finite-Sum,
and b)  importance weights learnt from 
 ZO-Min-Max.
}}
  \label{fig: ensemble_attack_app}
\end{figure}

\paragraph{Additional results.}
{In Figure\,\ref{fig: ensemble_attack_app}-(a), 
We   compare ZO-Min-Max with ZO-Finite-Sum, where the latter minimizes
the average loss  over all model-class combinations. As we can see, our approach significantly
improves the worst-case attack performance (corresponding to M$1$C$1$). Here the worst case represents the
most robust model-class pair against the attack. This suggests that ZO-Min-Max takes into account different robustness levels of model-class pairs through the design of importance weights $\mathbf w$. This can also be
evidenced from Figure\,\ref{fig: ensemble_attack_app}-(b):  M$1$C$1$ has the largest weight while  M$2$C$2$ corresponds to the smallest weight.
}

\begin{figure}[htb]
  \begin{center}
    \includegraphics[width=0.5\textwidth]{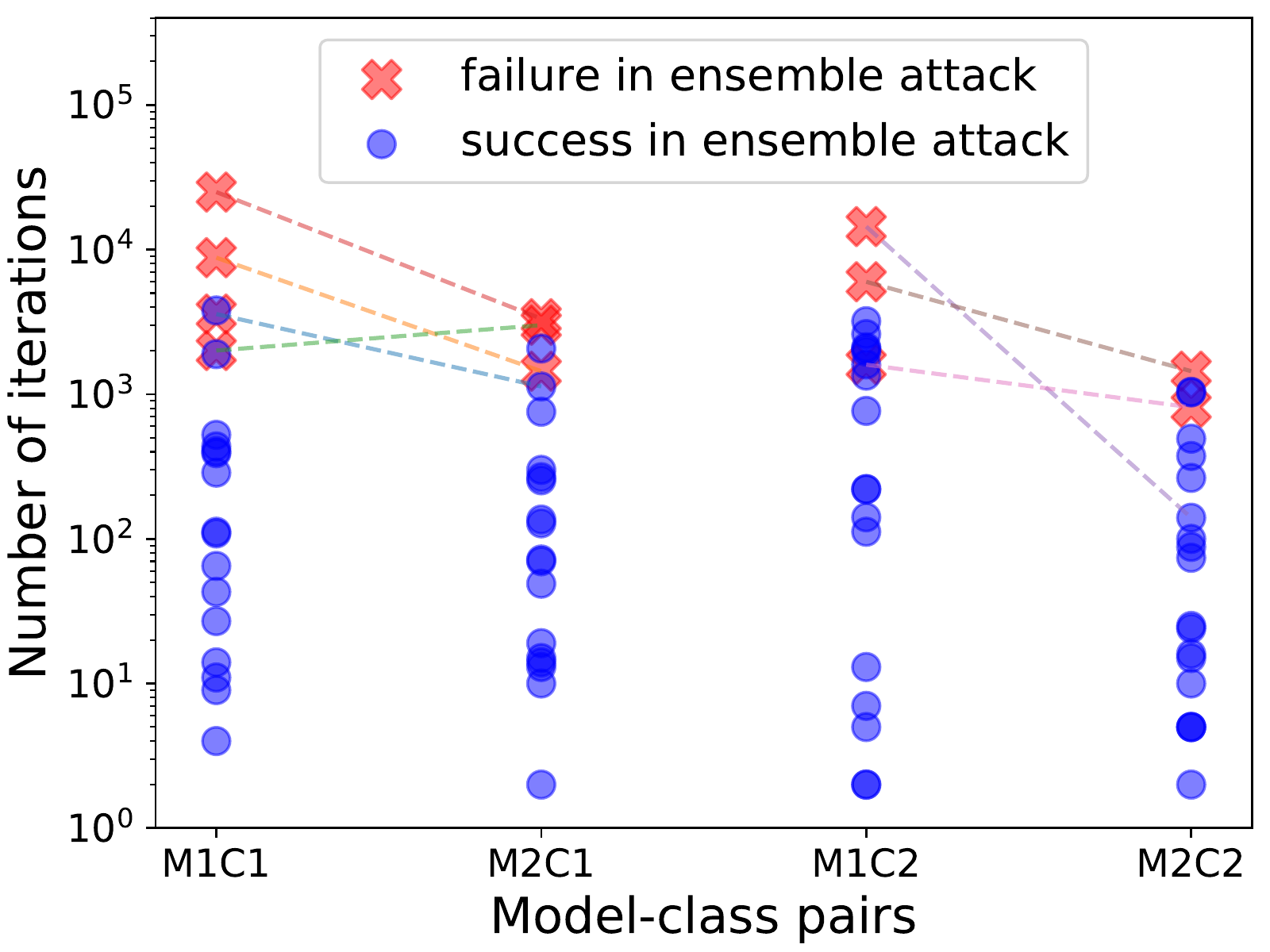}
  \end{center}
      \vspace*{-0.1in}
  \caption{\footnotesize{
  Success or failure of our ensemble attack versus successful  per-image PGD attack.
 }}
  \label{fig: ensemble_vs_perImage_attack}
\end{figure}

{
  In Figure\,\ref{fig: ensemble_vs_perImage_attack}, we  contrast the success or failure (marked by blue or red in the plot) of attacking each image using the obtained universal perturbation $\mathbf x$  with the attacking difficulty (in terms of required iterations for   successful adversarial example) of using per-image non-universal  PGD attack  \citep{madry2017towards}.
    We observe that the success rate of the ensemble universal attack is around $80\%$   at each model-class pair, where the failed cases  (red cross markers)  also need a large amount of iterations to succeed at the case of per-image PGD attack. And images that are difficult to attack keep consistent across models; see  dash lines  to associate the same images between two models in Figure\,\ref{fig: ensemble_vs_perImage_attack}.
}

\clearpage
\newpage

\section{Additional Details on Poisoning Attack}\label{app: poison}

\paragraph{Experiment setup.}
In our experiment, 
we  generate a synthetic dataset that contains $n = 1000$ samples $( \mathbf z_i, t_i )$. We randomly draw the feature vector $\mathbf z_i \in \mathbb R^{100}$ from  $\mathcal N(\mathbf 0, \mathbf I)$, and determine 
 $t_i =1$ if   ${1}/{(1+ e^{- (\mathbf z_i^T \boldsymbol{\theta}^{*} + \nu_i) })} > 0.5$. Here we choose $\boldsymbol{\theta}^{*} = \mathbf 1$ as the ground-truth model parameters, and  $\nu_i \in \mathcal N(0, 10^{-3})$ as  random noise. We randomly split the generated dataset into the training dataset $\mathcal D_{\mathrm{tr}}$ $(70\%)$ and the testing dataset $\mathcal D_{\mathrm{te}}$ $(30\%)$.
We specify our learning model as the logistic regression model for binary classification. Thus, the loss function in 
problem \eqref{eq: poison_robust_attack} is chosen as 
   $ F_{\mathrm{tr}}(\mathbf x , \boldsymbol{\theta}; \mathcal D_{\mathrm{tr}}) \Def   h (\mathbf x, \boldsymbol{\theta}; \mathcal D_{\mathrm{tr},1}) +  h (\mathbf 0, \boldsymbol{\theta}; D_{\mathrm{tr},2})$, where $\mathcal D_{\mathrm{tr}} = \mathcal D_{\mathrm{tr},1} \cup  \mathcal D_{\mathrm{tr},2}$, $ \mathcal D_{\mathrm{tr},1}$ represents the subset of the training dataset that will be poisoned, $|\mathcal D_{\mathrm{tr},1}|/|\mathcal D_{\mathrm{tr}}|$ denotes the poisoning ratio, 
   $h (\mathbf x, \boldsymbol{\theta}; \mathcal D) = - ( {1}/{|\mathcal D|} ) \sum_{(\mathbf z_i, t_i) \in \mathcal D} [t_i \log ( h(\mathbf x, \boldsymbol{\theta}; \mathbf z_i) )
+ (1-t_i) \log ( 1- h(\mathbf x, \boldsymbol{\theta}; \mathbf z_i) )] $, and $h(\mathbf x , \boldsymbol{\theta}; \mathbf z_i) =1/(1+e^{-(\mathbf z_i+ \mathbf x)^T \boldsymbol{\theta}})$.  In problem \eqref{eq: poison_robust_attack}, we also set $\epsilon = 2$ and $\lambda = 10^{-3}$.

 In 
  Algorithm\,\ref{alg: ZO_2side}, unless specified otherwise we choose the  the mini-batch size $b = 100$, the number of random direction vectors $q =5$, the learning rate $\alpha = 0.02$ and $\beta = 0.05$, and the total number of iterations $T = 50000$. We report the empirical results over $10$ independent trials with random initialization.

\begin{figure}[htb]
\centerline{ 
\begin{tabular}{ccc}
\includegraphics[width=.5\textwidth,height=!]{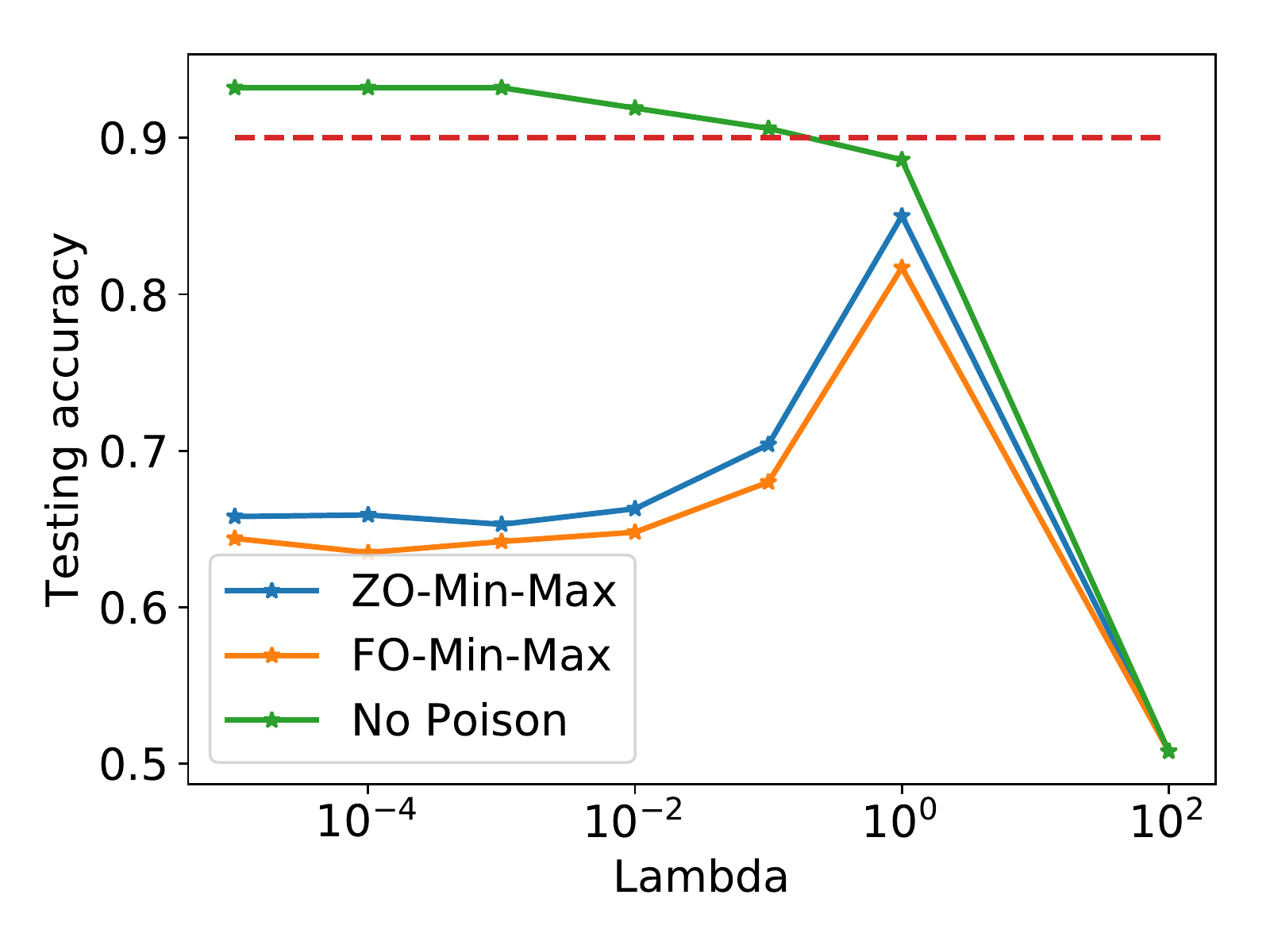}  
\end{tabular} 
}
\caption{\footnotesize{ Empirical performance of ZO-Min-Max in design of poisoning attack:  Testing accuracy   versus regularization parameter $\lambda$. 
}}
  \label{fig: poison_attack_reg}
\end{figure}

\paragraph{Addition results.}
{In Figure\,\ref{fig: poison_attack_reg}, we show the testing accuracy of the poisoned model as the regularization parameter $\lambda$ varies. We observe that the poisoned model accuracy could be improved  as $\lambda$ increases, e.g., $\lambda = 1$. However, this leads to a decrease in   clean model accuracy   (below $90\%$ at $\lambda = 1$). This implies a robustness-accuracy tradeoff.
If $\lambda$ continues to increase, both the clean and poisoned accuracy will decrease dramatically as the training loss in   \eqref{eq: poison_robust_attack} is less optimized.}

\begin{figure}[htb]
\centerline{ 
\begin{tabular}{c}
\includegraphics[width=.5\textwidth,height=!]{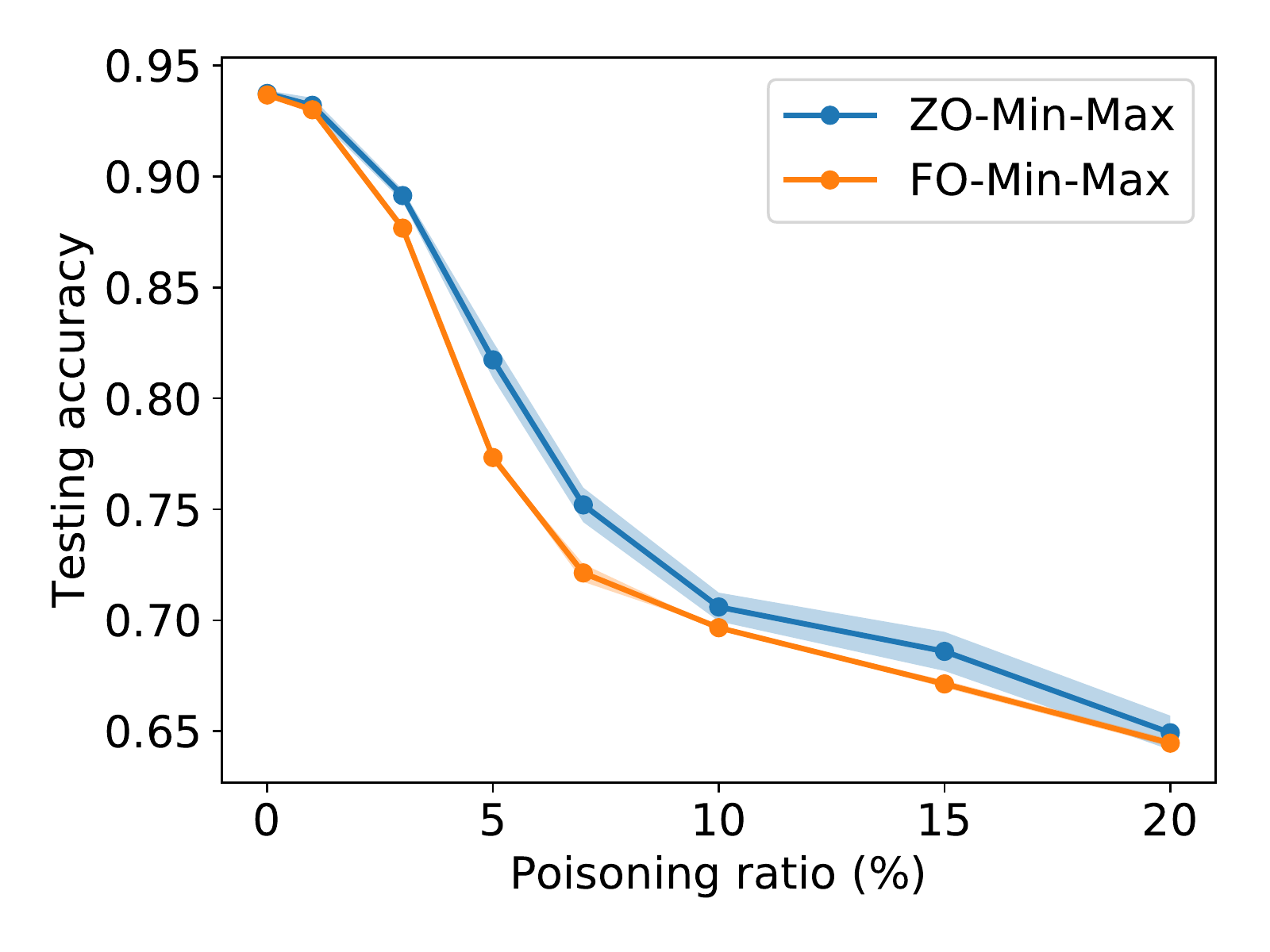} 
\end{tabular} 
}
\caption{\footnotesize{{Testing accuracy of data poisoning attacks generated by ZO-Min-Max (vs. FO-Min-Max) for different  data poisoning ratios. 
}}}
  \label{fig: poison_attack_app}
\end{figure}

In Figure\,\ref{fig: poison_attack_app}, we present the testing accuracy of the learnt model under different data poisoning ratios. As we can see,  only $5\%$ poisoned training data can significantly break the testing accuracy of a well-trained model.  

%% file: Main_ICML2020.bbl
\begin{thebibliography}{69}
\providecommand{\natexlab}[1]{#1}
\providecommand{\url}[1]{\texttt{#1}}
\expandafter\ifx\csname urlstyle\endcsname\relax
  \providecommand{\doi}[1]{doi: #1}\else
  \providecommand{\doi}{doi: \begingroup \urlstyle{rm}\Url}\fi

\bibitem[Aggarwal et~al.(2019)Aggarwal, Bouneffouf, Samulowitz, Buesser, Hoang,
  Khurana, Liu, Pedapati, Ram, Rawat, et~al.]{aggarwal2019can}
Aggarwal, C., Bouneffouf, D., Samulowitz, H., Buesser, B., Hoang, T., Khurana,
  U., Liu, S., Pedapati, T., Ram, P., Rawat, A., et~al.
\newblock How can ai automate end-to-end data science?
\newblock \emph{arXiv preprint arXiv:1910.14436}, 2019.

\bibitem[Al-Dujaili et~al.(2018{\natexlab{a}})Al-Dujaili, Hemberg, and
  O'Reilly]{al2018approximating}
Al-Dujaili, A., Hemberg, E., and O'Reilly, U.-M.
\newblock Approximating nash equilibria for black-box games: A bayesian
  optimization approach.
\newblock \emph{arXiv preprint arXiv:1804.10586}, 2018{\natexlab{a}}.

\bibitem[Al-Dujaili et~al.(2018{\natexlab{b}})Al-Dujaili, Huang, Hemberg, and
  O’Reilly]{al2018adversarial}
Al-Dujaili, A., Huang, A., Hemberg, E., and O’Reilly, U.-M.
\newblock Adversarial deep learning for robust detection of binary encoded
  malware.
\newblock In \emph{2018 IEEE Security and Privacy Workshops (SPW)}, pp.\
  76--82. IEEE, 2018{\natexlab{b}}.

\bibitem[Al-Dujaili et~al.(2018{\natexlab{c}})Al-Dujaili, Srikant, Hemberg, and
  O'Reilly]{al2018application}
Al-Dujaili, A., Srikant, S., Hemberg, E., and O'Reilly, U.-M.
\newblock On the application of danskin's theorem to derivative-free minimax
  optimization.
\newblock \emph{arXiv preprint arXiv:1805.06322}, 2018{\natexlab{c}}.

\bibitem[Audet \& Hare(2017)Audet and Hare]{audet2017derivative}
Audet, C. and Hare, W.
\newblock \emph{Derivative-free and blackbox optimization}.
\newblock Springer, 2017.

\bibitem[Balasubramanian \& Ghadimi(2018)Balasubramanian and
  Ghadimi]{Balasubramanian2018nips}
Balasubramanian, K. and Ghadimi, S.
\newblock Zeroth-order (non)-convex stochastic optimization via conditional
  gradient and gradient updates.
\newblock In \emph{Advances in Neural Information Processing Systems}, pp.\
  3455--3464, 2018.

\bibitem[Berahas et~al.(2019)Berahas, Cao, Choromanski, and
  Scheinberg]{berahas2019theoretical}
Berahas, A.~S., Cao, L., Choromanski, K., and Scheinberg, K.
\newblock A theoretical and empirical comparison of gradient approximations in
  derivative-free optimization.
\newblock \emph{arXiv preprint arXiv:1905.01332}, 2019.

\bibitem[Bogunovic et~al.(2018)Bogunovic, Scarlett, Jegelka, and
  Cevher]{bogunovic2018adversarially}
Bogunovic, I., Scarlett, J., Jegelka, S., and Cevher, V.
\newblock Adversarially robust optimization with gaussian processes.
\newblock In \emph{Proc. of Advances in Neural Information Processing Systems},
  pp.\  5765--5775, 2018.

\bibitem[Boyd \& Vandenberghe(2004)Boyd and Vandenberghe]{boyd2004convex}
Boyd, S. and Vandenberghe, L.
\newblock \emph{Convex optimization}.
\newblock Cambridge university press, 2004.

\bibitem[Branke \& Rosenbusch(2008)Branke and Rosenbusch]{branke2008new}
Branke, J. and Rosenbusch, J.
\newblock New approaches to coevolutionary worst-case optimization.
\newblock In \emph{International Conference on Parallel Problem Solving from
  Nature}, pp.\  144--153. Springer, 2008.

\bibitem[Carlini \& Wagner(2017)Carlini and Wagner]{carlini2017towards}
Carlini, N. and Wagner, D.
\newblock Towards evaluating the robustness of neural networks.
\newblock In \emph{Security and Privacy (SP), 2017 IEEE Symposium on}, pp.\
  39--57. IEEE, 2017.

\bibitem[Chen et~al.(2018)Chen, Yi, and Gu]{chen2018frank}
Chen, J., Yi, J., and Gu, Q.
\newblock A frank-wolfe framework for efficient and effective adversarial
  attacks.
\newblock \emph{arXiv preprint arXiv:1811.10828}, 2018.

\bibitem[Chen et~al.(2017)Chen, Zhang, Sharma, Yi, and Hsieh]{chen2017zoo}
Chen, P.-Y., Zhang, H., Sharma, Y., Yi, J., and Hsieh, C.-J.
\newblock Zoo: Zeroth order optimization based black-box attacks to deep neural
  networks without training substitute models.
\newblock In \emph{Proceedings of the 10th ACM Workshop on Artificial
  Intelligence and Security}, pp.\  15--26. ACM, 2017.

\bibitem[Chen \& Giannakis(2018)Chen and Giannakis]{chen2017bandit}
Chen, T. and Giannakis, G.~B.
\newblock Bandit convex optimization for scalable and dynamic {IoT} management.
\newblock \emph{IEEE Internet of Things Journal}, 2018.

\bibitem[Chen et~al.(2019)Chen, Liu, Sun, and Hong]{chen2018convergence}
Chen, X., Liu, S., Sun, R., and Hong, M.
\newblock On the convergence of a class of adam-type algorithms for non-convex
  optimization.
\newblock \emph{International Conference on Learning Representations}, 2019.

\bibitem[Danskin(1966)]{danskin1966theory}
Danskin, J.~M.
\newblock The theory of max-min, with applications.
\newblock \emph{SIAM Journal on Applied Mathematics}, 14\penalty0 (4):\penalty0
  641--664, 1966.

\bibitem[Deng et~al.(2009)Deng, Dong, Socher, Li, Li, and
  Fei-Fei]{deng2009imagenet}
Deng, J., Dong, W., Socher, R., Li, L.-J., Li, K., and Fei-Fei, L.
\newblock Imagenet: A large-scale hierarchical image database.
\newblock In \emph{Computer Vision and Pattern Recognition, 2009. CVPR 2009.
  IEEE Conference on}, pp.\  248--255. IEEE, 2009.

\bibitem[Dhurandhar et~al.(2019)Dhurandhar, Pedapati, Balakrishnan, Chen,
  Shanmugam, and Puri]{dhurandhar2019model}
Dhurandhar, A., Pedapati, T., Balakrishnan, A., Chen, P.-Y., Shanmugam, K., and
  Puri, R.
\newblock Model agnostic contrastive explanations for structured data.
\newblock \emph{arXiv preprint arXiv:1906.00117}, 2019.

\bibitem[Duchi et~al.(2015)Duchi, Jordan, Wainwright, and
  Wibisono]{duchi2015optimal}
Duchi, J.~C., Jordan, M.~I., Wainwright, M.~J., and Wibisono, A.
\newblock Optimal rates for zero-order convex optimization: The power of two
  function evaluations.
\newblock \emph{IEEE Transactions on Information Theory}, 61\penalty0
  (5):\penalty0 2788--2806, 2015.

\bibitem[Finlay \& Oberman(2019)Finlay and Oberman]{finlay2019scaleable}
Finlay, C. and Oberman, A.~M.
\newblock Scaleable input gradient regularization for adversarial robustness.
\newblock \emph{arXiv preprint arXiv:1905.11468}, 2019.

\bibitem[Flokas et~al.(2019)Flokas, Vlatakis-Gkaragkounis, and
  Piliouras]{flokas2019efficiently}
Flokas, L., Vlatakis-Gkaragkounis, E.-V., and Piliouras, G.
\newblock Efficiently avoiding saddle points with zero order methods: No
  gradients required.
\newblock \emph{arXiv preprint arXiv:1910.13021}, 2019.

\bibitem[Gao et~al.(2014)Gao, Jiang, and Zhang]{gao2014information}
Gao, X., Jiang, B., and Zhang, S.
\newblock On the information-adaptive variants of the {ADMM}: an iteration
  complexity perspective.
\newblock \emph{Optimization Online}, 12, 2014.

\bibitem[Ghadimi \& Lan(2013)Ghadimi and Lan]{ghadimi2013stochastic}
Ghadimi, S. and Lan, G.
\newblock Stochastic first-and zeroth-order methods for nonconvex stochastic
  programming.
\newblock \emph{SIAM Journal on Optimization}, 23\penalty0 (4):\penalty0
  2341--2368, 2013.

\bibitem[Ghadimi et~al.(2016)Ghadimi, Lan, and Zhang]{ghadimi2016mini}
Ghadimi, S., Lan, G., and Zhang, H.
\newblock Mini-batch stochastic approximation methods for nonconvex stochastic
  composite optimization.
\newblock \emph{Mathematical Programming}, 155\penalty0 (1-2):\penalty0
  267--305, 2016.

\bibitem[Gidel et~al.(2017)Gidel, Jebara, and Lacoste-Julien]{gidel17a}
Gidel, G., Jebara, T., and Lacoste-Julien, S.
\newblock {Frank-Wolfe Algorithms for Saddle Point Problems}.
\newblock In \emph{Proceedings of the 20th International Conference on
  Artificial Intelligence and Statistics}, volume~54, pp.\  362--371. PMLR,
  20--22 Apr 2017.

\bibitem[Golovin et~al.(2019)Golovin, Karro, Kochanski, Lee, Song,
  et~al.]{golovin2019gradientless}
Golovin, D., Karro, J., Kochanski, G., Lee, C., Song, X., et~al.
\newblock Gradientless descent: High-dimensional zeroth-order optimization.
\newblock \emph{arXiv preprint arXiv:1911.06317}, 2019.

\bibitem[Goodfellow et~al.(2014)Goodfellow, Shlens, and
  Szegedy]{goodfellow2014explaining}
Goodfellow, I.~J., Shlens, J., and Szegedy, C.
\newblock Explaining and harnessing adversarial examples.
\newblock \emph{arXiv preprint arXiv:1412.6572}, 2014.

\bibitem[Hamedani et~al.(2018)Hamedani, Jalilzadeh, Aybat, and
  Shanbhag]{yazdandoost2018iteration}
Hamedani, E.~Y., Jalilzadeh, A., Aybat, N.~S., and Shanbhag, U.~V.
\newblock Iteration complexity of randomized primal-dual methods for
  convex-concave saddle point problems.
\newblock \emph{arXiv preprint arXiv:1806.04118}, 2018.

\bibitem[He et~al.(2016)He, Zhang, Ren, and Sun]{he2016deep}
He, K., Zhang, X., Ren, S., and Sun, J.
\newblock Deep residual learning for image recognition.
\newblock In \emph{Proceedings of the IEEE conference on computer vision and
  pattern recognition}, pp.\  770--778, 2016.

\bibitem[Herrmann(1999)]{herrmann1999genetic}
Herrmann, J.~W.
\newblock A genetic algorithm for minimax optimization problems.
\newblock In \emph{CEC}, volume~2, pp.\  1099--1103. IEEE, 1999.

\bibitem[Ilyas et~al.(2018)Ilyas, Engstrom, Athalye, and Lin]{ilyas2018black}
Ilyas, A., Engstrom, L., Athalye, A., and Lin, J.
\newblock Black-box adversarial attacks with limited queries and information.
\newblock \emph{arXiv preprint arXiv:1804.08598}, 2018.

\bibitem[Ilyas et~al.(2019)Ilyas, Engstrom, and Madry]{ilyas2018prior}
Ilyas, A., Engstrom, L., and Madry, A.
\newblock Prior convictions: Black-box adversarial attacks with bandits and
  priors.
\newblock In \emph{International Conference on Learning Representations}, 2019.
\newblock URL \url{https://openreview.net/forum?id=BkMiWhR5K7}.

\bibitem[Jagielski et~al.(2018)Jagielski, Oprea, Biggio, Liu, Nita-Rotaru, and
  Li]{jagielski2018manipulating}
Jagielski, M., Oprea, A., Biggio, B., Liu, C., Nita-Rotaru, C., and Li, B.
\newblock Manipulating machine learning: Poisoning attacks and countermeasures
  for regression learning.
\newblock In \emph{2018 IEEE Symposium on Security and Privacy (SP)}, pp.\
  19--35. IEEE, 2018.

\bibitem[Jensen(2003)]{jensen2003new}
Jensen, M.~T.
\newblock A new look at solving minimax problems with coevolutionary genetic
  algorithms.
\newblock In \emph{Metaheuristics: computer decision-making}, pp.\  369--384.
  Springer, 2003.

\bibitem[Jin et~al.(2019)Jin, Netrapalli, and Jordan]{jin2019minmax}
Jin, C., Netrapalli, P., and Jordan, M.~I.
\newblock Minmax optimization: Stable limit points of gradient descent ascent
  are locally optimal.
\newblock \emph{arXiv preprint arXiv:1902.00618}, 2019.

\bibitem[Jones et~al.(2001)Jones, Oliphant, Peterson, et~al.]{scipy2001}
Jones, E., Oliphant, T., Peterson, P., et~al.
\newblock {SciPy}: Open source scientific tools for {Python}, 2001.
\newblock URL \url{http://www.scipy.org/}.

\bibitem[Larson et~al.(2019)Larson, Menickelly, and Wild]{larson2019derivative}
Larson, J., Menickelly, M., and Wild, S.~M.
\newblock Derivative-free optimization methods.
\newblock \emph{Acta Numerica}, 28:\penalty0 287--404, 2019.

\bibitem[Li et~al.(2019)Li, Li, Wang, Zhang, and Gong]{li2019nattack}
Li, Y., Li, L., Wang, L., Zhang, T., and Gong, B.
\newblock Nattack: Learning the distributions of adversarial examples for an
  improved black-box attack on deep neural networks.
\newblock \emph{arXiv preprint arXiv:1905.00441}, 2019.

\bibitem[Liu et~al.(2018{\natexlab{a}})Liu, Zhang, and Yu]{liu2018caad}
Liu, J., Zhang, W., and Yu, N.
\newblock Caad 2018: Iterative ensemble adversarial attack.
\newblock \emph{arXiv preprint arXiv:1811.03456}, 2018{\natexlab{a}}.

\bibitem[Liu et~al.(2018{\natexlab{b}})Liu, Chen, Chen, and
  Hero]{liu2017zeroth}
Liu, S., Chen, J., Chen, P.-Y., and Hero, A.~O.
\newblock Zeroth-order online admm: Convergence analysis and applications.
\newblock In \emph{Proceedings of the Twenty-First International Conference on
  Artificial Intelligence and Statistics}, volume~84, pp.\  288--297, April
  2018{\natexlab{b}}.

\bibitem[Liu et~al.(2018{\natexlab{c}})Liu, Kailkhura, Chen, Ting, Chang, and
  Amini]{liu2018_NIPS}
Liu, S., Kailkhura, B., Chen, P.-Y., Ting, P., Chang, S., and Amini, L.
\newblock Zeroth-order stochastic variance reduction for nonconvex
  optimization.
\newblock In \emph{Proc. of Advances in Neural Information Processing Systems},
  2018{\natexlab{c}}.

\bibitem[Liu et~al.(2019)Liu, Chen, Chen, and Hong]{liu2018signsgd}
Liu, S., Chen, P.-Y., Chen, X., and Hong, M.
\newblock sign{SGD} via zeroth-order oracle.
\newblock In \emph{Proc. of International Conference on Learning
  Representations}, 2019.
\newblock URL \url{https://openreview.net/forum?id=BJe-DsC5Fm}.

\bibitem[Liu et~al.(2016)Liu, Chen, Liu, and Song]{liu2016delving}
Liu, Y., Chen, X., Liu, C., and Song, D.
\newblock Delving into transferable adversarial examples and black-box attacks.
\newblock \emph{arXiv preprint arXiv:1611.02770}, 2016.

\bibitem[{Lu} et~al.(2019){Lu}, {Tsaknakis}, and {Hong}]{luts19}
{Lu}, S., {Tsaknakis}, I., and {Hong}, M.
\newblock Block alternating optimization for non-convex min-max problems:
  Algorithms and applications in signal processing and communications.
\newblock In \emph{Proc. of IEEE International Conference on Acoustics, Speech
  and Signal Processing}, pp.\  4754--4758, May 2019.

\bibitem[Lu et~al.(2019)Lu, Tsaknakis, Hong, and Chen]{lu2019hybrid}
Lu, S., Tsaknakis, I., Hong, M., and Chen, Y.
\newblock Hybrid block successive approximation for one-sided non-convex
  min-max problems: Algorithms and applications.
\newblock \emph{arXiv preprint arXiv:1902.08294}, 2019.

\bibitem[Madry et~al.(2018)Madry, Makelov, Schmidt, Tsipras, and
  Vladu]{madry2017towards}
Madry, A., Makelov, A., Schmidt, L., Tsipras, D., and Vladu, A.
\newblock Towards deep learning models resistant to adversarial attacks.
\newblock \emph{ICLR}, 2018.

\bibitem[Moosavi-Dezfooli et~al.(2019)Moosavi-Dezfooli, Fawzi, Uesato, and
  Frossard]{moosavi2019robustness}
Moosavi-Dezfooli, S.-M., Fawzi, A., Uesato, J., and Frossard, P.
\newblock Robustness via curvature regularization, and vice versa.
\newblock In \emph{Proceedings of the IEEE Conference on Computer Vision and
  Pattern Recognition}, pp.\  9078--9086, 2019.

\bibitem[Nesterov(2007)]{nesterov2007dual}
Nesterov, Y.
\newblock Dual extrapolation and its applications to solving variational
  inequalities and related problems.
\newblock \emph{Mathematical Programming}, 109\penalty0 (2-3):\penalty0
  319--344, 2007.

\bibitem[Nesterov \& Spokoiny(2015)Nesterov and Spokoiny]{nesterov2015random}
Nesterov, Y. and Spokoiny, V.
\newblock Random gradient-free minimization of convex functions.
\newblock \emph{Foundations of Computational Mathematics}, 2\penalty0
  (17):\penalty0 527--566, 2015.

\bibitem[Nouiehed et~al.(2019)Nouiehed, Sanjabi, Lee, and
  Razaviyayn]{nouiehed2019solving}
Nouiehed, M., Sanjabi, M., Lee, J.~D., and Razaviyayn, M.
\newblock Solving a class of non-convex min-max games using iterative first
  order methods.
\newblock \emph{arXiv preprint arXiv:1902.08297}, 2019.

\bibitem[Parikh et~al.(2014)Parikh, Boyd, et~al.]{parikh2014proximal}
Parikh, N., Boyd, S., et~al.
\newblock Proximal algorithms.
\newblock \emph{Foundations and Trends{\textregistered} in Optimization},
  1\penalty0 (3):\penalty0 127--239, 2014.

\bibitem[Picheny et~al.(2019)Picheny, Binois, and Habbal]{picheny2019bayesian}
Picheny, V., Binois, M., and Habbal, A.
\newblock A bayesian optimization approach to find nash equilibria.
\newblock \emph{Journal of Global Optimization}, 73\penalty0 (1):\penalty0
  171--192, 2019.

\bibitem[Qian et~al.(2019)Qian, Zhu, Tang, Jin, Sun, and Li]{qian2019robust}
Qian, Q., Zhu, S., Tang, J., Jin, R., Sun, B., and Li, H.
\newblock Robust optimization over multiple domains.
\newblock In \emph{Proceedings of the AAAI Conference on Artificial
  Intelligence}, volume~33, pp.\  4739--4746, 2019.

\bibitem[Rafique et~al.(2018)Rafique, Liu, Lin, and Yang]{rafique2018non}
Rafique, H., Liu, M., Lin, Q., and Yang, T.
\newblock Non-convex min-max optimization: Provable algorithms and applications
  in machine learning.
\newblock \emph{arXiv preprint arXiv:1810.02060}, 2018.

\bibitem[Rios \& Sahinidis(2013)Rios and Sahinidis]{rios2013derivative}
Rios, L.~M. and Sahinidis, N.~V.
\newblock Derivative-free optimization: a review of algorithms and comparison
  of software implementations.
\newblock \emph{Journal of Global Optimization}, 56\penalty0 (3):\penalty0
  1247--1293, 2013.

\bibitem[Sanjabi et~al.(2018{\natexlab{a}})Sanjabi, Ba, Razaviyayn, and
  Lee]{Sanjabi18}
Sanjabi, M., Ba, J., Razaviyayn, M., and Lee, J.~D.
\newblock On the convergence and robustness of training gans with regularized
  optimal transport.
\newblock In \emph{Proceedings of the 32Nd International Conference on Neural
  Information Processing Systems}, pp.\  7091--7101, 2018{\natexlab{a}}.

\bibitem[Sanjabi et~al.(2018{\natexlab{b}})Sanjabi, Ba, Razaviyayn, and
  Lee]{sanjabi2018convergence}
Sanjabi, M., Ba, J., Razaviyayn, M., and Lee, J.~D.
\newblock On the convergence and robustness of training gans with regularized
  optimal transport.
\newblock In \emph{Advances in Neural Information Processing Systems}, pp.\
  7091--7101, 2018{\natexlab{b}}.

\bibitem[Schmiedlechner et~al.(2018)Schmiedlechner, Al-Dujaili, Hemberg, and
  O'Reilly]{schmiedlechner2018towards}
Schmiedlechner, T., Al-Dujaili, A., Hemberg, E., and O'Reilly, U.-M.
\newblock Towards distributed coevolutionary gans.
\newblock \emph{arXiv preprint arXiv:1807.08194}, 2018.

\bibitem[Shamir(2017)]{shamir2017optimal}
Shamir, O.
\newblock An optimal algorithm for bandit and zero-order convex optimization
  with two-point feedback.
\newblock \emph{Journal of Machine Learning Research}, 18\penalty0
  (52):\penalty0 1--11, 2017.

\bibitem[Steinhardt et~al.(2017)Steinhardt, Koh, and
  Liang]{steinhardt2017certified}
Steinhardt, J., Koh, P. W.~W., and Liang, P.~S.
\newblock Certified defenses for data poisoning attacks.
\newblock In \emph{Advances in neural information processing systems}, pp.\
  3517--3529, 2017.

\bibitem[Szegedy et~al.(2016)Szegedy, Vanhoucke, Ioffe, Shlens, and
  Wojna]{szegedy2016rethinking}
Szegedy, C., Vanhoucke, V., Ioffe, S., Shlens, J., and Wojna, Z.
\newblock Rethinking the inception architecture for computer vision.
\newblock In \emph{IEEE Conference on Computer Vision and Pattern Recognition
  (CVPR)}, pp.\  2818--2826, 2016.

\bibitem[Tran et~al.(2018)Tran, Li, and Madry]{tran2018spectral}
Tran, B., Li, J., and Madry, A.
\newblock Spectral signatures in backdoor attacks.
\newblock In \emph{Advances in Neural Information Processing Systems}, pp.\
  8000--8010, 2018.

\bibitem[Tu et~al.(2018)Tu, Ting, Chen, Liu, Zhang, Yi, Hsieh, and
  Cheng]{tu2018autozoom}
Tu, C.-C., Ting, P., Chen, P.-Y., Liu, S., Zhang, H., Yi, J., Hsieh, C.-J., and
  Cheng, S.-M.
\newblock Autozoom: Autoencoder-based zeroth order optimization method for
  attacking black-box neural networks.
\newblock \emph{arXiv preprint arXiv:1805.11770}, 2018.

\bibitem[Wald(1945)]{wald1945statistical}
Wald, A.
\newblock Statistical decision functions which minimize the maximum risk.
\newblock \emph{Annals of Mathematics}, pp.\  265--280, 1945.

\bibitem[Wang et~al.(2019{\natexlab{a}})Wang, Yao, Shan, Li, Viswanath, Zheng,
  and Zhao]{wang2019neural}
Wang, B., Yao, Y., Shan, S., Li, H., Viswanath, B., Zheng, H., and Zhao, B.~Y.
\newblock Neural cleanse: Identifying and mitigating backdoor attacks in neural
  networks.
\newblock \emph{Neural Cleanse: Identifying and Mitigating Backdoor Attacks in
  Neural Networks}, pp.\ ~0, 2019{\natexlab{a}}.

\bibitem[Wang \& Wu(2019)Wang and Wu]{wang2019flo}
Wang, C. and Wu, Q.
\newblock Flo: Fast and lightweight hyperparameter optimization for automl.
\newblock \emph{arXiv preprint arXiv:1911.04706}, 2019.

\bibitem[Wang et~al.(2019{\natexlab{b}})Wang, Zhang, Liu, Chen, Xu, Fardad, and
  Li]{wang2019unified}
Wang, J., Zhang, T., Liu, S., Chen, P.-Y., Xu, J., Fardad, M., and Li, B.
\newblock Towards a unified min-max framework for adversarial exploration and
  robustness, 2019{\natexlab{b}}.

\bibitem[Ward et~al.(2019)Ward, Wu, and Bottou]{pmlr-v97-ward19a}
Ward, R., Wu, X., and Bottou, L.
\newblock {A}da{G}rad stepsizes: Sharp convergence over nonconvex landscapes.
\newblock In \emph{Proceedings of the 36th International Conference on Machine
  Learning}, pp.\  6677--6686, 2019.

\bibitem[Watson \& Pollack(2001)Watson and Pollack]{watson2001coevolutionary}
Watson, R.~A. and Pollack, J.~B.
\newblock Coevolutionary dynamics in a minimal substrate.
\newblock In \emph{Proceedings of the 3rd Annual Conference on Genetic and
  Evolutionary Computation}, pp.\  702--709. Morgan Kaufmann Publishers Inc.,
  2001.

\end{thebibliography}
